\documentclass{article}

\PassOptionsToPackage{numbers, compress}{natbib}
\usepackage{neurips_data_2024}

\usepackage[utf8]{inputenc} 
\usepackage[T1]{fontenc}    
\usepackage{hyperref}       
\usepackage{url}            
\usepackage{booktabs}       
\usepackage{amsfonts}       
\usepackage{nicefrac}       
\usepackage{microtype}      
\usepackage{xcolor}         
\usepackage{enumitem}
\usepackage{subcaption}
\title{From Perfect to Noisy World Simulation:\\ Customizable Embodied Multi-modal Perturbations for SLAM Robustness Benchmarking}

\author{%
  Xiaohao Xu${}^{1}$,  Tianyi Zhang${}^{2}$, Sibo Wang${}^{1}$,  Xiang Li${}^{2}$,  Yongqi Chen${}^{1}$, Ye Li${}^{1}$, \\ \textbf{Bhiksha Raj}${}^{2}$,   \textbf{Matthew Johnson-Roberson}${}^{2}$, \textbf{Xiaonan Huang}${}^{1}$ \\ {${}^{1}$ University of Michigan, Ann Arbor}, ${}^{2}$ Carnegie Mellon University
}

\usepackage{times}
\usepackage{pifont}
\usepackage{multirow}
\usepackage{collcell}
\usepackage{colortbl}
\usepackage{tikz}
\usepackage{tabularx}
\usepackage{graphicx}
\usepackage{amsmath}
\usepackage{cases}
\usepackage{bbding}
\usepackage{wrapfig}
\usepackage{utfsym}
\usepackage{amsthm, thmtools}

\newcommand{\xxh}[1]{{\color{black}  #1}}

\usepackage{caption}
\usepackage{multirow}

\usepackage{changes}
\usepackage{pifont}

\usepackage{array}
\usepackage{makecell}

\definechangesauthor[name={Xiaohao Xu}, color=orange]{X.}
\definechangesauthor[name={Xiaohao Xu}, color=red]{Y.}

\begin{document}

\maketitle

\begin{abstract}
Embodied agents require robust navigation systems to operate in unstructured environments, making the robustness of Simultaneous Localization and Mapping (SLAM) models critical to embodied agent autonomy. While real-world datasets are invaluable, simulation-based benchmarks offer a scalable approach for robustness evaluations. However, the creation of a challenging and controllable noisy world with diverse perturbations remains under-explored.
To this end, we propose a novel, customizable pipeline for noisy data synthesis, aimed at assessing the resilience of multi-modal SLAM models against various perturbations. 
 The pipeline comprises a comprehensive taxonomy of sensor and motion perturbations for embodied multi-modal (specifically RGB-D) sensing, categorized by their sources and propagation order, allowing for procedural composition. We also provide a toolbox for synthesizing these perturbations, enabling the transformation of clean environments into challenging noisy simulations. 
Utilizing the pipeline, we instantiate the large-scale \textit{Noisy-Replica} benchmark, which includes diverse perturbation types, to evaluate the risk tolerance of existing advanced RGB-D SLAM models. Our extensive analysis uncovers the susceptibilities  of both neural (NeRF and Gaussian Splatting -based) and non-neural SLAM models to disturbances, despite their demonstrated accuracy in standard benchmarks. Our code is publicly available at \url{https://github.com/Xiaohao-Xu/SLAM-under-Perturbation}.

\end{abstract}

\section{Introduction}
\label{sec:introduction}
{T}he growing prevalence of embodied agents deployed in complex and dynamic environments~\cite{kaufmann2023champion,SubT}, \textit{i.e.}, \textit{Noisy World}, underscores the critical need for robustness in embodied systems. This robustness, essential for effective operation, is significantly influenced by the agent's ability to withstand  perturbations.
Consequently, robustness evaluation in such settings~\cite{RobustNav} has emerged as a critical research area.
For embodied agents, Simultaneous Localization and Mapping (SLAM)~\cite{slam-survey1,TNNLS-SLAM-survey} is a fundamental task to achieving autonomy.  Therefore, our focus is on developing a comprehensive and reliable benchmark to assess SLAM robustness against disturbances.

Recent advances in embodied SLAM system assessment have primarily focused on collecting challenging datasets. These datasets expose SLAM systems to domain-specific environmental degradation, broadening our understanding of real-world operational challenges~\cite{oxford-robotcar,Multi-Spectral,Burri25012016,Geiger2012CVPR,zhang2021multi,pfrommer2017penncosyvio,TUM-RGBD,zuniga2020vi,carlevaris2016university,Delmerico19icra,helmberger2022hilti,tian2023resilient,schubert2018tum,zhao2024subt}. 
However, due to the inherent difficulties in data collection and labeling in the wild, {existing real-world datasets remain limited in size,} hindering holistic evaluation.
To overcome these limitations, simulation-based benchmarks~\cite{RobustNav,VI-navigation-with-simulation,dosovitskiy2017carla,tartanvo,driving-matrix} have emerged as a promising approach. They offer the advantage of creating infinite and diverse scenarios for rigorous testing of SLAM models. Additionally, these benchmarks allow for the adaptive crafting of increasingly challenging environments, driving the continuous improvement of SLAM robustness~\cite{tartanvo}. Simulated methods also enable the study of the disentangled effects of individual perturbations on SLAM performance, revealing potential weaknesses. While current simulators may not fully replicate real-world fidelity, rapid advancements in visual content synthesis~\cite{rombach2022high,raistrick2023infinite} are progressively closing this gap.

\begin{figure*}
\centering 
    \includegraphics[width=\linewidth]{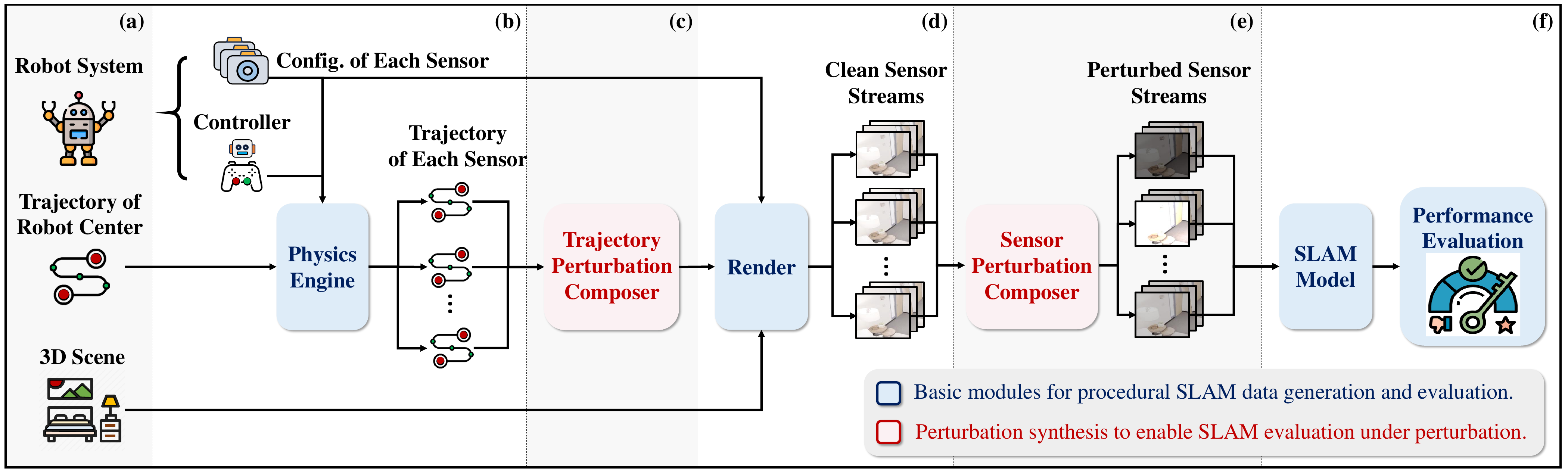} 
    \caption{\textbf{Noisy data synthesis for robustness evaluation of embodied perception (specifically SLAM) models under perturbations.} Our insight is to customize perturbations (\textcolor[rgb]{0.75,0.0,0.0}{red} blocks)  during  conventional procedural (clean) data generation (\textcolor[rgb]{0.5,0.6,1.0}{blue} blocks).}
    \label{fig:overview-simulator-pipeline}
\end{figure*}

\begin{table}[t]
\caption{Comparison on synthetic visual SLAM benchmarks.}
\centering 
\label{tab:datasets}
\setlength{\tabcolsep}{4pt}
\resizebox{\textwidth}{!}
{
\begin{tabular}{l|rl|r|c|cccc|c}
\toprule
\toprule

\multirow{2}{*}{{\textbf{Benchmark}}} &
  \multirow{2}{*}{{\textbf{\#Seq}}}  &
  \multirow{2}{*}{{\textbf{Modality}}} &\multirow{2}{*}{\begin{tabular}[c]{@{}c@{}}\textbf{\#Perturbed} \\ \textbf{Setting}\end{tabular} }  &\multirow{2}{*}{\begin{tabular}[c]{@{}c@{}}\textbf{Multi-modal} \\ \textbf{Perturbation}\end{tabular} }    & \multicolumn{4}{c|}{{\textbf{Perturbation Category}}}  &  \multirow{2}{*}{\begin{tabular}[c]{@{}c@{}}\textbf{Editable} \\ \textbf{Capability}\end{tabular} }  
 
 \\ \cmidrule{6-9} %
 &
   &    &    &  & 
  {RGB}  &
  {Motion} &
  {Depth} &
  {RGB-D Sync.} &
  {} 
 \\  %
 \midrule
Replica~\cite{straub2019replica} & 8   & RGB-D & 0  &   \textcolor{red}{\ding{55}}  &   \textcolor{red}{\ding{55}} &   \textcolor{red}{\ding{55}} &   \textcolor{red}{\ding{55}} &   \textcolor{red}{\ding{55}} &   \textcolor{red}{\ding{55}}\\ 
TartanAir~\cite{wang2020tartanair} & 30  & RGB &   \begin{tabular}[c]{@{}c@{}}8\end{tabular}  &   \textcolor{red}{\ding{55}}   &  \textcolor[rgb]{0,0.6875,0.3125}{\ding{51}} &   \textcolor[rgb]{0,0.6875,0.3125}{\ding{51}}&   \textcolor{red}{\ding{55}} &   \textcolor{red}{\ding{55}} &   \textcolor{red}{\ding{55}} \\ 
\begin{tabular}[c]{@{}c@{}}{\textbf{Noisy-Replica}} (\textbf{Ours}) \end{tabular} & {\textbf{{1,000}}}  &  \textbf{RGB-D} & \begin{tabular}[c]{@{}c@{}}\textbf{124} \end{tabular} &   \textcolor[rgb]{0,0.6875,0.3125}{\ding{51}}   &   \textcolor[rgb]{0,0.6875,0.3125}{\ding{51}} &   \textcolor[rgb]{0,0.6875,0.3125}{\ding{51}} &   \textcolor[rgb]{0,0.6875,0.3125}{\ding{51}} &   \textcolor[rgb]{0,0.6875,0.3125}{\ding{51}} &   \textcolor[rgb]{0,0.6875,0.3125}{{\ding{51}}} 
  \\ \bottomrule \bottomrule
\end{tabular}}
\end{table}

Despite the increasing availability of nearly photo-realistic 3D scene datasets and simulators for SLAM evaluation~\cite{straub2019replica,dai2017scannet,deitke2022️,zheng2020structured3d}, they often lack varied and controllable disturbances. Consequently, these simulations typically represent idealized, perturbation-free environments (\textit{Perfect World}), leaving the simulated perturbed environment (\textit{Noisy World}) largely unexplored.
To address this gap, we propose a comprehensive perturbation taxonomy for embodied multi-modal (specifically RGB-D) sensing systems. This taxonomy includes perturbations originating from RGB-D sensing (RGB imaging and depth imaging corruptions), locomotion of the embodied agent (motion-related deviations), and communication among multiple sensors (multi-sensor de-synchronization). We illustrate how these perturbations propagate within the system, enabling the composition of mixed perturbations.
Based on this taxonomy, we develop a perturbation composition toolbox that seamlessly integrates with existing simulation tools~\cite{denninger2020blenderproc,deitke2022️,straub2019replica}, transforming the simulated environment from a \textit{Perfect World} into a more challenging \textit{Noisy World} for robustness evaluation. As shown in Fig.~\ref{fig:overview-simulator-pipeline}, we propose a  noisy data synthesis pipeline, designed to assess SLAM resilience under customizable perturbations. This pipeline adapts to different hardware configurations (\textit{e.g.}, sensor placement) and software components (\textit{e.g.}, SLAM models), incorporating both motion and sensor perturbations with varying severity levels to simulate sensor pose vibrations and environmental disturbances.

To assess SLAM robustness, we use the proposed comprehensive taxonomy of perturbations and noisy data synthesis pipeline to instantiate the \textit{Noisy-Replica} benchmark, based on photo-realistic 3D scenes from the \textit{Replica}~\cite{straub2019replica} dataset. As shown in Table~\ref{tab:datasets}, \textit{Noisy-Replica} surpasses existing benchmarks in diversity and scope. It offers editable perturbation capabilities covering 124 distinct RGB-D perturbation settings across 1,000 long video sequences, which is two orders of magnitude larger than standard SLAM benchmarks~\cite{straub2019replica, TUM-RGBD}. We then analyze the effects of individual perturbations on both  neural (using NeRF~\cite{rosinol2022nerf} or Gaussian Splat~\cite{gaussiansplatting} map representations)  and non-neural SLAM models. Our findings reveal that while advanced SLAM models excel in standard clean SLAM benchmarks~\cite{TUM-RGBD,straub2019replica}, they exhibit vulnerabilities and a propensity for failure when exposed to perturbations. Furthermore, we demonstrate that combined perturbations pose greater challenges for SLAM systems, and the interaction of multiple perturbation types can create complex impacts.

To summarize, our contributions are: 
\textbf{1}) We propose a comprehensive taxonomy of perturbations for embodied multi-modal (specifically RGB-D) sensing systems, along with a perturbation synthesis toolbox.
\textbf{2}) We introduce a noisy data synthesis pipeline for customizable robustness assessment. Focusing on the SLAM task, we utilize this pipeline to initialize the first large-scale RGB-D SLAM robustness benchmark, \textit{Noisy-Replica}, featuring diverse editable sensor and motion perturbations.
\textbf{3}) To the best of our knowledge, we conduct the first robustness study of neural RGB-D SLAM models under perturbations. Our extensive dataset and robustness benchmarking offer a systematic approach and environment to evaluate SLAM models, revealing the vulnerabilities of both existing neural and non-neural SLAM models to individual and combined perturbations.

\section{Related Work}\label{sec:related_work}

\noindent\textbf{{Robustness benchmarking.}}
To ensure the reliable deployment of mobile robots, their perception modules must demonstrate resilience to shifts in natural distributions~\cite{zhang2017resilient,taori2020measuring}. A pioneering robustness benchmark~\cite{hendrycks2019robustness} analyzes image corruption robustness by evaluating the performance of image classification methods against common corruptions and perturbations. Building upon this, subsequent research has expanded the scope of investigation to encompass other perception tasks. These tasks include 2D/3D object detection \cite{michaelis2019benchmarking,carlson2018modeling,kong2023robodepth, kong2023robo3d}, segmentation \cite{kamann2020benchmarking,xu2022towards,li2023robust}, and embodied navigation \cite{RobustNav,yokoyama2022benchmarking}.
These studies underscore the significance of evaluating models' robustness to corruptions. In SLAM, the challenges extend beyond just handling image-level corruptions, like those due to camera malfunctions. It is also crucial to account for dynamic variations in sensor corruption and deviations in sensor transformation simultaneously over time. These variations arise from time-variant environmental effects and the diverse motion of robots, respectively. 
In this study, we propose a perturbation taxonomy for RGB-D SLAM in dynamic (\textit{e.g.}, varying illumination) and unstructured environments (\textit{e.g.,} uneven terrains that can cause vibrations for mobile robots).

\noindent\textbf{{SLAM methods.}}
This overview highlights visual-related SLAM systems. More comprehensive reviews of SLAM systems can be obtained from various resources such as \cite{slam-survey1,macario2022comprehensive,kazerouni2022survey}. Classical single-modal SLAM methods, like visual-only models exemplified by ORB-SLAM~\cite{orbslam}, have demonstrated remarkable accuracy in `clean' benchmark settings \cite{schubert2018tum,straub2019replica}.
To address the complexities of real-world environments, researchers have explored various techniques \cite{orbslam2,orbslam3,MIMOSA,rosinol2021kimera} that incorporate multi-view sensors and fuse diverse modalities, such as visual-inertia and RGB-D. 
Furthermore, several approaches \cite{teed2021droid,tartanvo,coslam,imap,niceslam,zhang2023goslam,rosinol2022nerf} have utilized neural networks and neural representations to enhance generalization and improve the dense mapping quality. Despite notable improvements in accuracy, the robustness of these models against perturbations remains under-explored.

\noindent\textbf{{Robustness evaluation for SLAM.}}
The robustness of SLAM systems is essential for their reliable and accurate operation in dynamic and challenging real-world environments~\cite{slam-survey1}. This robustness is critical not only to handle sensor faults, but also to ensure long-term performance.
To facilitate the robustness evaluation of SLAM models, several datasets \cite{schubert2018tum,helmberger2022hilti,tian2023resilient,SubT,zhao2023subt}  have been collected in degraded environments with perturbations such as low illumination or motion blur.  
Furthermore, SLAMBench \cite{bujanca2021robust} compares the performance of several classical SLAM models across multiple challenging datasets and reveals the vulnerability of these SLAM models. 
Considering that constructing real-world datasets via robot platforms for SLAM can be challenging and unscalable, Wang \textit{et al.} \cite{wang2020tartanair} have utilized photo-realistic simulation environments to create a pioneering simulated SLAM benchmark called TartanAir for robustness evaluation. In this study, we expand the scope of evaluation to include the robustness of multi-modal SLAM models—encompassing both classical and neural SLAM methods—against a broader spectrum of sensor corruptions and motion patterns (\textit{e.g.}, varying speed and \textcolor{black}{motion-induced deviations of sensors' trajectories}).

\begin{figure*}[ht!]
	\centering
\includegraphics[width=\textwidth]{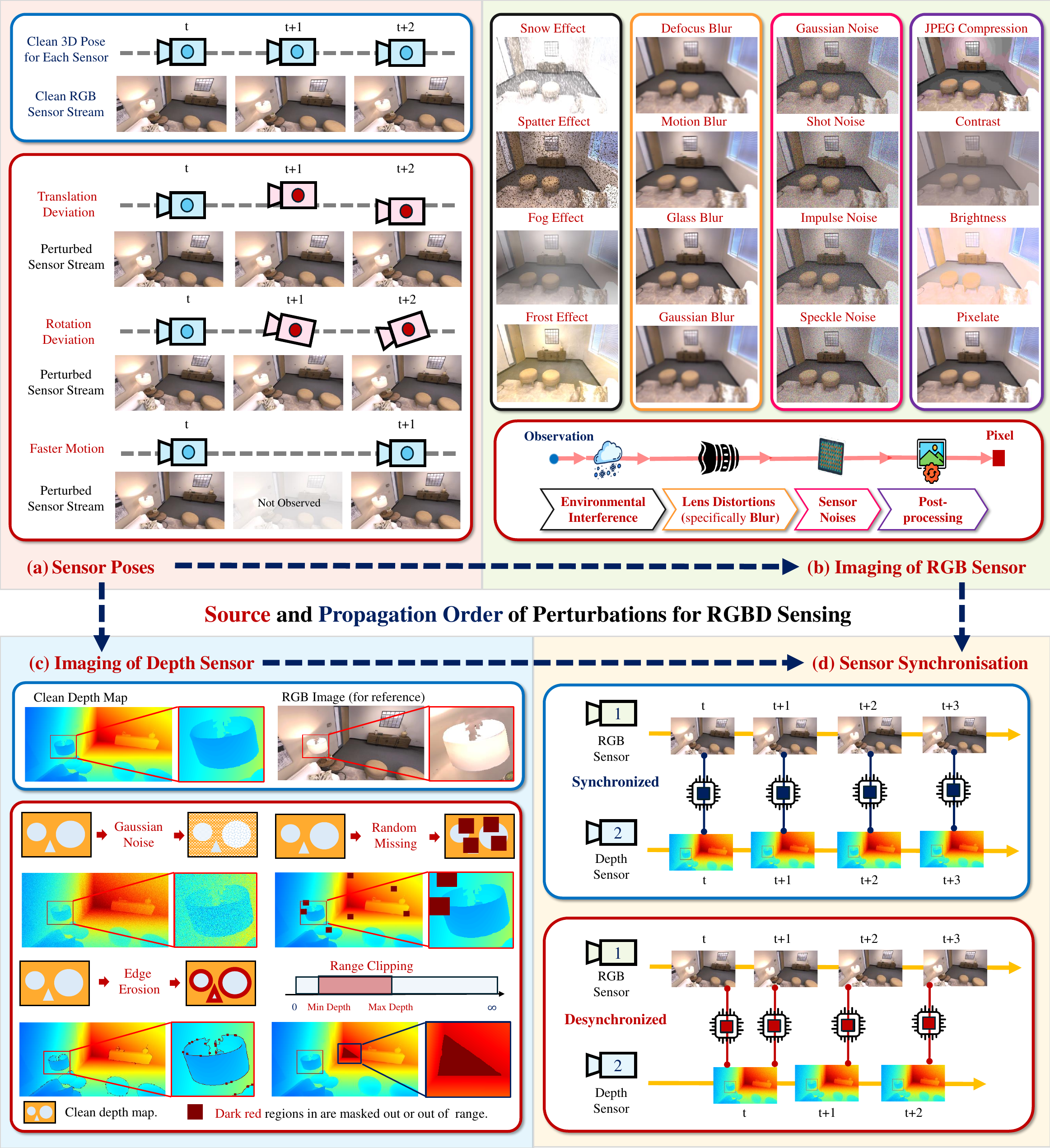} 
	\caption{\xxh{\textbf{Taxonomy of perturbations for embodied RGB-D sensing.} The sources of perturbations include: (\textbf{a}) sensor pose errors, (\textbf{b}) RGB and (\textbf{c}) depth imaging corruptions, and (\textbf{d}) RGB-D sensor synchronization errors. Dashed arrows illustrate the propagation order of individual perturbations.} } 
	\label{fig:all-perturb-taxonomy}
\end{figure*}

\section{Formulation of SLAM under Perturbations}

SLAM aims to concurrently construct a map of the environment and estimate the pose (position and orientation) of an embodied agent~\cite{probabilistic_robot}. Given a sequence of observations $\mathbf{z}_{1:t}$ from timestamp $1$ to $t$, the goal is to estimate the environmental map $\mathbf{m}$ (\textit{e.g.}, 3D mesh or point cloud) and the trajectory $\mathbf{x}_{1:t}$ over time. \xxh{The agent's pose at a specific timestamp $i$ ($1\leq i \leq t$) is denoted by spatial coordinates and orientation.}
The probabilistic posterior articulates our belief about the map and trajectory based on the observations which encodes the effect of ego motions: 
\begin{equation}
p(\mathbf{m}, \mathbf{x}_{1:t} | \mathbf{z}_{1:t})
\label{eq:posterior}
\end{equation}
\xxh{where $\mathbf{x}_t = [\mathbf{R}_t, \mathbf{t}_t]$ with $\mathbf{R}_t \in \boldsymbol{SO}(3)$ being a rotation matrix and $\mathbf{t}_t \in \mathbb{R}^3$ a translation vector for each timestamp $t$.}

For embodied agents, perturbations can arise from three main sources: sensor imaging, motion, and communication. \textbf{1}) Imaging perturbations refer to noise and processing effects introduced to sensor readings. \textbf{2}) Motion perturbations result from locomotion complexities (\textit{e.g.}, vibrations) and agility (\textit{e.g.}, fast and dynamic movements), leading to complex sensor pose sequences and unstable observations. \textbf{3}) Communication perturbations can cause de-synchronization of multiple sensors.

\section{Noisy Data Synthesis with  Customizable Perturbations}\label{sec:method}

\subsection{{Noisy Data Synthesis Pipeline}}

Fig.~\ref{fig:overview-simulator-pipeline} shows the proposed noisy data synthesis pipeline for model robustness benchmarking of embodied RGB-D sensing systems. The pipeline is highly customizable and incorporates controllable perturbations to simulate sensor noises and locomotion disturbances.

 The initial phase is to configure the robot system, the \xxh{desired trajectory of the robot center (\textit{e.g.}, the center of gravity) in the world frame}, and the 3D scene (see Fig.~\ref{fig:overview-simulator-pipeline}a). We can optionally utilize off-the-shelf physics engines (such as MuJoCo~\cite{mujoco}) in conjunction with motion controllers to obtain \xxh{the trajectory (\textit{i.e.}, sensor poses along time) of each individual sensor in the world frame} (see Fig.~\ref{fig:overview-simulator-pipeline}b).  Subsequently, these sensor-specific trajectories are passed to the trajectory perturbation composer to introduce motion deviations to the pose, thereby better emulating vibrations of sensors on a mobile robot (see Fig.~\ref{fig:overview-simulator-pipeline}c). The render, implemented via OpenGL~\cite{shreiner2009opengl},  derives clean sensor data streams conditional on the trajectory and sensor configurations (see Fig.~\ref{fig:overview-simulator-pipeline}d). Sensor imaging and synchronization perturbations are introduced into the sensor streams to mimic real-world observational anomalies and sensor failures (see Fig.~\ref{fig:overview-simulator-pipeline}e).   By utilizing the generated noisy data, which encompasses perturbed sensor streams as inputs and perturbed trajectory and 3D scene as ground-truth labels, the robustness of SLAM models to perturbations can be rigorously assessed (see Fig.~\ref{fig:overview-simulator-pipeline}f).

\subsection{{Perturbation Taxonomy for Embodied RGB-D Sensing}}

\noindent\textbf{Perturbation sources.} As shown in Fig. \ref{fig:all-perturb-taxonomy}, perturbations affecting embodied RGB-D sensing systems can originate from sensor pose deviations, inaccuracies within the RGB-D imaging processes, and de-synchronization issues between RGB and depth sensors. Due to space limitation, we briefly illustrate perturbation sources as follows. Please see Appendix Sec.~\ref{sec:perturbation-taxonomy-appendix} for  details. 

{\textbf{(a) Perturbation on sensor poses.}}
Fig.~\ref{fig:all-perturb-taxonomy}a depicts perturbations affecting sensor poses, encompassing \textit{Motion/Trajectory Deviations} (by applying a rotation perturbation $\Delta \mathbf{R} \in \boldsymbol{SO}(3)$ and a translation perturbation $\Delta \mathbf{t} \in \mathbb{R}^3$) and \textit{Faster Motion Effect } (by downsampling the original sensor stream).

\noindent{\textbf{(b) Perturbation on RGB sensor imaging.}}
\xxh{The perturbations on RGB imaging are designed to model potential error sources throughout the entire RGB image formation and processing pipeline, from the 3D world to the final 2D image. The perturbation sources include environmental interference effects that affect light transmission, blurring effects partially caused by lens-related distortions, sensor noises, and post-processing effects on the image.} Prominent perturbations~\cite{hendrycks2019robustness} (see Fig.~\ref{fig:all-perturb-taxonomy}b) that we considered are: 
\textbf{1}) {{environmental interference}}: \textit{Snow Effect}, \textit{Frost Effect}, \textit{Fog Effect}, and \textit{Spatter effect}; \textbf{2}) {{lens-related distortions (specifically blur)}}: \textit{{Gaussian Blur}}, \textit{{Glass Blur}}, \textit{{Motion Blur}}, and  \textit{{Defocus Blur}}; \textbf{3}) sensor Noises: \textit{Gaussian Noise}, \textit{Shot Noise}, \textit{Impulse Noise}, and \textit{Speckle Noise}; \textbf{4}) {{post-processing}}: \textit{Brightness Increase}, \textit{Contrast Decrease}, \textit{JPEG Compression}, and \textit{Pixelate}.

\newcommand{\colorcell}[1]{%
  \ifdim #1 pt > 0.018pt 
    \cellcolor{gray!50}{#1}
  \else
    \ifdim #1 pt > 0.014 pt 
      \cellcolor{gray!30}{#1}
    \else  
      \ifdim #1 pt > 0.010 pt 
        \cellcolor{gray!15}{#1}
      \else
        #1 
      \fi
    \fi
  \fi
}

\begin{table*}[t]
\caption{Performance (measured by ATE$\downarrow$ (m)) under static (\textbf{Top}) and dynamic (\textbf{Bottom}) RGB imaging perturbations for Neural SLAM models. {Cells with darker \colorbox{lightgray}{shades} indicate higher ATE.}}
\label{tab:image_perturb}
\centering\setlength{\tabcolsep}{0.4mm}
\resizebox{\textwidth}{!}{
\begin{tabular}{l|c|cc|cccc|cccc|cccc|cccc}
    \toprule \toprule
    \multirow{2}{*}{\textbf{Method}} &  \multirow{2}{*}{\textbf{Clean}}&  \multicolumn{2}{|c|}{\textbf{Perturbed}}  & \multicolumn{4}{|c|}{\textbf{Blur Effect}} & \multicolumn{4}{c|}{\textbf{Noise Effect}} & \multicolumn{4}{c|}{\textbf{Environmental Interference}}  & \multicolumn{4}{c}{\textbf{Post-processing}} 
    \\ \cmidrule{5-8} \cmidrule{9-12} \cmidrule{13-16} \cmidrule{17-20}
    &  & \textbf{Mean} & \textbf{Max}  & \textbf{Motion} & \textbf{Defocus} & \textbf{Gaussian} & \textbf{Glass} & \textbf{Gaussian} & \textbf{Shot} & \textbf{Impulse} & \textbf{Speckle} & \textbf{Fog} & \textbf{Frost} & \textbf{Snow} & \textbf{Spatter} & \textbf{Bright} & \textbf{Contra.} & \textbf{Jpeg}  & \textbf{Pixelate}
    \\\midrule\midrule
GO-SLAM (Mono)~\cite{zhang2023goslam} & \colorcell{0.0039} & \colorcell{0.0903}  & \colorcell{0.7207} & \colorcell{0.0151} & \colorcell{0.0052} & \colorcell{0.0052} & \colorcell{0.0089} & \colorcell{0.0776} & \colorcell{0.0456} & \colorcell{0.0296} & \colorcell{0.0190} & \colorcell{0.2157} & \colorcell{0.7207} & \colorcell{0.1921} & \colorcell{0.0859} & \colorcell{0.0046} & \colorcell{0.0047} & \colorcell{0.0095} & \colorcell{0.0046} \\  \midrule
iMAP (RGB-D)~\cite{imap} & \colorcell{0.1209} & \colorcell{0.1568} & \colorcell{0.3831} & \colorcell{0.1424} & \colorcell{0.1671} & \colorcell{0.1811} & \colorcell{0.0672} & \colorcell{0.0278} & \colorcell{0.0779} & \colorcell{0.1710} & \colorcell{0.1087} & \colorcell{0.1913} & \colorcell{0.1316} & \colorcell{0.1665} & \colorcell{0.1473} & \colorcell{0.1903} & \colorcell{0.3831} & \colorcell{0.1884} & \colorcell{0.1669}  \\
Nice-SLAM (RGB-D)~\cite{niceslam}& \colorcell{0.0147} & \colorcell{0.0253} & \colorcell{0.0654}  & \colorcell{0.0307} & \colorcell{0.0151} & \colorcell{0.0161} & \colorcell{0.0188} & \colorcell{0.0254} & \colorcell{0.0377} & \colorcell{0.0353} & \colorcell{0.0151} & \colorcell{0.0186} & \colorcell{0.0160} & \colorcell{0.0323} & \colorcell{0.0320} & \colorcell{0.0654} & \colorcell{0.0161} & \colorcell{0.0150} & \colorcell{0.0145}\\
CO-SLAM (RGB-D)~\cite{coslam} & \colorcell{0.0090} & \colorcell{0.0104} &  \colorcell{0.0125}  & \colorcell{0.0115} & \colorcell{0.0096} & \colorcell{0.0097} & \colorcell{0.0097} & \colorcell{0.0125} & \colorcell{0.0101} & \colorcell{0.0099} & \colorcell{0.0105} & \colorcell{0.0118} & \colorcell{0.0113} & \colorcell{0.0104} & \colorcell{0.0098} & \colorcell{0.0103} & \colorcell{0.0112} & \colorcell{0.0094} & \colorcell{0.0094}\\
GO-SLAM (RGB-D)~\cite{zhang2023goslam} & \colorcell{0.0046} & \colorcell{0.0574} & \colorcell{0.6271} & \colorcell{0.0135} & \colorcell{0.0052} & \colorcell{0.0052} & \colorcell{0.0090} & \colorcell{0.0169} & \colorcell{0.0140} & \colorcell{0.0171} & \colorcell{0.0100} & \colorcell{0.1211} & \colorcell{0.6271} & \colorcell{0.0416} & \colorcell{0.0164} & \colorcell{0.0047} & \colorcell{0.0054} & \colorcell{0.0065} & \colorcell{0.0050} \\
SplaTAM-S (RGB-D)~\cite{keetha2023splatam}& \colorcell{0.0045} & \colorcell{0.0062} & \colorcell{0.0160} & \colorcell{0.0160} & \colorcell{0.0052}  & \colorcell{0.0049} & \colorcell{0.0048} & \colorcell{0.0054} & \colorcell{0.0050} & \colorcell{0.0044} & \colorcell{0.0051} & \colorcell{0.0085} & \colorcell{0.0063} & \colorcell{0.0048} & \colorcell{0.0051} & \colorcell{0.0038} & \colorcell{0.0133} & \colorcell{0.0044} & \colorcell{0.0048}  \\ 
\midrule
    \multicolumn{20}{c}{Optimal performance (min ATE) achieved by using all SLAM models:}  \\ \midrule
  min ATE  & \multirow{3}{*}{${0.0039}$} & {{{{0.0056}}}}   & {${0.0115}$} & $0.0115$ & $0.0052$  & $0.0049$ & $0.0048$ & $0.0054$ & $0.0050$ & $0.0044$ & $0.0051$ & $0.0085$ & $0.0063$ & $0.0048$ & $0.0051$ & $0.0038$ & $0.0054$ & $0.0044$ & $0.0046$  
\\ \cmidrule{5-8}\cmidrule{9-12}\cmidrule{13-16}\cmidrule{17-20}
 mean(min ATE)& & {{{\colorcell{0.0056}}}}   & {\colorcell{0.0066}}  &  \multicolumn{4}{c|}{\colorcell{0.0066}}&  \multicolumn{4}{c|}{\colorcell{0.0050}}&  \multicolumn{4}{c|}{\colorcell{0.0062}} &  \multicolumn{4}{c}{\colorcell{0.0046}}
\\ 
   \midrule     \midrule
GO-SLAM (Mono)~\cite{zhang2023goslam} & \colorcell{0.0039} & \colorcell{0.0933} & \colorcell{0.7395} & \colorcell{0.0155} & \colorcell{0.0065}   & \colorcell{0.0060} & \colorcell{0.0090}  & \colorcell{0.0509} & \colorcell{0.0253} & \colorcell{0.0396} & \colorcell{0.0158} & \colorcell{0.2668} & \colorcell{0.7395}& \colorcell{0.2254}  & \colorcell{0.0474} & \colorcell{0.0066} & \colorcell{0.0050} & \colorcell{0.0298} & \colorcell{0.0044} 
\\
\midrule
iMAP (RGB-D)~\cite{imap} & \colorcell{0.1209} & \colorcell{0.1756} & \colorcell{0.2873} & \colorcell{0.1243} & \colorcell{0.1042} & \colorcell{0.2149}  & \colorcell{0.1221} & \colorcell{0.1354} & \colorcell{0.1170} & \colorcell{0.1967} & \colorcell{0.1576} & \colorcell{0.2279} & \colorcell{0.2873} & \colorcell{0.2412} & \colorcell{0.1528} & \colorcell{0.2141} & \colorcell{0.2576} & \colorcell{0.1607} & \colorcell{0.0955} 
\\
Nice-SLAM (RGB-D)~\cite{niceslam}& \colorcell{0.0147} & \colorcell{0.0214} &\colorcell{0.0409} & \colorcell{0.0157} & \colorcell{0.0252} & \colorcell{0.0359} & \colorcell{0.0211} & \colorcell{0.0288} & \colorcell{0.0409} & \colorcell{0.0146} & \colorcell{0.0155} & \colorcell{0.0167} & \colorcell{0.0211} & \colorcell{0.0197} & \colorcell{0.0187} & \colorcell{0.0206} & \colorcell{0.0155} & \colorcell{0.0146} & \colorcell{0.0170}
\\
CO-SLAM (RGB-D)~\cite{coslam}& \colorcell{0.0090} & \colorcell{0.0105} & \colorcell{0.0117} & \colorcell{0.0107} & \colorcell{0.0095} & \colorcell{0.0115} & \colorcell{0.0093} & \colorcell{0.0106} & \colorcell{0.0103} & \colorcell{0.0102} & \colorcell{0.0098} & \colorcell{0.0117} & \colorcell{0.0116} & \colorcell{0.0111} & \colorcell{0.0109} & \colorcell{0.0106} & \colorcell{0.0111} & \colorcell{0.0095} & \colorcell{0.0097}
\\
GO-SLAM (RGB-D)~\cite{zhang2023goslam} & \colorcell{0.0046} & \colorcell{0.0363}  & \colorcell{0.2213} & \colorcell{0.0130} & \colorcell{0.0057} & \colorcell{0.0055} & \colorcell{0.0078} & \colorcell{0.0185} & \colorcell{0.0117} & \colorcell{0.0139} & \colorcell{0.0098} & \colorcell{0.1685} & \colorcell{0.2213} & \colorcell{0.0637} & \colorcell{0.0166} & \colorcell{0.0051} & \colorcell{0.0052} & \colorcell{0.0092} & \colorcell{0.0049} 
\\ 
SplaTAM-S (RGB-D)~\cite{keetha2023splatam}& \colorcell{0.0045} & \colorcell{0.008} & \colorcell{0.045} & \colorcell{0.0191} & \colorcell{0.0053}  & \colorcell{0.0052} & \colorcell{0.0050} & \colorcell{0.0058} & \colorcell{0.0072} & \colorcell{0.0044} & \colorcell{0.0067} & \colorcell{0.0062} & \colorcell{0.0062} & \colorcell{0.045} & \colorcell{0.0041} & \colorcell{0.0054} & \colorcell{0.0096} & \colorcell{0.0046} & \colorcell{0.0045}  
\\ \midrule

    \multicolumn{20}{c}{{Optimal performance (min ATE) achieved by using all SLAM models:}}  \\ \midrule
  min ATE  & \multirow{3}{*}{${0.0039}$} & {{{{0.0061}}}}   & {${0.0111}$} & ${0.0107}$ & ${0.0053}$  & ${0.0052}$ & ${0.0050}$ & ${0.0058}$ & ${0.0072}$ & ${0.0044}$ & ${0.0067}$ & ${0.0062}$ & ${0.0062}$ & {${0.0111}$} & ${0.0041}$ & ${0.0051}$ & ${0.0050}$ & ${0.0046}$ & ${0.0044}$  
\\ \cmidrule{5-8}\cmidrule{9-12}\cmidrule{13-16}\cmidrule{17-20}
 mean(min ATE)& & {{{\colorcell{0.0061}}}}   & {\colorcell{0.0069}}  &  \multicolumn{4}{c|}{\colorcell{0.0066}}&  \multicolumn{4}{c|}{\colorcell{0.0060}}&  \multicolumn{4}{c|}{\colorcell{0.0069}} &  \multicolumn{4}{c}{\colorcell{0.0048}}
\\ 
\bottomrule  \bottomrule
        \multicolumn{20}{l}{{\textit{{G}} represent settings that include failure sequences where no final trajectory is generated due to running out of GPU memory (more than 48GB).}}\\
        \multicolumn{20}{l}{{{The number in front of \textit{G}} represents the average ATE as failure sequences are set as a value of 1.0.}}\\
\end{tabular}
}
\end{table*}

\noindent{\textbf{(c) Perturbation on depth sensor imaging}} The depth distribution of the existing simulated benchmark Replica~\cite{straub2019replica} differs noticeably from real data TUM-RGBD~\cite{schubert2018tum} (see Appendix Sec.~\ref{subsec:perturbation-taxonomy-appendix-depth}), which motivates us to propose a set of  perturbation operations designed specifically for depth images (see Fig.~\ref{fig:all-perturb-taxonomy}c): \textbf{1}) noise-rated perturbation: \textit{Gaussian Noise}; \textbf{2}) depth missing: \textit{Edge Erosion} and \textit{Random Missing Depth}; \textbf{3}) depth perception limitation: \textit{Range Clipping}.

\noindent{\textbf{(d) Perturbation on RGB-D sensor synchronization.}}  
To emulate sensor delays in cases where multiple sensors within an RGB-D sensing system are not synchronized, we introduce temporal misalignment between the sensor streams (see Fig.~\ref{fig:all-perturb-taxonomy}d). 

\noindent\textbf{Perturbation propagation order.} Within a embodied RGB-D sensing system, the order (dashed arrows of Fig.~\ref{fig:all-perturb-taxonomy}) in which perturbations occur and interact follows the sensing and data processing procedure. Initially, sensor motion deviations directly impact the accuracy of estimated sensor poses, creating a ripple effect downstream. Subsequently, external noises introduced during the RGB and depth imaging process further corrupt the data, compounding the initial pose errors. Finally, desynchronization between multiple sensor streams can lead to misalignment during data fusion.

\noindent\textbf{Perturbation mode and severity.} Perturbations are examined in two modes: static and dynamic. Static perturbations maintain a constant severity throughout a sensor stream or a pose sequence, while dynamic perturbations exhibit frame-to-frame variations, mimicking time-variant perturbations. In addition, the perturbation is investigated on different levels of severity and strength.

\section{Benchmarking RGB-D SLAM Robustness under Perturbations}\label{sec:experiment}

Leveraging our noisy data synthesis pipeline, we instantiate \textit{{Noisy-Replica}}, a benchmark designed for robustness evaluation of {RGB-D} SLAM models under perturbations.
In the following sections, we delve into the details of \textit{Noisy-Replica} benchmark and evaluate the performance of neural and non-neural RGB-D SLAM models under perturbation.

\subsection{{\textit{Noisy-Replica} Benchmark Construction}}

\vspace{1.0mm}\noindent{\textbf{Data source for rendering}}. We render the RGB-D sensor streams using 3D scene models sourced from the \textit{Replica} dataset~\cite{straub2019replica}, which comprises real 3D scans of indoor scenes. We select the same set of eight rooms and offices as the (clean) Replica-SLAM dataset~\cite{imap} for consistent comparison. \xxh{Each sequence has 2,000 frames at 1200×680 resolution.}  See Appendix Sec.~\ref{subsec:assumption} for details on the assumptions used for benchmark instantiation.

\vspace{1.0mm}\noindent{\textbf{Perturbation setup}}. 
We compose a diverse set of RGB-D perturbations using our noisy data synthesis pipeline. 
When motion deviations is disabled, we utilize the same trajectory as~\cite{imap} to render clean sensor streams, and then introduce perturbations for each frame. \textbf{1}) For RGB perturbations, we follow the perturbation magnitudes used in image classification robustness benchmark~\cite{hendrycks2019robustness}. \textbf{2}) For the severity of  depth perturbations, we refer to the depth range and distribution in real-world dataset TUM-RGBD~\cite{schubert2018tum}. \textbf{3}) For motion deviations, we perturb the original trajectories in~\cite{imap} by introducing additional translation and orientation deviations in sensor poses. \textbf{4}) To simulate faster motion, we down-sample sensor streams from the clean source by 2, 4, and 8 times. \textbf{5}) To mimic sensor de-synchronization, we introduce frame delay (5, 10, and 20 frames) between RGB and depth sensor streams. See Appendix Sec.~\ref{subsec:benchmark-stat} for details about benchmark statistics.  

\noindent\textbf{{Benchmarking baseline models.}}
\xxh{While previous SLAM robustness evaluations primarily focus on classical SLAM methods \cite{wang2020tartanair,bujanca2021robust} (\textit{e.g.}, ORB-SLAM3 \cite{orbslam3}), our benchmark additionally encompasses top-performing Neural SLAM models on standard SLAM benchmarks for robustness benchmarking analyses, including iMAP \cite{imap}, Nice-SLAM \cite{niceslam}, CO-SLAM \cite{coslam}, GO-SLAM \cite{zhang2023goslam}, and SplaTAM-S \cite{keetha2023splatam}. } See Appendix Sec.~\ref{subsec:methods} for more details about baseline models.

\noindent\textbf{{Evaluation metrics.}} 
We primarily use Absolute Trajectory Error (ATE) \cite{prokhorov2019measuring} for evaluation. For classical SLAM models, we also use Relative Pose Error (RPE) and Success Rate (SR) \cite{orbslam3,wang2020tartanair} metrics. Smaller ATE and RPE values ($\downarrow$) and larger SR values ($\uparrow$) indicate better performance.

\begin{figure*}[t]
	\centering
\includegraphics[width=\textwidth]{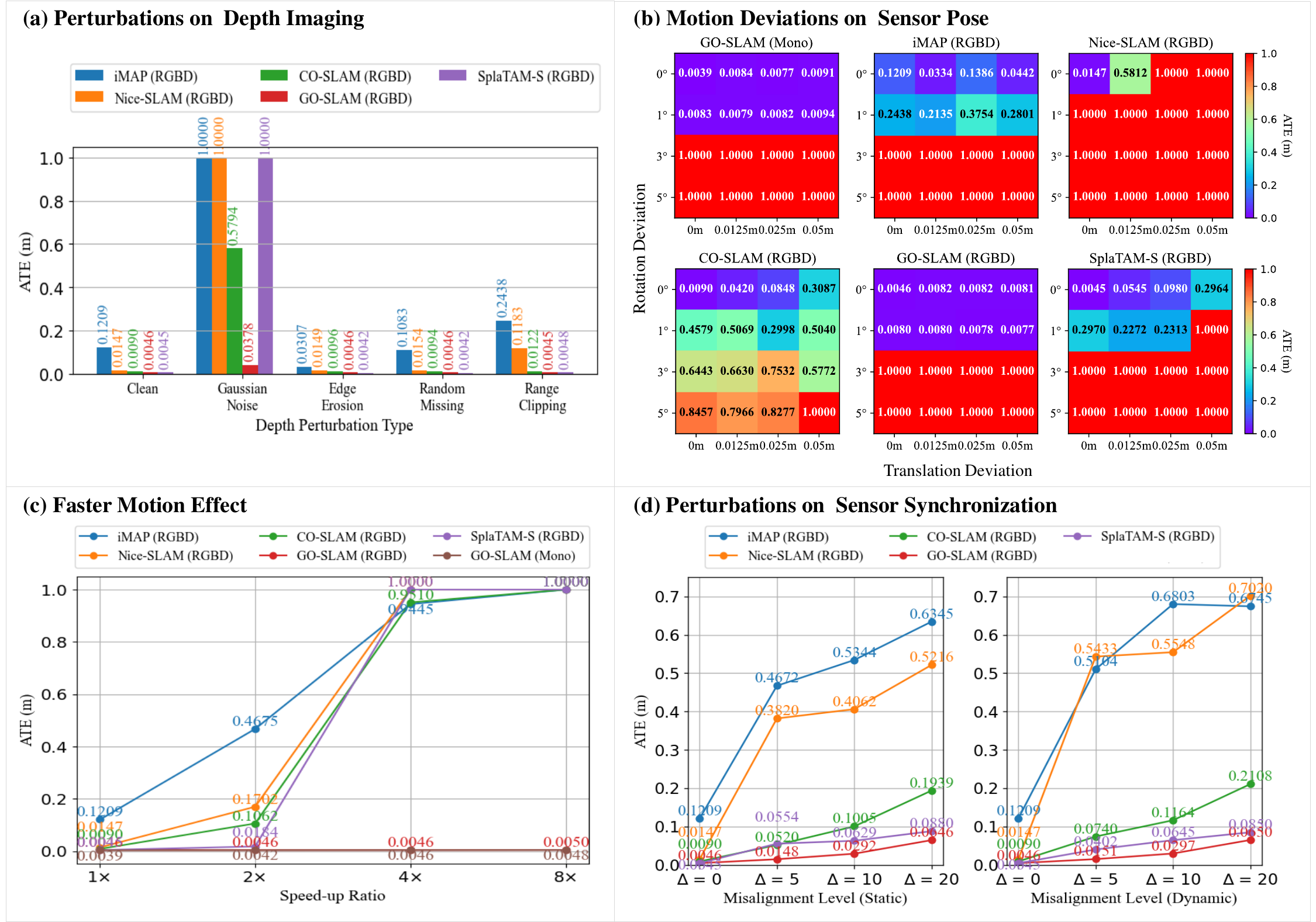}
	\caption{\xxh{Performance (measured by ATE$\downarrow$ (m)) of Neural SLAM models  under diverse perturbations. For visualization, sequences resulting in failure are assigned an ATE value of 1.0.}}
	\label{fig:neural-all}
\end{figure*}

\begin{figure*}[t]
	\centering
\includegraphics[width=\textwidth]{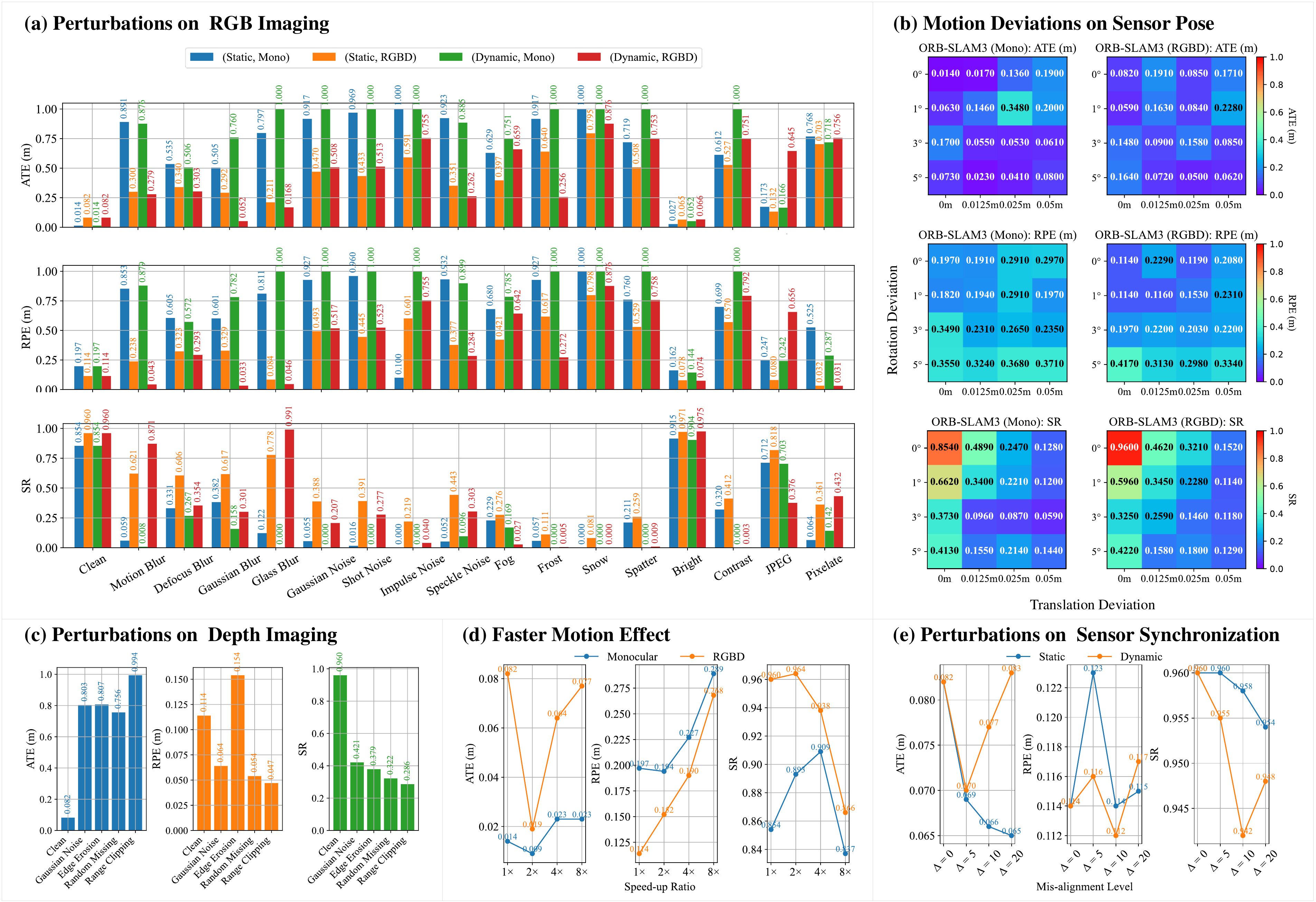}
	\caption{\xxh{Performance (measured by ATE$\downarrow$ (m), RPE$\downarrow$ (m), and SR$\uparrow$) of ORB-SLAM3~\cite{orbslam3} under diverse perturbations. For visualization, sequences resulting in failure are assigned an ATE/RPE value of 1.0 and a Success Rate of 0. }}
	\label{fig:ORBSLAM-all}
\end{figure*}
\subsection{{Benchmarking Results on Isolated Perturbations}}\label{sec:results}

This section summarizes the findings on the \textit{Noisy-Replica} benchmark, evaluating the robustness of RGB-D SLAM models against various perturbations. To reduce randomness, each experiment is conducted three times on eight 3D scenes, averaging the results across 24 experiments per perturbation. Detailed benchmarking tables are available in Appendix Sec.~\ref{sec:detailed-result-with-error-bar}.

\vspace{1.0mm}\noindent{\textbf{Sensor perturbation on RGB imaging.}} We present the performance of  Neural SLAM models and the classical SLAM model ORB-SLAM3 under RGB imaging perturbations in Table~\ref{tab:image_perturb} and Fig.~\ref{fig:ORBSLAM-all}a, respectively. For Neural SLAM models, to gauge the expected average and worst-case performance of each model, we present the mean and maximum ATE values across various perturbation settings. We offer the following analyses and insights: \textbf{1}) Different types of RGB imaging perturbations impact SLAM performance to varying degrees, with environmental effects like adverse weather conditions posing the most significant challenge, followed by sensor noises, while post-processing perturbations like image compression have a relatively minor influence.
\textbf{2}) Dynamic perturbations consistently present a greater challenge than static perturbations for SLAM systems, evident in both individual model performance and the optimal performance achievable by all SLAM models. The performance decline under dynamic perturbations underscores the difficulty of handling real-time visual disturbances.
\textbf{3}) Neural SLAM models exhibit robustness to most noise types due to their learning-based components. In contrast, the non-neural model ORB-SLAM3 encounters complete tracking loss under certain perturbations, resulting in a low success rate of pose tracking.

\vspace{0.5mm}\noindent{\textbf{Sensor perturbation on depth imaging.}}
Fig.~\ref{fig:neural-all}a and Fig.~\ref{fig:ORBSLAM-all}c demonstrate the impact of depth perturbations on Neural SLAM models and the ORB-SLAM3 model, respectively. \textbf{1}) Most neural-based SLAM models exhibit minimal performance degradation when faced with partial depth missing perturbations (\textit{{e.g.}}, \textit{random missing values}, \textit{edge erosion}, and \textit{range clipping}). This robustness can be attributed to their effective pixel-wise optimization mechanisms. In contrast, ORB-SLAM3 suffers significant tracking loss and experiences a notable decline in trajectory estimation performance when encountering missing depth data. \textbf{2}) Introducing \textit{Gaussian noise} to depth maps has a more pronounced impact on all evaluated models compared to missing data. This results in a considerable increase in trajectory estimation error, due to the noise directly interfering with the observed depth.

\vspace{1.0mm}\noindent{\textbf{Motion deviations on sensor poses.}}
As shown in Fig.~\ref{fig:neural-all}b and Fig.~\ref{fig:ORBSLAM-all}b, motion deviations, both in translation and rotation, of sensor poses significantly degrade the trajectory estimation accuracy of SLAM models, even with small deviations like 2.5 cm in translation or 3 degrees in rotation.
The combination of translation and rotation deviations amplifies the trajectory estimation error, leading to failure of nearly all SLAM models.
SplaTAM-S, which exhibits robustness under sensor imaging perturbations, faces failures in most settings under motion deviations.
ORB-SLAM3 and GO-SLAM demonstrate better robustness to motion deviations, likely due to their incorporation of loop closure and global bundle adjustment techniques.
Overall, trajectory estimation accuracy in advanced SLAM models is highly sensitive to motion deviations of sensor poses.

\vspace{1.0mm}\noindent\textbf{Faster motion effect.} \xxh{Fig.~\ref{fig:neural-all}c and Fig.~\ref{fig:ORBSLAM-all}d demonstrate the impact of faster motion effects on Neural SLAM models and ORB-SLAM3,} revealing the limitations of most approaches in achieving acceptable performance at higher speeds. Notably, GO-SLAM excels in handling faster motion scenarios, thanks to the integration of global bundle adjustment mechanism. Besides, the classical SLAM model ORB-SLAM3 also demonstrates robustness in tackling high-speed scenarios.

\vspace{1.0mm}\noindent{\textbf{Perturbation on sensor synchronization.}}
Fig.~\ref{fig:neural-all}d and Fig.~\ref{fig:ORBSLAM-all}e present the result under different severity levels of multi-sensor misalignment, characterized by the frame interval ($\Delta$) between RGB and depth sensor streams. The performance of iMAP and Nice-SLAM significantly deteriorates as the misalignment intervals increase. In contrast, CO-SLAM, GO-SLAM, and SplaTAM-S demonstrate a certain degree of tolerance towards misalignment. Generally, increasing de-synchronization frames leads to larger performance drop. 

\noindent\textbf{Discussion on the effect of isolated perturbations.} Our main takeaways are: \textbf{1}) No single model can handle all perturbed settings. Different SLAM models demonstrate varying robustness across different types of perturbations, emphasizing the need for tailored approaches for specific scenarios and application requirements.
\textbf{2}) There is a lack of correlation between a model's performance in clean and perturbed settings. Methods that excel in standard clean conditions may exhibit significantly degraded performance under specific perturbations.
These findings underscore the importance of evaluating SLAM systems across diverse perturbed settings, in addition to clean settings, to comprehensively and reliably assess their robustness.

\begin{wraptable}{r}{0.52\textwidth}
\vspace{-4mm}
\centering
\caption{{{Effect of mixed perturbations on ATE$\downarrow$ (m).} {Cells with darker \colorbox{lightgray}{shades} indicate higher ATE.} }}
\label{tab:mixed_perturbation}
\centering\setlength{\tabcolsep}{0.5mm}
\resizebox{0.52\textwidth}{!}{
\begin{tabular}{l|c|cccccc}
\toprule\toprule

\textbf{Perturbation Type} &\textbf{Clean}& \multicolumn{6}{c}{\textbf{Perturbation Composition}} \\ \midrule
RGB Snow Effect             & &  \checkmark&  \checkmark&  \checkmark                 & \checkmark &\checkmark &\checkmark \\  
RGB Motion Blur     &    &   &  \checkmark&  \checkmark &\checkmark     &  \checkmark                &\checkmark            \\
RGB Gaussian Noise   &&     &  &      \checkmark   &\checkmark  &\checkmark   &          \checkmark   \\
RGB JPEG Compress.  &    && &         & \checkmark  &\checkmark   &\checkmark                \\ 
Depth Gaussian Noise   &  & &  &         &                  &     \checkmark              &   \checkmark                         \\  
RGBD De-sync.  &   &&   &               &                  &                  &                     \checkmark      \\ \midrule
GO-SLAM~\cite{zhang2023goslam} & {0.005}&  \cellcolor{gray!12}{0.056} & \cellcolor{gray!28}{0.139} & \cellcolor{gray!40}{0.196} & \cellcolor{gray!26}{0.127}  & \cellcolor{gray!42}{0.211} & \cellcolor{gray!66}{0.327}                 \\ 
SplaTAM-S~\cite{keetha2023splatam} & {0.005}&  \cellcolor{gray!01}{0.005} & \cellcolor{gray!28}{0.007} & \cellcolor{gray!40}{0.008} & \cellcolor{gray!40}{0.008}  & \cellcolor{gray!42}{0.132} & \cellcolor{gray!66}{0.307}                 \\ 


\bottomrule\bottomrule
\end{tabular}}
\vspace{-2mm}
\end{wraptable}

\subsection{{Case Study on the Effect of Mixed Perturbations}}

We conduct a case study to investigate the impact of mixed perturbations, combining decoupled perturbations based on their propagation order (see Fig.~\ref{fig:all-perturb-taxonomy}) in SLAM systems, using medium-severity static perturbations, as showcased in Table~\ref{tab:mixed_perturbation}.
Our key observations are: \textbf{1}) {{Certain combinations of mixed perturbations can degrade performance more than individual perturbations alone}}, as seen in the example of \textit{Snow Effect} and \textit{Motion Blur} leading to higher trajectory estimation error for the GO-SLAM model.
\textbf{2}) However, {{mixing multiple perturbations does not always worsen performance}}, as in the case of \textit{JPEG Compression} slightly reducing trajectory estimation error for GO-SLAM.
The overall performance under mixed perturbations depends on the complex interplay of various perturbations.

\section{Conclusion and Future Work}
\label{sec:conclusions}
\noindent\textbf{Conclusion.} In this work, we first presented a comprehensive taxonomy of perturbations for embodied RGB-D sensing system and introduced a versatile noisy data synthesis pipeline, which can be utilized to transform perturbation-free scenes, \textit{i.e.}, \textit{Perfect World}, into customizable perturbed datasets, \textit{i.e.}, \textit{Noisy World}, laying the ground work for rigorous robustness benchmarking. Then, we created the \textit{Noisy-Replica} benchmark -- an extensive initiative designed to assess the resilience of  {RGB-D} SLAM models against a wide range of perturbations. Our evaluation has revealed vulnerabilities in current SLAM systems when exposed to various perturbations. These findings not only highlight the limitations of existing models and their potential failures in real-world, unstructured environments, but also offer valuable insights for future research aimed at developing robust embodied agents.

\noindent\textbf{\textcolor{black}{Limitations and future work.}}
While this work provides a preliminary robustness analysis of SLAM for embodied RGB-D system, revealing the fragility, our work has several limitations that future research could address. \textbf{1}) For perturbation synthesis, leveraging generative models like \cite{ramesh2021zero} could enhance the quality and realism of testing environments. Additionally, exploring more types of perturbations, such as regional sensor corruptions \cite{moseley2021extreme} and adversarial perturbations \cite{moosavi2017universal}, would be valuable. \textbf{2}) For robustness evaluation, future studies could investigate the complex interplay between multiple perturbations and evaluate model robustness under more diverse coupled noises. Further research could also assess the robustness of more diverse SLAM models, including those using voxel representation \cite{reijgwart2019voxgraph,wang2021elastic}, and additional modalities (\textit{e.g.}, IMU \cite{rosinol2021kimera} and LiDAR \cite{zhang2014loam}). Exploring the robustness of active SLAM \cite{actiev-slam} and multi-agent SLAM \cite{Kimera-Multi} represents an exciting frontier. \textbf{3}) Beyond robustness benchmarking, future work could explore model robustness enhancement, with techniques like sensor correction \cite{nerfwater} and calibration~\cite{basso2018robust,10387679}. Additional future directions for exploration are provided in Appendix Sec.~\ref{sec:more future work}.

\begin{ack}
This work is partially supported by Office of Naval Research (Grant \#: N00014-24-1-2137; Program Manager: Michael “Q” Qin). The authors appreciate partial GPU resources provided by Prof. Ram Vasudevan from UMich Robotics. The authors are grateful to Dr. Lu Li from CMU, Prof. Siheng Chen from SJTU, Dr. Wenshan Wang from CMU, and Dr. Youmin Zhang from University of Bologna for valuable discussions. The authors express their gratitude to Prof. Sara Nezami Nav, Prof. Pamela Bogart,   Mrs. Lucy Kates, and Mr. Todd Maslyk from UMich ELI for proofreading.
\end{ack}

\bibliographystyle{unsrtnat}
{
\small
\bibliography{main}

\begin{thebibliography}{85}
\providecommand{\natexlab}[1]{#1}
\providecommand{\url}[1]{\texttt{#1}}
\expandafter\ifx\csname urlstyle\endcsname\relax
  \providecommand{\doi}[1]{doi: #1}\else
  \providecommand{\doi}{doi: \begingroup \urlstyle{rm}\Url}\fi

\bibitem[Kaufmann et~al.(2023)Kaufmann, Bauersfeld, Loquercio, M{\"u}ller, Koltun, and Scaramuzza]{kaufmann2023champion}
Elia Kaufmann, Leonard Bauersfeld, Antonio Loquercio, Matthias M{\"u}ller, Vladlen Koltun, and Davide Scaramuzza.
\newblock Champion-level drone racing using deep reinforcement learning.
\newblock \emph{Nature}, 620\penalty0 (7976):\penalty0 982--987, 2023.

\bibitem[Ebadi et~al.(2023)Ebadi, Bernreiter, Biggie, Catt, Chang, Chatterjee, Denniston, Deschênes, Harlow, Khattak, Nogueira, Palieri, Petráček, Petrlík, Reinke, Krátký, Zhao, Agha-mohammadi, Alexis, Heckman, Khosoussi, Kottege, Morrell, Hutter, Pauling, Pomerleau, Saska, Scherer, Siegwart, Williams, and Carlone]{SubT}
Kamak Ebadi, Lukas Bernreiter, Harel Biggie, Gavin Catt, Yun Chang, Arghya Chatterjee, Christopher~E. Denniston, Simon-Pierre Deschênes, Kyle Harlow, Shehryar Khattak, Lucas Nogueira, Matteo Palieri, Pavel Petráček, Matěj Petrlík, Andrzej Reinke, Vít Krátký, Shibo Zhao, Ali-akbar Agha-mohammadi, Kostas Alexis, Christoffer Heckman, Kasra Khosoussi, Navinda Kottege, Benjamin Morrell, Marco Hutter, Fred Pauling, François Pomerleau, Martin Saska, Sebastian Scherer, Roland Siegwart, Jason~L. Williams, and Luca Carlone.
\newblock Present and future of slam in extreme environments: The darpa subt challenge.
\newblock \emph{IEEE Transactions on Robotics}, pages 1--20, 2023.

\bibitem[Chattopadhyay et~al.(2021)Chattopadhyay, Hoffman, Mottaghi, and Kembhavi]{RobustNav}
Prithvijit Chattopadhyay, Judy Hoffman, Roozbeh Mottaghi, and Aniruddha Kembhavi.
\newblock Robustnav: Towards benchmarking robustness in embodied navigation.
\newblock In \emph{IEEE/CVF International Conference on Computer Vision}, pages 15691--15700, 2021.

\bibitem[Cadena et~al.(2016)Cadena, Carlone, Carrillo, Latif, Scaramuzza, Neira, Reid, and Leonard]{slam-survey1}
Cesar Cadena, Luca Carlone, Henry Carrillo, Yasir Latif, Davide Scaramuzza, José Neira, Ian Reid, and John~J. Leonard.
\newblock Past, present, and future of simultaneous localization and mapping: Toward the robust-perception age.
\newblock \emph{IEEE Transactions on Robotics}, 32\penalty0 (6):\penalty0 1309--1332, 2016.

\bibitem[Chen et~al.(2023)Chen, Wang, Lu, Trigoni, and Markham]{TNNLS-SLAM-survey}
Changhao Chen, Bing Wang, Chris~Xiaoxuan Lu, Niki Trigoni, and Andrew Markham.
\newblock Deep learning for visual localization and mapping: A survey.
\newblock \emph{IEEE Transactions on Neural Networks and Learning Systems}, pages 1--21, 2023.

\bibitem[Maddern et~al.(2017)Maddern, Pascoe, Linegar, and Newman]{oxford-robotcar}
Will Maddern, Geoffrey Pascoe, Chris Linegar, and Paul Newman.
\newblock 1 year, 1000 km: The oxford robotcar dataset.
\newblock \emph{The International Journal of Robotics Research}, 36\penalty0 (1):\penalty0 3--15, 2017.

\bibitem[Choi et~al.(2018)Choi, Kim, Hwang, Park, Yoon, An, and Kweon]{Multi-Spectral}
Yukyung Choi, Namil Kim, Soonmin Hwang, Kibaek Park, Jae~Shin Yoon, Kyounghwan An, and In~So Kweon.
\newblock Kaist multi-spectral day/night data set for autonomous and assisted driving.
\newblock \emph{IEEE Transactions on Intelligent Transportation Systems}, 19\penalty0 (3):\penalty0 934--948, 2018.

\bibitem[Burri et~al.(2016)Burri, Nikolic, Gohl, Schneider, Rehder, Omari, Achtelik, and Siegwart]{Burri25012016}
Michael Burri, Janosch Nikolic, Pascal Gohl, Thomas Schneider, Joern Rehder, Sammy Omari, Markus~W Achtelik, and Roland Siegwart.
\newblock The {EuRoC} micro aerial vehicle datasets.
\newblock \emph{The International Journal of Robotics Research}, 2016.

\bibitem[Geiger et~al.(2012)Geiger, Lenz, and Urtasun]{Geiger2012CVPR}
Andreas Geiger, Philip Lenz, and Raquel Urtasun.
\newblock Are we ready for autonomous driving? {T}he {KITTI} {V}ision {B}enchmark {S}uite.
\newblock In \emph{Conference on Computer Vision and Pattern Recognition (CVPR)}, 2012.

\bibitem[Zhang et~al.(2021)Zhang, Camurri, and Fallon]{zhang2021multi}
Lintong Zhang, Marco Camurri, and Maurice Fallon.
\newblock Multi-camera lidar inertial extension to the newer college dataset.
\newblock \emph{arXiv preprint arXiv:2112.08854}, 2021.

\bibitem[Pfrommer et~al.(2017)Pfrommer, Sanket, Daniilidis, and Cleveland]{pfrommer2017penncosyvio}
Bernd Pfrommer, Nitin Sanket, Kostas Daniilidis, and Jonas Cleveland.
\newblock Penncosyvio: A challenging visual inertial odometry benchmark.
\newblock In \emph{2017 IEEE International Conference on Robotics and Automation (ICRA)}, pages 3847--3854. IEEE, 2017.

\bibitem[Sturm et~al.(2012)Sturm, Engelhard, Endres, Burgard, and Cremers]{TUM-RGBD}
Jürgen Sturm, Nikolas Engelhard, Felix Endres, Wolfram Burgard, and Daniel Cremers.
\newblock A benchmark for the evaluation of rgb-d slam systems.
\newblock In \emph{2012 IEEE/RSJ International Conference on Intelligent Robots and Systems}, pages 573--580, 2012.

\bibitem[Zu{\~n}iga-No{\"e}l et~al.(2020)Zu{\~n}iga-No{\"e}l, Jaenal, Gomez-Ojeda, and Gonzalez-Jimenez]{zuniga2020vi}
David Zu{\~n}iga-No{\"e}l, Alberto Jaenal, Ruben Gomez-Ojeda, and Javier Gonzalez-Jimenez.
\newblock The uma-vi dataset: Visual--inertial odometry in low-textured and dynamic illumination environments.
\newblock \emph{The International Journal of Robotics Research}, 39\penalty0 (9):\penalty0 1052--1060, 2020.

\bibitem[Carlevaris-Bianco et~al.(2016)Carlevaris-Bianco, Ushani, and Eustice]{carlevaris2016university}
Nicholas Carlevaris-Bianco, Arash~K Ushani, and Ryan~M Eustice.
\newblock University of michigan north campus long-term vision and lidar dataset.
\newblock \emph{The International Journal of Robotics Research}, 35\penalty0 (9):\penalty0 1023--1035, 2016.

\bibitem[Delmerico et~al.(2019)Delmerico, Cieslewski, Rebecq, Faessler, and Scaramuzza]{Delmerico19icra}
Jeffrey Delmerico, Titus Cieslewski, Henri Rebecq, Matthias Faessler, and Davide Scaramuzza.
\newblock Are we ready for autonomous drone racing? the {UZH-FPV} drone racing dataset.
\newblock In \emph{{IEEE} Int. Conf. Robot. Autom. ({ICRA})}, 2019.

\bibitem[Helmberger et~al.(2022)Helmberger, Morin, Berner, Kumar, Cioffi, and Scaramuzza]{helmberger2022hilti}
Michael Helmberger, Kristian Morin, Beda Berner, Nitish Kumar, Giovanni Cioffi, and Davide Scaramuzza.
\newblock The hilti slam challenge dataset.
\newblock \emph{IEEE Robotics and Automation Letters}, 7\penalty0 (3):\penalty0 7518--7525, 2022.

\bibitem[Tian et~al.(2023)Tian, Chang, Quang, Schang, Nieto-Granda, How, and Carlone]{tian2023resilient}
Yulun Tian, Yun Chang, Long Quang, Arthur Schang, Carlos Nieto-Granda, Jonathan~P How, and Luca Carlone.
\newblock Resilient and distributed multi-robot visual slam: Datasets, experiments, and lessons learned.
\newblock \emph{arXiv preprint arXiv:2304.04362}, 2023.

\bibitem[Schubert et~al.(2018)Schubert, Goll, Demmel, Usenko, St{\"u}ckler, and Cremers]{schubert2018tum}
David Schubert, Thore Goll, Nikolaus Demmel, Vladyslav Usenko, J{\"o}rg St{\"u}ckler, and Daniel Cremers.
\newblock The tum vi benchmark for evaluating visual-inertial odometry.
\newblock In \emph{2018 IEEE/RSJ International Conference on Intelligent Robots and Systems}, pages 1680--1687, 2018.

\bibitem[Zhao et~al.(2024)Zhao, Gao, Wu, Singh, Jiang, Sun, Sarawata, Whittaker, Higgins, Su, Du, Xu, Keller, Karhade, Nogueira, Saha, Qiu, Zhang, Wang, Wang, and Scherer]{zhao2024subt}
Shibo Zhao, Yuanjun Gao, Tianhao Wu, Damanpreet Singh, Rushan Jiang, Haoxiang Sun, Mansi Sarawata, Warren~C Whittaker, Ian Higgins, Shaoshu Su, Yi~Du, Can Xu, John Keller, Jay Karhade, Lucas Nogueira, Sourojit Saha, Yuheng Qiu, Ji~Zhang, Wenshan Wang, Chen Wang, and Sebastian Scherer.
\newblock {SubT-MRS} dataset: Pushing slam towards all-weather environments.
\newblock In \emph{IEEE/CVF Conference on Computer Vision and Pattern Recognition (CVPR)}, 2024.

\bibitem[Sayre-McCord et~al.(2018)Sayre-McCord, Guerra, Antonini, Arneberg, Brown, Cavalheiro, Fang, Gorodetsky, McCoy, Quilter, Riether, Tal, Terzioglu, Carlone, and Karaman]{VI-navigation-with-simulation}
Thomas Sayre-McCord, Winter Guerra, Amado Antonini, Jasper Arneberg, Austin Brown, Guilherme Cavalheiro, Yajun Fang, Alex Gorodetsky, Dave McCoy, Sebastian Quilter, Fabian Riether, Ezra Tal, Yunus Terzioglu, Luca Carlone, and Sertac Karaman.
\newblock Visual-inertial navigation algorithm development using photorealistic camera simulation in the loop.
\newblock In \emph{2018 IEEE International Conference on Robotics and Automation (ICRA)}, pages 2566--2573, 2018.

\bibitem[Dosovitskiy et~al.(2017)Dosovitskiy, Ros, Codevilla, Lopez, and Koltun]{dosovitskiy2017carla}
Alexey Dosovitskiy, German Ros, Felipe Codevilla, Antonio Lopez, and Vladlen Koltun.
\newblock Carla: An open urban driving simulator.
\newblock In \emph{Conference on robot learning}, pages 1--16. PMLR, 2017.

\bibitem[Wang et~al.(2021)Wang, Hu, and Scherer]{tartanvo}
Wenshan Wang, Yaoyu Hu, and Sebastian Scherer.
\newblock Tartanvo: A generalizable learning-based vo.
\newblock In \emph{Conference on Robot Learning}, pages 1761--1772. PMLR, 2021.

\bibitem[Johnson-Roberson et~al.(2017)Johnson-Roberson, Barto, Mehta, Sridhar, Rosaen, and Vasudevan]{driving-matrix}
Matthew Johnson-Roberson, Charles Barto, Rounak Mehta, Sharath~Nittur Sridhar, Karl Rosaen, and Ram Vasudevan.
\newblock Driving in the matrix: Can virtual worlds replace human-generated annotations for real world tasks?
\newblock In \emph{2017 IEEE International Conference on Robotics and Automation (ICRA)}, pages 746--753, 2017.

\bibitem[Rombach et~al.(2022)Rombach, Blattmann, Lorenz, Esser, and Ommer]{rombach2022high}
Robin Rombach, Andreas Blattmann, Dominik Lorenz, Patrick Esser, and Bj{\"o}rn Ommer.
\newblock High-resolution image synthesis with latent diffusion models.
\newblock In \emph{Proceedings of the IEEE/CVF conference on computer vision and pattern recognition}, pages 10684--10695, 2022.

\bibitem[Raistrick et~al.(2023)Raistrick, Lipson, Ma, Mei, Wang, Zuo, Kayan, Wen, Han, Wang, et~al.]{raistrick2023infinite}
Alexander Raistrick, Lahav Lipson, Zeyu Ma, Lingjie Mei, Mingzhe Wang, Yiming Zuo, Karhan Kayan, Hongyu Wen, Beining Han, Yihan Wang, et~al.
\newblock Infinite photorealistic worlds using procedural generation.
\newblock In \emph{Proceedings of the IEEE/CVF Conference on Computer Vision and Pattern Recognition}, pages 12630--12641, 2023.

\bibitem[Straub et~al.(2019)Straub, Whelan, Ma, Chen, Wijmans, Green, Engel, Mur-Artal, Ren, Verma, et~al.]{straub2019replica}
Julian Straub, Thomas Whelan, Lingni Ma, Yufan Chen, Erik Wijmans, Simon Green, Jakob~J Engel, Raul Mur-Artal, Carl Ren, Shobhit Verma, et~al.
\newblock The replica dataset: A digital replica of indoor spaces.
\newblock \emph{arXiv preprint arXiv:1906.05797}, 2019.

\bibitem[Wang et~al.(2020)Wang, Zhu, Wang, Hu, Qiu, Wang, Hu, Kapoor, and Scherer]{wang2020tartanair}
Wenshan Wang, Delong Zhu, Xiangwei Wang, Yaoyu Hu, Yuheng Qiu, Chen Wang, Yafei Hu, Ashish Kapoor, and Sebastian Scherer.
\newblock Tartanair: A dataset to push the limits of visual slam.
\newblock In \emph{2020 IEEE/RSJ International Conference on Intelligent Robots and Systems (IROS)}, pages 4909--4916. IEEE, 2020.

\bibitem[Dai et~al.(2017)Dai, Chang, Savva, Halber, Funkhouser, and Nie{\ss}ner]{dai2017scannet}
Angela Dai, Angel~X Chang, Manolis Savva, Maciej Halber, Thomas Funkhouser, and Matthias Nie{\ss}ner.
\newblock Scannet: Richly-annotated 3d reconstructions of indoor scenes.
\newblock In \emph{IEEE/CVF Conference on Computer Vision and Pattern Recognition}, pages 5828--5839, 2017.

\bibitem[Deitke et~al.(2022)Deitke, VanderBilt, Herrasti, Weihs, Ehsani, Salvador, Han, Kolve, Kembhavi, and Mottaghi]{deitke2022️}
Matt Deitke, Eli VanderBilt, Alvaro Herrasti, Luca Weihs, Kiana Ehsani, Jordi Salvador, Winson Han, Eric Kolve, Aniruddha Kembhavi, and Roozbeh Mottaghi.
\newblock Procthor: Large-scale embodied ai using procedural generation.
\newblock \emph{Advances in Neural Information Processing Systems}, 35:\penalty0 5982--5994, 2022.

\bibitem[Zheng et~al.(2020)Zheng, Zhang, Li, Tang, Gao, and Zhou]{zheng2020structured3d}
Jia Zheng, Junfei Zhang, Jing Li, Rui Tang, Shenghua Gao, and Zihan Zhou.
\newblock Structured3d: A large photo-realistic dataset for structured 3d modeling.
\newblock In \emph{Computer Vision--ECCV 2020: 16th European Conference, Glasgow, UK, August 23--28, 2020, Proceedings, Part IX 16}, pages 519--535. Springer, 2020.

\bibitem[Denninger et~al.(2020)Denninger, Sundermeyer, Winkelbauer, Olefir, Hodan, Zidan, Elbadrawy, Knauer, Katam, and Lodhi]{denninger2020blenderproc}
Maximilian Denninger, Martin Sundermeyer, Dominik Winkelbauer, Dmitry Olefir, Tomas Hodan, Youssef Zidan, Mohamad Elbadrawy, Markus Knauer, Harinandan Katam, and Ahsan Lodhi.
\newblock Blenderproc: Reducing the reality gap with photorealistic rendering.
\newblock In \emph{International Conference on Robotics: Sciene and Systems, RSS 2020}, 2020.

\bibitem[Rosinol et~al.(2022)Rosinol, Leonard, and Carlone]{rosinol2022nerf}
Antoni Rosinol, John~J Leonard, and Luca Carlone.
\newblock Nerf-slam: Real-time dense monocular slam with neural radiance fields.
\newblock \emph{arXiv preprint arXiv:2210.13641}, 2022.

\bibitem[Kerbl et~al.(2023)Kerbl, Kopanas, Leimk{\"u}hler, and Drettakis]{gaussiansplatting}
Bernhard Kerbl, Georgios Kopanas, Thomas Leimk{\"u}hler, and George Drettakis.
\newblock 3d gaussian splatting for real-time radiance field rendering.
\newblock \emph{ACM Transactions on Graphics}, 42\penalty0 (4), 2023.

\bibitem[Zhang et~al.(2017)Zhang, Zhang, and Gupta]{zhang2017resilient}
Tan Zhang, Wenjun Zhang, and Madan~M Gupta.
\newblock Resilient robots: Concept, review, and future directions.
\newblock \emph{Robotics}, 6\penalty0 (4):\penalty0 22, 2017.

\bibitem[Taori et~al.(2020)Taori, Dave, Shankar, Carlini, Recht, and Schmidt]{taori2020measuring}
Rohan Taori, Achal Dave, Vaishaal Shankar, Nicholas Carlini, Benjamin Recht, and Ludwig Schmidt.
\newblock Measuring robustness to natural distribution shifts in image classification.
\newblock \emph{Advances in Neural Information Processing Systems}, 33:\penalty0 18583--18599, 2020.

\bibitem[Hendrycks and Dietterich(2019)]{hendrycks2019robustness}
Dan Hendrycks and Thomas Dietterich.
\newblock Benchmarking neural network robustness to common corruptions and perturbations.
\newblock \emph{Proceedings of the International Conference on Learning Representations}, 2019.

\bibitem[Michaelis et~al.(2019)Michaelis, Mitzkus, Geirhos, Rusak, Bringmann, Ecker, Bethge, and Brendel]{michaelis2019benchmarking}
Claudio Michaelis, Benjamin Mitzkus, Robert Geirhos, Evgenia Rusak, Oliver Bringmann, Alexander~S Ecker, Matthias Bethge, and Wieland Brendel.
\newblock Benchmarking robustness in object detection: Autonomous driving when winter is coming.
\newblock \emph{arXiv preprint arXiv:1907.07484}, 2019.

\bibitem[Carlson et~al.(2018)Carlson, Skinner, Vasudevan, and Johnson-Roberson]{carlson2018modeling}
Alexandra Carlson, Katherine~A. Skinner, Ram Vasudevan, and Matthew Johnson-Roberson.
\newblock Modeling camera effects to improve visual learning from synthetic data.
\newblock In \emph{Proceedings of the European Conference on Computer Vision (ECCV) Workshops}, September 2018.

\bibitem[Kong et~al.(2023{\natexlab{a}})Kong, Xie, Hu, Ng, Cottereau, and Ooi]{kong2023robodepth}
Lingdong Kong, Shaoyuan Xie, Hanjiang Hu, Lai~Xing Ng, Benoit~R Cottereau, and Wei~Tsang Ooi.
\newblock Robodepth: Robust out-of-distribution depth estimation under corruptions.
\newblock In \emph{Thirty-seventh Conference on Neural Information Processing Systems Datasets and Benchmarks Track}, 2023{\natexlab{a}}.
\newblock URL \url{https://openreview.net/forum?id=SNznC08OOO}.

\bibitem[Kong et~al.(2023{\natexlab{b}})Kong, Liu, Li, Chen, Zhang, Ren, Pan, Chen, and Liu]{kong2023robo3d}
Lingdong Kong, Youquan Liu, Xin Li, Runnan Chen, Wenwei Zhang, Jiawei Ren, Liang Pan, Kai Chen, and Ziwei Liu.
\newblock Robo3d: Towards robust and reliable 3d perception against corruptions.
\newblock In \emph{Proceedings of the IEEE/CVF International Conference on Computer Vision}, pages 19994--20006, 2023{\natexlab{b}}.

\bibitem[Kamann and Rother(2020{\natexlab{a}})]{kamann2020benchmarking}
Christoph Kamann and Carsten Rother.
\newblock Benchmarking the robustness of semantic segmentation models.
\newblock In \emph{Proceedings of the IEEE/CVF Conference on Computer Vision and Pattern Recognition}, pages 8828--8838, 2020{\natexlab{a}}.

\bibitem[Xu et~al.(2022)Xu, Wang, Ming, and Lu]{xu2022towards}
Xiaohao Xu, Jinglu Wang, Xiang Ming, and Yan Lu.
\newblock Towards robust video object segmentation with adaptive object calibration.
\newblock In \emph{Proceedings of the 30th ACM International Conference on Multimedia}, pages 2709--2718, 2022.

\bibitem[Li et~al.(2023)Li, Wang, Xu, Li, Raj, and Lu]{li2023robust}
Xiang Li, Jinglu Wang, Xiaohao Xu, Xiao Li, Bhiksha Raj, and Yan Lu.
\newblock Robust referring video object segmentation with cyclic structural consensus.
\newblock In \emph{Proceedings of the IEEE/CVF International Conference on Computer Vision}, pages 22236--22245, 2023.

\bibitem[Yokoyama et~al.(2022)Yokoyama, Luo, Batra, and Ha]{yokoyama2022benchmarking}
Naoki Yokoyama, Qian Luo, Dhruv Batra, and Sehoon Ha.
\newblock Benchmarking augmentation methods for learning robust navigation agents: the winning entry of the 2021 igibson challenge.
\newblock In \emph{2022 IEEE/RSJ International Conference on Intelligent Robots and Systems (IROS)}, pages 1748--1755. IEEE, 2022.

\bibitem[Macario~Barros et~al.(2022)Macario~Barros, Michel, Moline, Corre, and Carrel]{macario2022comprehensive}
Andr{\'e}a Macario~Barros, Maugan Michel, Yoann Moline, Gwenol{\'e} Corre, and Fr{\'e}d{\'e}rick Carrel.
\newblock A comprehensive survey of visual slam algorithms.
\newblock \emph{Robotics}, 11\penalty0 (1):\penalty0 24, 2022.

\bibitem[Kazerouni et~al.(2022)Kazerouni, Fitzgerald, Dooly, and Toal]{kazerouni2022survey}
Iman~Abaspur Kazerouni, Luke Fitzgerald, Gerard Dooly, and Daniel Toal.
\newblock A survey of state-of-the-art on visual slam.
\newblock \emph{Expert Systems with Applications}, 205:\penalty0 117734, 2022.

\bibitem[Mur-Artal et~al.(2015)Mur-Artal, Montiel, and Tardos]{orbslam}
Raul Mur-Artal, Jose Maria~Martinez Montiel, and Juan~D Tardos.
\newblock Orb-slam: a versatile and accurate monocular slam system.
\newblock \emph{IEEE transactions on robotics}, 31\penalty0 (5):\penalty0 1147--1163, 2015.

\bibitem[Mur-Artal and Tardós(2017)]{orbslam2}
Raúl Mur-Artal and Juan~D. Tardós.
\newblock Orb-slam2: An open-source slam system for monocular, stereo, and rgb-d cameras.
\newblock \emph{IEEE Transactions on Robotics}, 33\penalty0 (5):\penalty0 1255--1262, 2017.

\bibitem[Campos et~al.(2021)Campos, Elvira, Rodríguez, M.~Montiel, and D.~Tardós]{orbslam3}
Carlos Campos, Richard Elvira, Juan J.~Gómez Rodríguez, José~M. M.~Montiel, and Juan D.~Tardós.
\newblock Orb-slam3: An accurate open-source library for visual, visual–inertial, and multimap slam.
\newblock \emph{IEEE Transactions on Robotics}, 37\penalty0 (6):\penalty0 1874--1890, 2021.

\bibitem[Khedekar et~al.(2022)Khedekar, Kulkarni, and Alexis]{MIMOSA}
Nikhil Khedekar, Mihir Kulkarni, and Kostas Alexis.
\newblock Mimosa: A multi-modal slam framework for resilient autonomy against sensor degradation.
\newblock In \emph{2022 IEEE/RSJ International Conference on Intelligent Robots and Systems (IROS)}, pages 7153--7159, 2022.

\bibitem[Rosinol et~al.(2021)Rosinol, Violette, Abate, Hughes, Chang, Shi, Gupta, and Carlone]{rosinol2021kimera}
Antoni Rosinol, Andrew Violette, Marcus Abate, Nathan Hughes, Yun Chang, Jingnan Shi, Arjun Gupta, and Luca Carlone.
\newblock Kimera: From slam to spatial perception with 3d dynamic scene graphs.
\newblock \emph{The International Journal of Robotics Research}, 40\penalty0 (12-14):\penalty0 1510--1546, 2021.

\bibitem[Teed and Deng(2021)]{teed2021droid}
Zachary Teed and Jia Deng.
\newblock Droid-slam: Deep visual slam for monocular, stereo, and rgb-d cameras.
\newblock \emph{Advances in neural information processing systems}, 34:\penalty0 16558--16569, 2021.

\bibitem[Wang et~al.(2023)Wang, Wang, and Agapito]{coslam}
Hengyi Wang, Jingwen Wang, and Lourdes Agapito.
\newblock Co-slam: Joint coordinate and sparse parametric encodings for neural real-time slam.
\newblock In \emph{Proceedings of the IEEE/CVF Conference on Computer Vision and Pattern Recognition}, pages 13293--13302, 2023.

\bibitem[Sucar et~al.(2021)Sucar, Liu, Ortiz, and Davison]{imap}
Edgar Sucar, Shikun Liu, Joseph Ortiz, and Andrew~J Davison.
\newblock imap: Implicit mapping and positioning in real-time.
\newblock In \emph{Proceedings of the IEEE/CVF International Conference on Computer Vision}, pages 6229--6238, 2021.

\bibitem[Zhu et~al.(2022)Zhu, Peng, Larsson, Xu, Bao, Cui, Oswald, and Pollefeys]{niceslam}
Zihan Zhu, Songyou Peng, Viktor Larsson, Weiwei Xu, Hujun Bao, Zhaopeng Cui, Martin~R Oswald, and Marc Pollefeys.
\newblock Nice-slam: Neural implicit scalable encoding for slam.
\newblock In \emph{Proceedings of the IEEE/CVF Conference on Computer Vision and Pattern Recognition}, pages 12786--12796, 2022.

\bibitem[Zhang et~al.(2023)Zhang, Tosi, Mattoccia, and Poggi]{zhang2023goslam}
Youmin Zhang, Fabio Tosi, Stefano Mattoccia, and Matteo Poggi.
\newblock Go-slam: Global optimization for consistent 3d instant reconstruction.
\newblock In \emph{Proceedings of the IEEE/CVF International Conference on Computer Vision (ICCV)}, October 2023.

\bibitem[Zhao et~al.(2023)Zhao, Singh, Sun, Jiang, Gao, Wu, Karhade, Whittaker, Higgins, Xu, et~al.]{zhao2023subt}
Shibo Zhao, Damanpreet Singh, Haoxiang Sun, Rushan Jiang, YuanJun Gao, Tianhao Wu, Jay Karhade, Chuck Whittaker, Ian Higgins, Jiahe Xu, et~al.
\newblock Subt-mrs: A subterranean, multi-robot, multi-spectral and multi-degraded dataset for robust slam.
\newblock \emph{arXiv preprint arXiv:2307.07607}, 2023.

\bibitem[Bujanca et~al.(2021)Bujanca, Shi, Spear, Zhao, Lennox, and Luj{\'a}n]{bujanca2021robust}
Mihai Bujanca, Xuesong Shi, Matthew Spear, Pengpeng Zhao, Barry Lennox, and Mikel Luj{\'a}n.
\newblock Robust slam systems: Are we there yet?
\newblock In \emph{2021 IEEE/RSJ International Conference on Intelligent Robots and Systems (IROS)}, pages 5320--5327. IEEE, 2021.

\bibitem[Thrun(2002)]{probabilistic_robot}
Sebastian Thrun.
\newblock Probabilistic robotics.
\newblock \emph{Commun. ACM}, 45\penalty0 (3):\penalty0 52–57, 2002.

\bibitem[Todorov et~al.(2012)Todorov, Erez, and Tassa]{mujoco}
Emanuel Todorov, Tom Erez, and Yuval Tassa.
\newblock Mujoco: A physics engine for model-based control.
\newblock In \emph{2012 IEEE/RSJ International Conference on Intelligent Robots and Systems}, pages 5026--5033, 2012.

\bibitem[Shreiner et~al.(2009)]{shreiner2009opengl}
Dave Shreiner et~al.
\newblock \emph{OpenGL programming guide: the official guide to learning OpenGL, versions 3.0 and 3.1}.
\newblock Pearson Education, 2009.

\bibitem[Keetha et~al.(2023)Keetha, Karhade, Jatavallabhula, Yang, Scherer, Ramanan, and Luiten]{keetha2023splatam}
Nikhil Keetha, Jay Karhade, Krishna~Murthy Jatavallabhula, Gengshan Yang, Sebastian Scherer, Deva Ramanan, and Jonathan Luiten.
\newblock Splatam: Splat, track map 3d gaussians for dense rgb-d slam.
\newblock \emph{arXiv}, 2023.

\bibitem[Prokhorov et~al.(2019)Prokhorov, Zhukov, Barinova, Anton, and Vorontsova]{prokhorov2019measuring}
David Prokhorov, Dmitry Zhukov, Olga Barinova, Konushin Anton, and Anna Vorontsova.
\newblock Measuring robustness of visual slam.
\newblock In \emph{2019 16th International conference on machine vision applications (MVA)}, pages 1--6. IEEE, 2019.

\bibitem[Ramesh et~al.(2021)Ramesh, Pavlov, Goh, Gray, Voss, Radford, Chen, and Sutskever]{ramesh2021zero}
Aditya Ramesh, Mikhail Pavlov, Gabriel Goh, Scott Gray, Chelsea Voss, Alec Radford, Mark Chen, and Ilya Sutskever.
\newblock Zero-shot text-to-image generation.
\newblock In \emph{International Conference on Machine Learning}, pages 8821--8831. PMLR, 2021.

\bibitem[Moseley et~al.(2021)Moseley, Bickel, L{\'o}pez-Francos, and Rana]{moseley2021extreme}
Ben Moseley, Valentin Bickel, Ignacio~G L{\'o}pez-Francos, and Loveneesh Rana.
\newblock Extreme low-light environment-driven image denoising over permanently shadowed lunar regions with a physical noise model.
\newblock In \emph{Proceedings of the IEEE/CVF Conference on Computer Vision and Pattern Recognition}, pages 6317--6327, 2021.

\bibitem[Moosavi-Dezfooli et~al.(2017)Moosavi-Dezfooli, Fawzi, Fawzi, and Frossard]{moosavi2017universal}
Seyed-Mohsen Moosavi-Dezfooli, Alhussein Fawzi, Omar Fawzi, and Pascal Frossard.
\newblock Universal adversarial perturbations.
\newblock In \emph{Proceedings of the IEEE conference on computer vision and pattern recognition}, pages 1765--1773, 2017.

\bibitem[{Reijgwart, Victor and Millane, Alexander and Oleynikova, Helen and Siegwart, Roland and Cadena, Cesar and Nieto, Juan}({2019})]{reijgwart2019voxgraph}
{Reijgwart, Victor and Millane, Alexander and Oleynikova, Helen and Siegwart, Roland and Cadena, Cesar and Nieto, Juan}.
\newblock {Voxgraph: Globally consistent, volumetric mapping using signed distance function submaps}.
\newblock \emph{{IEEE Robotics and Automation Letters}}, {5}\penalty0 ({1}):\penalty0 {227--234}, {2019}.

\bibitem[{Wang, Yiduo and Funk, Nils and Ramezani, Milad and Papatheodorou, Sotiris and Popovi{\'c}, Marija and Camurri, Marco and Leutenegger, Stefan and Fallon, Maurice}({2021})]{wang2021elastic}
{Wang, Yiduo and Funk, Nils and Ramezani, Milad and Papatheodorou, Sotiris and Popovi{\'c}, Marija and Camurri, Marco and Leutenegger, Stefan and Fallon, Maurice}.
\newblock {Elastic and efficient LiDAR reconstruction for large-scale exploration tasks}.
\newblock In \emph{{2021 IEEE International Conference on Robotics and Automation (ICRA)}}, pages {5035--5041}. {IEEE}, {2021}.

\bibitem[Zhang and Singh(2014)]{zhang2014loam}
Ji~Zhang and Sanjiv Singh.
\newblock {LOAM}: Lidar odometry and mapping in real-time.
\newblock \emph{Robotics: Science and Systems Conference (RSS)}, pages 109--111, 01 2014.

\bibitem[Placed et~al.(2023)Placed, Strader, Carrillo, Atanasov, Indelman, Carlone, and Castellanos]{actiev-slam}
Julio~A. Placed, Jared Strader, Henry Carrillo, Nikolay Atanasov, Vadim Indelman, Luca Carlone, and José~A. Castellanos.
\newblock A survey on active simultaneous localization and mapping: State of the art and new frontiers.
\newblock \emph{IEEE Transactions on Robotics}, 39\penalty0 (3):\penalty0 1686--1705, 2023.

\bibitem[Tian et~al.(2022)Tian, Chang, Herrera~Arias, Nieto-Granda, How, and Carlone]{Kimera-Multi}
Yulun Tian, Yun Chang, Fernando Herrera~Arias, Carlos Nieto-Granda, Jonathan~P. How, and Luca Carlone.
\newblock Kimera-multi: Robust, distributed, dense metric-semantic slam for multi-robot systems.
\newblock \emph{IEEE Transactions on Robotics}, 38\penalty0 (4):\penalty0 2022--2038, 2022.

\bibitem[Zhang and Johnson-Roberson(2023)]{nerfwater}
Tianyi Zhang and Matthew Johnson-Roberson.
\newblock Beyond nerf underwater: Learning neural reflectance fields for true color correction of marine imagery.
\newblock \emph{IEEE Robotics and Automation Letters}, 8\penalty0 (10):\penalty0 6467--6474, 2023.

\bibitem[Basso et~al.(2018)Basso, Menegatti, and Pretto]{basso2018robust}
Filippo Basso, Emanuele Menegatti, and Alberto Pretto.
\newblock Robust intrinsic and extrinsic calibration of rgb-d cameras.
\newblock \emph{IEEE Transactions on Robotics}, 34\penalty0 (5):\penalty0 1315--1332, 2018.

\bibitem[Murai et~al.(2024)Murai, Alzugaray, Kelly, and Davison]{10387679}
Riku Murai, Ignacio Alzugaray, Paul~H.J. Kelly, and Andrew~J. Davison.
\newblock Distributed simultaneous localisation and auto-calibration using gaussian belief propagation.
\newblock \emph{IEEE Robotics and Automation Letters}, 9\penalty0 (3):\penalty0 2136--2143, 2024.
\newblock \doi{10.1109/LRA.2024.3352361}.

\bibitem[Gebru et~al.(2021)Gebru, Morgenstern, Vecchione, Vaughan, Wallach, III, and Crawford]{datasheet}
Timnit Gebru, Jamie Morgenstern, Briana Vecchione, Jennifer~Wortman Vaughan, Hanna Wallach, Hal~Daum\'{e} III, and Kate Crawford.
\newblock Datasheets for datasets.
\newblock \emph{Commun. ACM}, 64\penalty0 (12):\penalty0 86–92, nov 2021.
\newblock ISSN 0001-0782.
\newblock \doi{10.1145/3458723}.
\newblock URL \url{https://doi.org/10.1145/3458723}.

\bibitem[Chang et~al.(2015)Chang, Funkhouser, Guibas, Hanrahan, Huang, Li, Savarese, Savva, Song, Su, et~al.]{chang2015shapenet}
Angel~X Chang, Thomas Funkhouser, Leonidas Guibas, Pat Hanrahan, Qixing Huang, Zimo Li, Silvio Savarese, Manolis Savva, Shuran Song, Hao Su, et~al.
\newblock Shapenet: An information-rich 3d model repository.
\newblock \emph{arXiv preprint arXiv:1512.03012}, 2015.

\bibitem[Kamann and Rother(2020{\natexlab{b}})]{kamann2020increasing}
Christoph Kamann and Carsten Rother.
\newblock Increasing the robustness of semantic segmentation models with painting-by-numbers.
\newblock In \emph{European Conference on Computer Vision}, pages 369--387. Springer, 2020{\natexlab{b}}.

\bibitem[Yu et~al.(2019)Yu, Song, Miao, Yang, Yang, and Chen]{yu2019nighttime}
Teng Yu, Kang Song, Pu~Miao, Guowei Yang, Huan Yang, and Chenglizhao Chen.
\newblock Nighttime single image dehazing via pixel-wise alpha blending.
\newblock \emph{IEEE Access}, 7:\penalty0 114619--114630, 2019.

\bibitem[Smith(1995)]{smith1995image}
Alvy~Ray Smith.
\newblock Image compositing fundamentals.
\newblock \emph{Microsoft Corporation}, 5, 1995.

\bibitem[Von~Bernuth et~al.(2019)Von~Bernuth, Volk, and Bringmann]{von2019simulating}
Alexander Von~Bernuth, Georg Volk, and Oliver Bringmann.
\newblock Simulating photo-realistic snow and fog on existing images for enhanced cnn training and evaluation.
\newblock In \emph{2019 IEEE Intelligent Transportation Systems Conference (ITSC)}, pages 41--46. IEEE, 2019.

\bibitem[Hasinoff(2014)]{hasinoff2014photon}
Samuel~W Hasinoff.
\newblock Photon, poisson noise.
\newblock \emph{Computer Vision, A Reference Guide}, 4\penalty0 (16):\penalty0 1, 2014.

\bibitem[Bianco et~al.(2018)Bianco, Memmolo, Leo, Montresor, Distante, Paturzo, Picart, Javidi, and Ferraro]{bianco2018strategies}
Vittorio Bianco, Pasquale Memmolo, Marco Leo, Silvio Montresor, Cosimo Distante, Melania Paturzo, Pascal Picart, Bahram Javidi, and Pietro Ferraro.
\newblock Strategies for reducing speckle noise in digital holography.
\newblock \emph{Light: Science and Applications}, 7\penalty0 (1):\penalty0 48, 2018.

\bibitem[Kim(2015)]{kim2015t}
Tae~Kyun Kim.
\newblock T test as a parametric statistic.
\newblock \emph{Korean journal of anesthesiology}, 68\penalty0 (6):\penalty0 540--546, 2015.

\bibitem[Spearman(1961)]{spearman1961proof}
Charles Spearman.
\newblock The proof and measurement of association between two things.
\newblock 1961.

\bibitem[Cohen et~al.(2009)Cohen, Huang, Chen, Benesty, Benesty, Chen, Huang, and Cohen]{cohen2009pearson}
Israel Cohen, Yiteng Huang, Jingdong Chen, Jacob Benesty, Jacob Benesty, Jingdong Chen, Yiteng Huang, and Israel Cohen.
\newblock Pearson correlation coefficient.
\newblock \emph{Noise reduction in speech processing}, pages 1--4, 2009.

\end{thebibliography}
}


\clearpage

\setcounter{table}{0}  
\setcounter{figure}{0}  
\renewcommand{\thetable}{\Alph{table}}
\renewcommand{\thefigure}{\Alph{figure}}

\appendix

\textbf{Table of Contents of the Appendix}\label{sec:appendix}

\begin{itemize}
    \item \textcolor{black}{\textbf{A. \nameref{sec:datasheet}}} \dotfill \pageref{sec:datasheet}
    \begin{itemize}
        \item \textcolor{black}{\textbf{A.1  \nameref{subsec:datasheet-motivation}}} \dotfill \pageref{subsec:datasheet-motivation}
        \item \textcolor{black}{\textbf{A.2  \nameref{subsec:datasheet-composition}}} \dotfill \pageref{subsec:datasheet-composition}
        \item \textcolor{black}{\textbf{A.3  \nameref{subsec:datasheet-collection}}} \dotfill \pageref{subsec:datasheet-collection}
        \item \textcolor{black}{\textbf{A.4  \nameref{subsec:datasheet-preprocess}}} \dotfill \pageref{subsec:datasheet-preprocess}
        \item \textcolor{black}{\textbf{A.5  \nameref{subsec:datasheet-uses}}} \dotfill \pageref{subsec:datasheet-uses}
        \item \textcolor{black}{\textbf{A.6  \nameref{subsec:datasheet-distribution}}} \dotfill \pageref{subsec:datasheet-distribution}
        \item \textcolor{black}{\textbf{A.7  \nameref{subsec:datasheet-mainteanance}}} \dotfill \pageref{subsec:datasheet-mainteanance}

    \end{itemize}
    \item \textcolor{black}{\textbf{B. \nameref{sec:perturbation-taxonomy-appendix}}} \dotfill \pageref{sec:perturbation-taxonomy-appendix}
    \begin{itemize}
        \item \textcolor{black}{\textbf{B.1  \nameref{subsec:perturbation-taxonomy-appendix-pose}}} \dotfill \pageref{subsec:perturbation-taxonomy-appendix-pose}
        \item \textcolor{black}{\textbf{B.2  \nameref{subsec:perturbation-taxonomy-appendix-rgb}}} \dotfill \pageref{subsec:perturbation-taxonomy-appendix-rgb}
        \item \textcolor{black}{\textbf{B.3  \nameref{subsec:perturbation-taxonomy-appendix-depth}}} \dotfill \pageref{subsec:perturbation-taxonomy-appendix-depth}
        \item \textcolor{black}{\textbf{B.4  \nameref{subsec:perturbation-taxonomy-appendix-rgbd}}} \dotfill \pageref{subsec:perturbation-taxonomy-appendix-rgbd}
    \end{itemize}
    \item \textcolor{black}{\textbf{C. \nameref{sec:benchmark-setup}}} \dotfill \pageref{sec:benchmark-setup}  
    \begin{itemize}
        \item \textcolor{black}{\textbf{C.1  \nameref{subsec:assumption}}} \dotfill \pageref{subsec:assumption}
        \item \textcolor{black}{\textbf{C.2  \nameref{subsec:benchmark-stat}}} \dotfill \pageref{subsec:benchmark-stat}
        \item \textcolor{black}{\textbf{C.3  \nameref{subsec:methods}}} \dotfill \pageref{subsec:methods}
        \item \textcolor{black}{\textbf{C.4  \nameref{subsec:hardware}}} \dotfill \pageref{subsec:hardware}
        \item \textcolor{black}{\textbf{C.5  \nameref{subsec:comparison_slam_benchmarks}}} \dotfill \pageref{subsec:comparison_slam_benchmarks}
    \end{itemize}
    \item \textcolor{black}{\textbf{D. \nameref{sec:additional-results}}} \dotfill \pageref{sec:additional-results}
      \begin{itemize}
        \item \textcolor{black}{\textbf{D.1  \nameref{subsec:additional-results-analyses}}} \dotfill \pageref{subsec:additional-results-analyses}
        \item \textcolor{black}{\textbf{D.2  \nameref{subsec:additional-results-discussion}}} \dotfill \pageref{subsec:additional-results-discussion}
    \end{itemize}
     \item \textcolor{black}{\textbf{E. \nameref{sec:qualitative_results}}} \dotfill \pageref{sec:qualitative_results}
    \begin{itemize}
        \item \textcolor{black}{\textbf{E.1  \nameref{subsec:qualitative_results-SLAM}}} \dotfill \pageref{subsec:qualitative_results-SLAM}
        \item \textcolor{black}{\textbf{E.2  \nameref{subsec:video-demo}}} \dotfill \pageref{subsec:video-demo}
    \end{itemize}
     \item \textcolor{black}{\textbf{F. \nameref{sec:more future work}}} \dotfill \pageref{sec:more future work}
     \item \textcolor{black}{\textbf{G. \nameref{sec:social-impact}}} \dotfill \pageref{sec:social-impact}
     \item \textcolor{black}{\textbf{H. \nameref{sec:availablility-maintenance}}} \dotfill \pageref{sec:availablility-maintenance}
     \item \textcolor{black}{\textbf{I. \nameref{sec:license}}} \dotfill \pageref{sec:license}
     \item \textcolor{black}{\textbf{J. \nameref{sec:public-assets}}} \dotfill \pageref{sec:public-assets}
     \item \textcolor{black}{\textbf{K. \nameref{sec:detailed-result-with-error-bar}}} \dotfill \pageref{sec:detailed-result-with-error-bar}

\end{itemize}

\clearpage
\section{Datasheet}
\label{sec:datasheet}

We document the necessary information about the proposed datasets and benchmarks following the guidelines of Gebru \textit{et al}.~\cite{datasheet}.

\begin{enumerate}[label=Q\arabic*]

\vspace{-0.5em}\begin{figure}[!h]
\subsection{Motivation}\label{subsec:datasheet-motivation}\vspace{-1em}
\end{figure}

\item \textbf{For what purpose was the dataset created?} Was there a specific task in mind? Was there a specific gap that needed to be filled? Please provide a description.

\begin{itemize}
\item Our benchmark was created to holistically evaluate the robustness of RGB-D SLAM models under diverse perturbations. Prior to our work, RGB-D SLAM models were typically evaluated under clean settings. With our customizable perturbation synthesis pipeline, we assess the models across a wide range of RGB-D perturbations crucial for real-world deployment: RGB imaging perturbations, depth imaging perturbations, motion-related perturbations, and RGB-D desynchronization.
\end{itemize}

\item \textbf{Who created the dataset (e.g., which team, research group) and on behalf of which entity (e.g., company, institution, organization)?}

\begin{itemize}
\item This benchmark is presented by researchers from the University of Michigan, Ann Arbor, and Carnegie Mellon University. Our aim is to advance the study, development, and deployment of more reliable and robust autonomous systems.
\end{itemize}

\item \textbf{Who funded the creation of the dataset?} If there is an associated grant, please provide the name of the grantor and the grant name and number.

\begin{itemize}
\item This work was partially supported by the Office of Naval Research (Grant \#: N00014-24-1-2137).
\end{itemize}

\item \textbf{Any other comments?}

\begin{itemize}
\item No.
\end{itemize}

\vspace{-0.5em}\begin{figure}[!h]
\subsection{Composition}\label{subsec:datasheet-composition}\vspace{-1em}
\end{figure}

\item \textbf{What do the instances that comprise the dataset represent (e.g., documents, photos, people, countries)?} \textit{Are there multiple types of instances (e.g., movies, users, and ratings; people and interactions between them; nodes and edges)? Please provide a description.}

\begin{itemize}
\item Our initialized \textit{Noisy-Replica} benchmark includes RGB-D video sequences rendered from scanned 3D scenes, the 6D trajectory at each timestamp, and the 3D scene point cloud.
\end{itemize}

\item \textbf{How many instances are there in total (of each type, if appropriate)?}

\begin{itemize}
\item The \textit{Noisy-Replica} benchmark contains 2000 RGB-D video sequences under various perturbation settings. Detailed statistics for each scenario are available in Sec.~\ref{subsec:benchmark-stat} of the Appendix.
\end{itemize}

\item \textbf{Does the dataset contain all possible instances or is it a sample (not necessarily random) of instances from a larger set?} \textit{If the dataset is a sample, what is the larger set? Is the sample representative of the larger set (e.g., geographic coverage)? If so, please describe how this representativeness was validated/verified. If it is not representative of the larger set, please describe why not (e.g., to cover a more diverse range of instances, because instances were withheld or unavailable).}

\begin{itemize}
\item The 3D scenes in our benchmark are sourced from the existing 3D scanned indoor scene dataset Replica~\cite{straub2019replica}, and we use all possible instances from these datasets.
\end{itemize}

\item \textbf{What data does each instance consist of?} \textit{“Raw” data (e.g., unprocessed text or images) or features? In either case, please provide a description.}

\begin{itemize}
\item Each instance consists of RGB-D images, trajectories, and the ground-truth 3D scene.
\end{itemize}

\item \textbf{Is there a label or target associated with each instance?} \textit{If so, please provide a description.}

\begin{itemize}
\item The RGB-D video sequences for each perturbed setting are rendered from a 3D scan from the Replica dataset~\cite{straub2019replica}.
\end{itemize}

\item \textbf{Is any information missing from individual instances?} \textit{If so, please provide a description, explaining why this information is missing (e.g., because it was unavailable). This does not include intentionally removed information, but might include, e.g., redacted text.}

\begin{itemize}
\item No.
\end{itemize}

\item \textbf{Are relationships between individual instances made explicit (e.g., users' movie ratings, social network links)?} \textit{If so, please describe how these relationships are made explicit.}

\begin{itemize}
\item Each RGB-D video sequence is rendered in a 3D scene of Replica, conditioned on a trajectory and a set of perturbations.
\end{itemize}

\item \textbf{Are there recommended data splits (e.g., training, development/validation, testing)?} \textit{If so, please provide a description of these splits, explaining the rationale behind them.}

\begin{itemize}
\item No. Our benchmark is intended solely for evaluation because RGB-D SLAM models typically follow an online model optimization/adaptation approach and do not require an additional training stage.
\end{itemize}

\item \textbf{Are there any errors, sources of noise, or redundancies in the dataset?} \textit{If so, please provide a description.}

\begin{itemize}
\item No.
\end{itemize}

\item \textbf{Is the dataset self-contained, or does it link to or otherwise rely on external resources (e.g., websites, tweets, other datasets)?} \textit{If it links to or relies on external resources, a) are there guarantees that they will exist, and remain constant, over time; b) are there official archival versions of the complete dataset (i.e., including the external resources as they existed at the time the dataset was created); c) are there any restrictions (e.g., licenses, fees) associated with any of the external resources that might apply to a future user? Please provide descriptions of all external resources and any restrictions associated with them, as well as links or other access points, as appropriate.}

\begin{itemize}
\item The benchmark is self-contained. We provide all the details and instructions at \url{https://github.com/Xiaohao-Xu/SLAM-under-Perturbation}.
\end{itemize}

\item \textbf{Does the dataset contain data that might be considered confidential (e.g., data that is protected by legal privilege or by doctor–patient confidentiality, data that includes the content of individuals’ non-public communications)?} \textit{If so, please provide a description.}

\begin{itemize}
\item No. The 3D scans used in our \textit{Noisy-Replica} benchmark are sourced from existing open-source datasets.
\end{itemize}

\item \textbf{Does the dataset contain data that, if viewed directly, might be offensive, insulting, threatening, or might otherwise cause anxiety?} \textit{If so, please describe why.}

\begin{itemize}
\item No.
\end{itemize}

\item \textbf{Does the dataset relate to people?} \textit{If not, you may skip the remaining questions in this section.}

\begin{itemize}
\item No. This dataset does not relate to people.
\end{itemize}

\item \textbf{Does the dataset identify any subpopulations (e.g., by age, gender)?}

\begin{itemize}
\item N/A.
\end{itemize}

\item \textbf{Is it possible to identify individuals (i.e., one or more natural persons), either directly or indirectly (i.e., in combination with other data) from the dataset?} \textit{If so, please describe how.}

\begin{itemize}
\item N/A.
\end{itemize}

\item \textbf{Does the dataset contain data that might be considered sensitive in any way (e.g., data that reveals racial or ethnic origins, sexual orientations, religious beliefs, political opinions or union memberships, or locations; financial or health data; biometric or genetic data; forms of government identification, such as social security numbers; criminal history)?} \textit{If so, please provide a description.}

\begin{itemize}
\item No.
\end{itemize}

\item \textbf{Any other comments?}

\begin{itemize}
\item We caution discretion on behalf of the user and call for responsible usage of the benchmark for research purposes only.
\end{itemize}

\vspace{-0.5em}\begin{figure}[!h]
\subsection{Collection Process}\label{subsec:datasheet-collection}\vspace{-1em}
\end{figure}

\item \textbf{How was the data associated with each instance acquired?} \textit{Was the data directly observable (e.g., raw text, movie ratings), reported by subjects (e.g., survey responses), or indirectly inferred/derived from other data (e.g., part-of-speech tags, model-based guesses for age or language)? If data was reported by subjects or indirectly inferred/derived from other data, was the data validated/verified? If so, please describe how.}

\begin{itemize}
\item The 3D scenes used for SLAM data generation in our benchmark are sourced from the existing open-source dataset Replica~\cite{straub2019replica}. The details of the benchmark construction are provided in the Experiment section (Sec. 5) of the main paper, with further details in Sec.~\ref{sec:benchmark-setup} of the Appendix.
\end{itemize}

\item \textbf{What mechanisms or procedures were used to collect the data (e.g., hardware apparatus or sensor, manual human curation, software program, software API)?} \textit{How were these mechanisms or procedures validated?}

\begin{itemize}
\item We did not collect additional raw data. Our main contribution is the development of a perturbation taxonomy and toolbox to transform existing clean SLAM datasets and 3D scenes into noisy SLAM datasets with perturbations for robustness evaluation.
\end{itemize}

\item \textbf{If the dataset is a sample from a larger set, what was the sampling strategy (e.g., deterministic, probabilistic with specific sampling probabilities)?}

\begin{itemize}
\item Our benchmark does not include new raw data collection. We render the RGB-D sensor streams using 3D scene models sourced from the Replica dataset~\cite{straub2019replica}, which comprises real 3D scans of indoor scenes. We selected the same set of eight rooms and offices as the (clean) Replica-SLAM dataset~\cite{imap} for consistent comparison.
\end{itemize}

\item \textbf{Who was involved in the data collection process (e.g., students, crowdworkers, contractors) and how were they compensated (e.g., how much were crowdworkers paid)?}

\begin{itemize}
\item N/A. Our benchmark does not include new raw data collection.
\end{itemize}

\item \textbf{Over what timeframe was the data collected? Does this timeframe match the creation timeframe of the data associated with the instances (e.g., recent crawl of old news articles)?} \textit{If not, please describe the timeframe in which the data associated with the instances was created.}

\begin{itemize}
\item N/A. Our benchmark does not include new raw data collection.
\end{itemize}

\item \textbf{Were any ethical review processes conducted (e.g., by an institutional review board)?} \textit{If so, please provide a description of these review processes, including the outcomes, as well as a link or other access point to any supporting documentation.}

\begin{itemize}
\item N/A. Our benchmark does not include new raw data collection.
\end{itemize}

\item \textbf{Does the dataset relate to people?} \textit{If not, you may skip the remaining questions in this section.}

\begin{itemize}
\item No.
\end{itemize}

\item \textbf{Did you collect the data from the individuals in question directly, or obtain it via third parties or other sources (e.g., websites)?}

\begin{itemize}
\item N/A. Our dataset does not relate to people.
\end{itemize}

\item \textbf{Were the individuals in question notified about the data collection?} \textit{If so, please describe (or show with screenshots or other information) how notice was provided, and provide a link or other access point to, or otherwise reproduce, the exact language of the notification itself.}

\begin{itemize}
\item N/A. Our dataset does not relate to people.
\end{itemize}

\item \textbf{Did the individuals in question consent to the collection and use of their data?} \textit{If so, please describe (or show with screenshots or other information) how consent was requested and provided, and provide a link or other access point to, or otherwise reproduce, the exact language to which the individuals consented.}

\begin{itemize}
\item N/A. Our dataset does not relate to people.
\end{itemize}

\item \textbf{If consent was obtained, were the consenting individuals provided with a mechanism to revoke their consent in the future or for certain uses?} \textit{If so, please provide a description, as well as a link or other access point to the mechanism (if appropriate).}

\begin{itemize}
\item N/A. Our dataset does not relate to people.
\end{itemize}

\item \textbf{Has an analysis of the potential impact of the dataset and its use on data subjects (e.g., a data protection impact analysis) been conducted?} \textit{If so, please provide a description of this analysis, including the outcomes, as well as a link or other access point to any supporting documentation.}

\begin{itemize}
\item We discuss the limitations of our current work in the Conclusion and Future Work section of the main paper, and we plan to further investigate and analyze the impact of our benchmark in future work. We acknowledge the potential data biases and limitations of our initial benchmark and have detailed the assumptions made during benchmark construction in Sec.~\ref{subsec:assumption} of the Appendix.
\end{itemize}

\item \textbf{Any other comments?}

\begin{itemize}
\item No.
\end{itemize}

\vspace{-0.5em}\begin{figure}[!h]
\subsection{Preprocessing, Cleaning, and/or Labeling}\label{subsec:datasheet-preprocess}
\vspace{-1em}
\end{figure}

\item \textbf{Was any preprocessing/cleaning/labeling of the data done (e.g., discretization or bucketing, tokenization, part-of-speech tagging, SIFT feature extraction, removal of instances, processing of missing values)?} \textit{If so, please provide a description. If not, you may skip the remainder of the questions in this section.}

\begin{itemize}
\item No preprocessing or labeling was performed.
\end{itemize}

\item \textbf{Was the “raw” data saved in addition to the preprocessed/cleaned/labeled data (e.g., to support unanticipated future uses)?} \textit{If so, please provide a link or other access point to the “raw” data.}

\begin{itemize}
\item N/A. No preprocessing or labeling was performed for creating the scenarios.
\end{itemize}

\item \textbf{Is the software used to preprocess/clean/label the instances available?} \textit{If so, please provide a link or other access point.}

\begin{itemize}
\item N/A. No preprocessing or labeling was performed for creating the scenarios.
\end{itemize}

\item \textbf{Any other comments?}

\begin{itemize}
\item No.
\end{itemize}

\vspace{-0.5em}\begin{figure}[!h]
\subsection{Uses}\label{subsec:datasheet-uses}
\vspace{-1em}
\end{figure}

\item \textbf{Has the dataset been used for any tasks already?} \textit{If so, please provide a description.}

\begin{itemize}
\item Not yet. We present a new benchmark.
\end{itemize}

\item \textbf{Is there a repository that links to any or all papers or systems that use the dataset?} \textit{If so, please provide a link or other access point.}

\begin{itemize}
\item We will provide links to works that use our benchmark at \url{https://github.com/Xiaohao-Xu/SLAM-under-Perturbation}.
\end{itemize}

\item \textbf{What (other) tasks could the dataset be used for?}

\begin{itemize}
\item The primary use case of our benchmark is to study the robustness of RGB-D SLAM models under perturbations.
\item While we did not explore this direction in the present work, our benchmark can be used for research on the robustness of RGB-D sensing-related downstream tasks in the future.
\end{itemize}

\item \textbf{Is there anything about the composition of the dataset or the way it was collected and preprocessed/cleaned/labeled that might impact future uses?} \textit{For example, is there anything that a future user might need to know to avoid uses that could result in unfair treatment of individuals or groups (e.g., stereotyping, quality of service issues) or other undesirable harms (e.g., financial harms, legal risks)? If so, please provide a description. Is there anything a future user could do to mitigate these undesirable harms?}

\begin{itemize}
\item No.
\end{itemize}

\item \textbf{Are there tasks for which the dataset should not be used?} \textit{If so, please provide a description.}

\begin{itemize}
\item No.
\end{itemize}

\item \textbf{Any other comments?}

\begin{itemize}
\item No.
\end{itemize}

\vspace{-0.5em}\begin{figure}[!h]
\subsection{Distribution and License}\label{subsec:datasheet-distribution}
\vspace{-1em}
\end{figure}

\item \textbf{Will the dataset be distributed to third parties outside of the entity (e.g.,
 company, institution, organization) on behalf of which the dataset was created?} \textit{If so, please provide a description.}

\begin{itemize}
\item Yes, this benchmark has been open-sourced.
\end{itemize}

\item \textbf{How will the dataset be distributed (e.g., tarball on website, API, GitHub)?} \textit{Does the dataset have a digital object identifier (DOI)?}

\begin{itemize} 
\item Our benchmark and the code used for evaluation are available at \url{https://github.com/Xiaohao-Xu/SLAM-under-Perturbation}.
\end{itemize}

\item \textbf{When will the dataset be distributed?}

\begin{itemize}
\item May 24, 2024, and onward.
\end{itemize}

\item \textbf{Will the dataset be distributed under a copyright or other intellectual property (IP) license, and/or under applicable terms of use (ToU)?} \textit{If so, please describe this license and/or ToU, and provide a link or other access point to, or otherwise reproduce, any relevant licensing terms or ToU, as well as any fees associated with these restrictions.}

\begin{itemize}
\item The 3D scene dataset Replica and the benchmarking methods used in our benchmark are sourced from existing open-source repositories, as illustrated in Sec.~\ref{sec:public-assets}. The license associated with them is followed accordingly.
\item Our code is released under the {Apache-2.0} license.
\end{itemize}

\item \textbf{Have any third parties imposed IP-based or other restrictions on the data associated with the instances?} \textit{If so, please describe these restrictions, and provide a link or other access point to, or otherwise reproduce, any relevant licensing terms, as well as any fees associated with these restrictions.}

\begin{itemize}
\item We release it under the {Apache-2.0} license.
\item We do not own the copyright of the original 3D scenes used for rendering our SLAM benchmark.
\end{itemize}

\item \textbf{Do any export controls or other regulatory restrictions apply to the dataset or to individual instances?} \textit{If so, please describe these restrictions, and provide a link or other access point to, or otherwise reproduce, any supporting documentation.}

\begin{itemize}
\item No.
\end{itemize}

\item \textbf{Any other comments?}

\begin{itemize}
\item No.
\end{itemize}

\vspace{-0.5em}\begin{figure}[!h]
\subsection{Maintenance}\label{subsec:datasheet-mainteanance}
\vspace{-1em}
\end{figure}

\item \textbf{Who will be supporting/hosting/maintaining the dataset?}

\begin{itemize}
\item The Robotics Department of the University of Michigan, Ann Arbor, will be supporting, hosting, and maintaining the benchmark.
\item The first author, Xiaohao Xu, will be the main manager of the benchmark.
\end{itemize}

\item \textbf{How can the owner/curator/manager of the dataset be contacted (e.g., email address)?}

\begin{itemize}
\item Robotics Department of the University of Michigan, Ann Arbor: \url{https://robotics.umich.edu/}
\item Xiaohao Xu: \url{xiaohaox@umich.edu}
\end{itemize}

\item \textbf{Is there an erratum?} \textit{If so, please provide a link or other access point.}

\begin{itemize}
\item There is no erratum for our initial release. Errata will be documented as future releases on the benchmark website.
\end{itemize}

\item \textbf{Will the dataset be updated (e.g., to correct labeling errors, add new instances, delete instances)?} \textit{If so, please describe how often, by whom, and how updates will be communicated to users (e.g., mailing list, GitHub)?}

\begin{itemize}
\item Yes, our benchmark will be updated. We plan to expand scenarios, metrics, and models to be evaluated.
\end{itemize}

\item \textbf{If the dataset relates to people, are there applicable limits on the retention of the data associated with the instances (e.g., were individuals in question told that their data would be retained for a fixed period of time and then deleted)?} \textit{If so, please describe these limits and explain how they will be enforced.}

\begin{itemize}
\item N/A. Our dataset does not relate to people.
\end{itemize}

\item \textbf{Will older versions of the dataset continue to be supported/hosted/maintained?} \textit{If so, please describe how. If not, please describe how its obsolescence will be communicated to users.}

\begin{itemize}
\item We will host other versions.
\end{itemize}

\item \textbf{If others want to extend/augment/build on/contribute to the dataset, is there a mechanism for them to do so?} \textit{If so, please provide a description. Will these contributions be validated/verified? If so, please describe how. If not, why not? Is there a process for communicating/distributing these contributions to other users? If so, please provide a description.}

\begin{itemize}
\item Users may contact us by reporting an issue on our benchmark GitHub page \url{https://github.com/Xiaohao-Xu/SLAM-under-Perturbation} or directly contacting the author of this project (Xiaohao Xu, \url{xiaohaox@umich.edu}) to request adding new scenarios, metrics, or models.
\end{itemize}

\item \textbf{Any other comments?}

\begin{itemize}
\item No.
\end{itemize}

\end{enumerate}

\clearpage

\section{More Details about Perturbation Taxonomy  }\label{sec:perturbation-taxonomy-appendix}

\begin{figure*}[ht!]
	\centering
	\includegraphics[width=\textwidth]{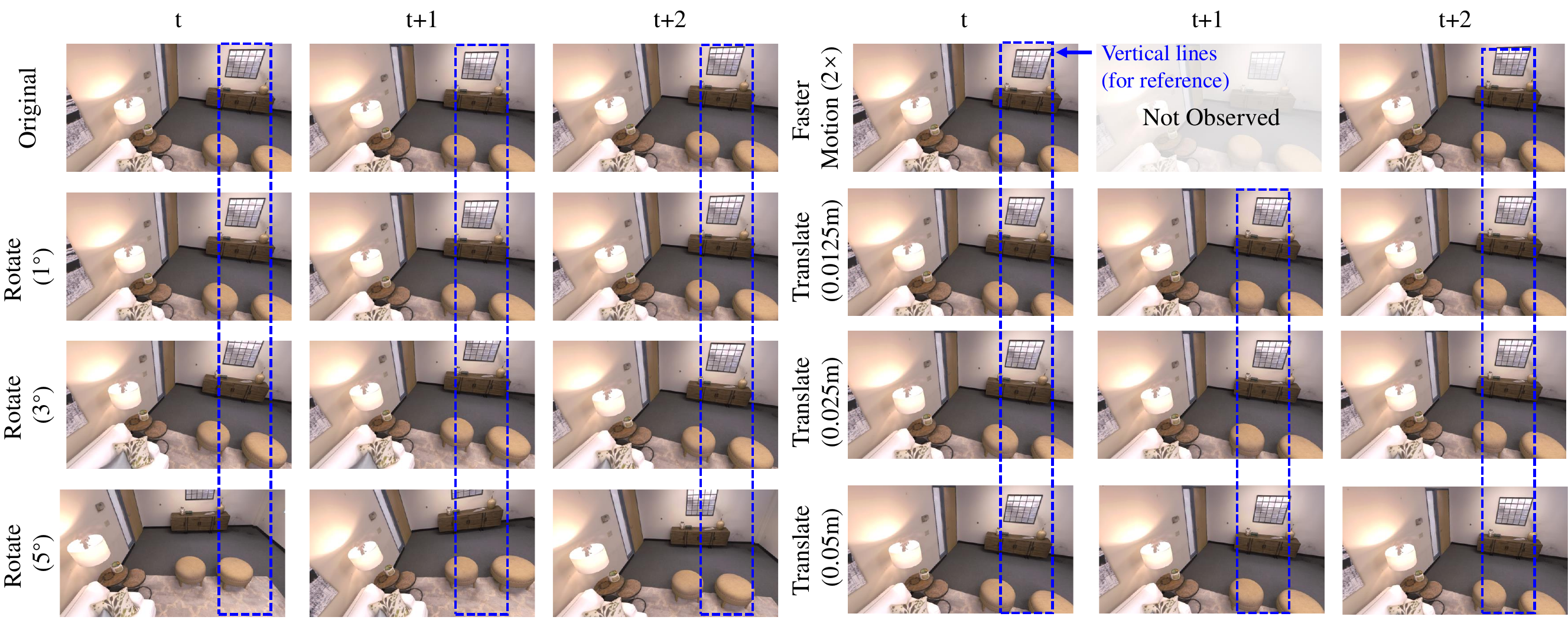} \caption{\textbf{{Rendered RGB image streams under trajectory-level perturbations}}, including translation deviations (Translate), rotation deviations (Rotate), and the faster motion effect. } 
	\label{fig:trajectory-perturbation-illustration}
\end{figure*}

\begin{figure}[t!]
	\centering
	\includegraphics[width=\textwidth]{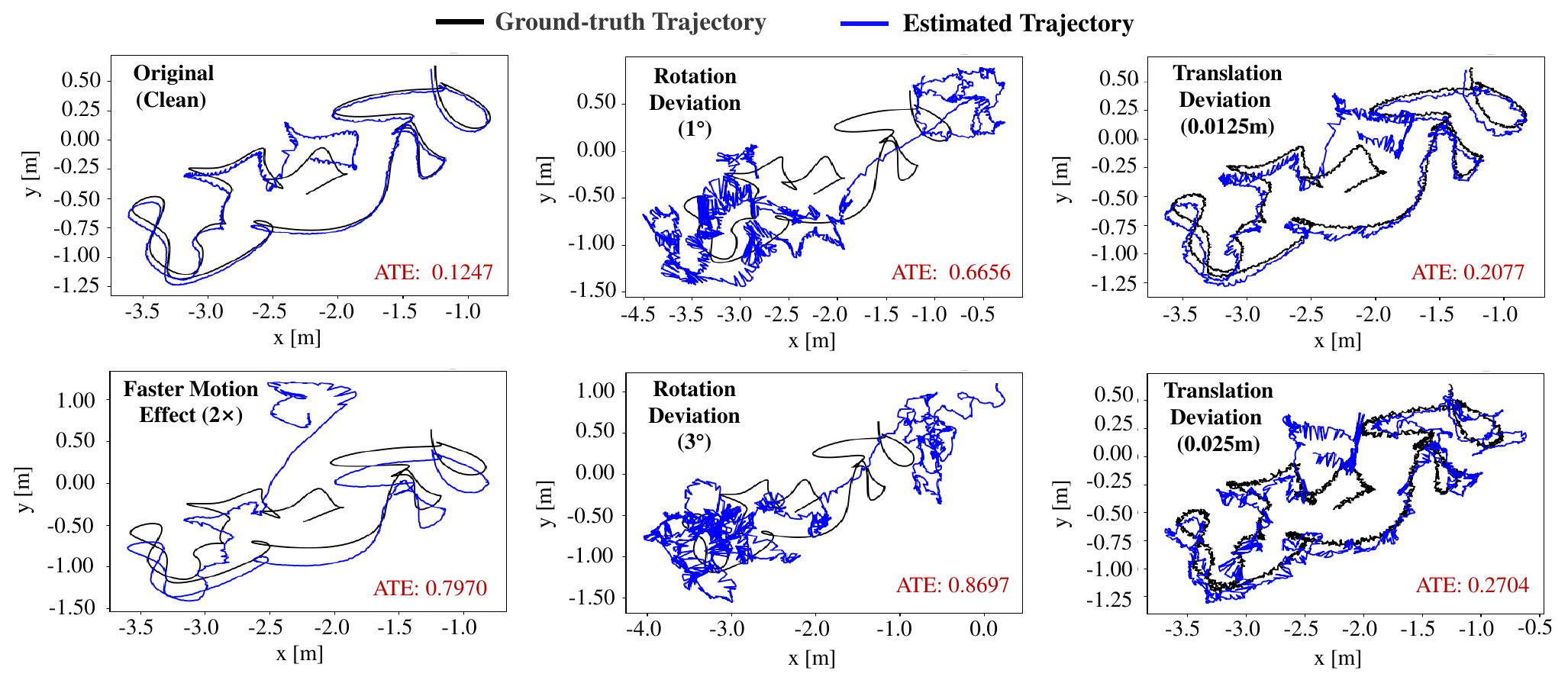} 
	\caption{\textbf{Illustrations of motion deviations and the faster motion effect.} We present the synthesized ground-truth trajectories (in black) and the estimated trajectories (in blue) obtained using the CO-SLAM~\cite{coslam} model. For clarity, we visualize the projected trajectory on the horizontal x-y plane derived from the 3D trajectory, which shows that slight trajectory deviations can have a  significant impact on the trajectory estimation performance.}
	\label{fig:taxonomy-trajectory-misalign-perturbation-simplified}
\end{figure} 

\subsection{{ Perturbation on Sensor Poses}}\label{subsec:perturbation-taxonomy-appendix-pose}

Real-world deployments of embodied agents face challenges such as robotic platform vibrations, uneven terrain, and dynamic motions, which impact sensor pose estimations and degrade SLAM performance. Most existing benchmarks~\cite{straub2019replica, chang2015shapenet} use smooth trajectories, resulting in stable sensor observations. However, this does not reflect the complexities of real-world scenarios with unstable sensor movements. For example, deploying a visual SLAM system on a legged robot traversing uneven terrain introduces significant motion deviations and vibrations, affecting sensor pose estimates and visual observations.
Our work addresses this gap by introducing trajectory and motion perturbations that simulate real-world complexities and instabilities in sensor poses. Our findings indicate that even advanced SLAM models can fail and lose tracking when confronted with unstable observations due to dynamic motions, which reveals the importance to consider motion perturbations.

Specifically, we consider the following two main categories of motion perturbations:

\noindent \textbf{\xxh{Motion deviations}}. To simulate sensor pose vibrations experienced by mobile embodied agents, we introduce motion deviations. These deviations perturb the original sensor pose by applying a rotation perturbation ($\Delta \mathbf{R} \in \boldsymbol{SO}(3)$) and a translation perturbation ($\Delta \mathbf{t} \in \mathbb{R}^3$). The perturbed rotation matrix is calculated as $\mathbf{R}' = \mathbf{R} \Delta \mathbf{R}$, and the perturbed translation vector as $\mathbf{t}' = \mathbf{t} + \Delta \mathbf{t}$. Specifically, the translation and rotation transformations are randomly perturbed using values sampled from a Gaussian distribution. 
In Fig.~\ref{fig:trajectory-perturbation-illustration}, we present the rendered sensor streams under varying severity levels of trajectory-level perturbations, encompassing translational deviations, rotational deviations, and faster motion effects. Although the rotational and translational deviations we examined result in minor changes in observations between adjacent frames, these perturbations lead to significant performance degradation across the majority of benchmarking SLAM models. As depicted in Fig.~\ref{fig:taxonomy-trajectory-misalign-perturbation-simplified}, even slight trajectory-level deviations can have a substantial impact on trajectory estimation performance.

\noindent \textbf{\xxh{Faster motion effect.}}
To evaluate the robustness of perception models for embodied agents under agile motion, we introduce a faster motion scenario by down-sampling the original sensor stream along the time axis.

\subsection{{ Perturbation on RGB Sensor Imaging}}\label{subsec:perturbation-taxonomy-appendix-rgb}

{The perturbations on RGB imaging are designed to model potential error sources throughout the entire RGB image formation and processing pipeline, from the 3D world to the final 2D image. The perturbation sources include environmental interference effects that affect light transmission, blurring effects partially caused by lens-related distortions, noise due to imperfections in the image sensors, and post-processing effects on the image.} Prominent RGB image perturbations~\cite{hendrycks2019robustness} (see Fig. 2b of the main paper) are detailed in the following paragraphs.

\noindent\textbf{{Environmental interference}}. Environmental interference
including weather effects~\cite{kamann2020increasing,yu2019nighttime} are commonly simulated via alpha blending techniques~\cite{smith1995image}. This involves blending a perturbed enviromental effect layer with the clean image to generate a composite perturbed image.
\begin{itemize}
\item \textbf{Snow effect}. To simulate the disturbances caused by snowfall, we construct a snow effect layer with random regional white values~\cite{von2019simulating}. 
    
\item \textbf{Frost effect}. The frost effect introduces a semi-transparent whitening overlay on the image~\cite{michaelis2019benchmarking}. This effect is modeled through a weighted combination of the original image and the whitened version of the image. 

    \item \textbf{Fog effect}. The fog effect results in a hazy observation. A simplified model~\cite{hendrycks2019robustness} to simulate this effect is achieved through linear interpolation between the original image and a constant gray-value image.

    \item \textbf{Spatter effect}. The spatter effect mimics the appearance of droplets on a lens or window. To achieve this effect, a layer comprising semi-transparent dark spots or streaks is blended with the image. Considering the local property of the spatters, we incorporate a hard mask onto the spatter effect layer. This mask designates transparent regions as 0 and perturbed (occupied) regions as 1, thereby controlling the visibility of the spatter effect in specific areas. 
 
\end{itemize}

\noindent \textbf{{Lens-related distortions (specifically blur)}}. 
\xxh{To simulate various blur effects caused by lens imperfections or camera motion, we convolve the input image with specific blur kernels, resulting in a blurred image. }
\begin{itemize}

\item  \textbf{Defocus blur}. This effect simulates the out-of-focus visuals caused by camera lens properties. It can be modeled by convolving the input image with a circular disc (bokeh) kernel.
 
\item  \textbf{Glass blur}. Emulating the appearance of viewing through textured or patterned glass, this effect adds complexity to the blurring process. The glass texture is approximated using an irregular kernel.

\item \textbf{Motion blur}. Rapid movement during image capture, either by the camera or objects in the scene, results in motion blur. This effect can be represented by convolving the image with a linear kernel oriented in the direction of motion.

\item \textbf{Gaussian blur}. 
Gaussian blur convolves the image with a Gaussian kernel. The standard deviation of Gaussian distribution determines the blurring level.
\end{itemize}

    \noindent \textbf{{Sensor noises}}. RGB image sensors inherently introduce noise during image acquisition, impacting image quality.
    \begin{itemize}
    \item \textbf{Gaussian noise}. To simulate the presence of Gaussian noise, we introduce additive noise  \xxh{for each pixel in the original clean image} to create a corrupted version of the image.
    The noise is sampled from a Gaussian distribution with zero mean. 

    \item \textbf{Shot noise}. 
    Shot noise is associated with the random arrival of photons or particles during the image capture process. It can be modeled using a Poisson distribution~\cite{hasinoff2014photon}.

    \item \textbf{Impulse noise}. \xxh{To simulate impulse noise, we introduce randomness to each pixel. For every pixel, we sample one random value  between 0 and 1. If the sampled value falls within the range of 0 to $a/2$ (where $a\in(0,1)$), the pixel is assigned the minimum intensity value, while it is set to the maximum intensity value if the value lies between $a/2$ and $a$. Finally, if the sampled value is between $a$ and 1, the pixel remains unchanged.}
        \item \textbf{Speckle noise}. Speckle noise \cite{bianco2018strategies}, applied to \xxh{each pixel $(x,y)$ of} the original clean image $I$, can be modeled as a multiplicative noise process:
        \begin{equation}
        I'(x,y) = I(x,y) \times (1 + \rho \times \eta)
        \end{equation}
    where $\rho$ controls the intensity level of the speckle noise, and $\eta$ represents the term of Gaussian noise.
\end{itemize}

   \noindent \textbf{{Post-processing}}. Image post-processing techniques can introduce perturbations that alter the original pixel values.
    \begin{itemize}
    \item \textbf{Brightness}. This effect adjusts the global luminance of the image. Image brightness is adjusted by adding a constant offset \xxh{to each pixel}.

    \item \textbf{Contrast}. This effect alters the tone variance  of \xxh{each pixel $(x,y)$ of} the image by linear scaling about the mean intensity $\mathcal{J}$:
   \xxh{\begin{equation}
    I'(x,y) = \beta \times (I(x,y) - \mathcal{J}) + \mathcal{J}
    \end{equation}}
    where $\beta$ controls the contrast level.

    \item \textbf{JPEG compression}. This effect simulates lossy compression artifacts when using the JPEG image compression. 
    
    \item \textbf{Pixelate}. This effect reduces resolution by dividing the image into blocks and setting all pixels in each block to the block's average value. 

    \end{itemize}

\begin{figure*}[t]
	\centering
\includegraphics[width=\textwidth]{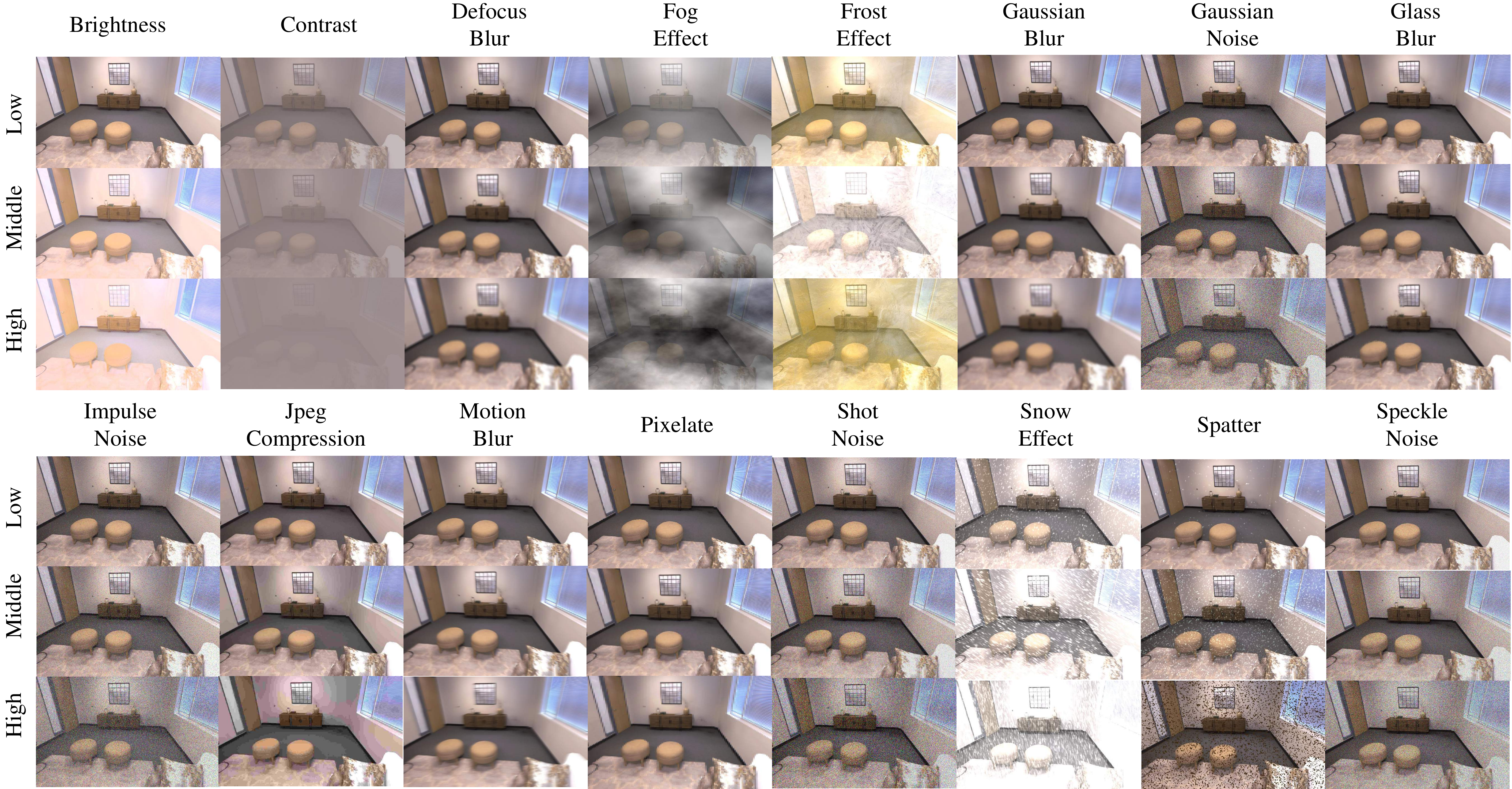} 
	\caption{
\textbf{Illustration of  RGB imaging perturbations under different severity levels}. We consider \textbf{16} common image corruption types~\cite{hendrycks2019robustness} from \textbf{\textit{4}} main categories of perturbations for  robustness evaluation: \textbf{(1)} \textbf{{noise-based distortions}}: \textit{Gaussian Noise}, \textit{Shot Noise}, \textit{Impulse Noise}, and \textit{Speckle Noise}; \textbf{(2)} \textbf{{blur-based effects}}: \textit{Defocus Blur}, \textit{Glass Blur}, \textit{Motion Blur}, and \textit{Gaussian Blur}; \textbf{(3)} \textbf{{environmental interferences}}: \textit{Snow Effect}, \textit{Frost Effect}, \textit{Fog Effect}, and \textit{Spatter Effect}. \textbf{(4)} \textbf{{post-processing manipulations}}: \textit{Brightness}, \textit{Contrast}, \textit{Pixelate}, and \textit{JPEG Compression}. Each perturbation type is further split into \textbf{\textit{3}} severity levels (low, middle, and high), which corresponds to Level 1, Level 3, and Level 5 of Table~\ref{tab:configuration_rgb_imaging}.  } 
	\label{fig:sensor-corruption-full}
\end{figure*}

\noindent\textbf{Implementation of RGB imaging perturbations.} As shown in Table~\ref{tab:configuration_rgb_imaging}, we define five severity levels for each type of RGB imaging perturbation, following established robustness evaluation literature~\cite{hendrycks2019robustness}. The specific implementation details can be found in our RGB Imaging Perturbation Synthesis Toolbox, available on \href{https://github.com/Xiaohao-Xu/SLAM-under-Perturbation/blob/main/benchmark/Co-SLAM/datasets/robustness.py}{our GitHub}. We illustrate RGB imaging perturbations under different severity levels in Fig.~\ref{fig:sensor-corruption-full}.

\begin{table}[t]
\centering

\caption{Specific configurations of the RGB imaging perturbations.}
\renewcommand\arraystretch{1.2}
\resizebox{\textwidth}{!} 
{ 
\begin{tabular}{l|c|ccccc}
\toprule \toprule
\textbf{Perturbation} & \textbf{Parameter} & \textbf{Level 1} & \textbf{Level 2} & \textbf{Level 3} & \textbf{Level 4} & \textbf{Level 5} \\
\midrule

{Snow Effect} & \makecell{(Mean, std, scale, \\ threshold, blur radius, \\ blur std, blending ratio)} & \makecell{0.1, 0.3, 3.0, \\ 0.5, 10.0, 4.0, 0.8} & \makecell{0.2, 0.3, 2, \\ 0.5, 12, 4, 0.7} & \makecell{0.55, 0.3, 4, \\ 0.9, 12, 8, 0.7} & \makecell{0.55, 0.3, 4.5, \\ 0.85, 12, 8, 0.65} & \makecell{0.55, 0.3, 2.5, \\ 0.85, 12, 12, 0.55} \\
Frost Effect & (Frost intensity, texture influence) & (1.00, 0.40) & (0.80, 0.60) & (0.70, 0.70) & (0.65, 0.70) & (0.60, 0.75) \\
Fog Effect & (Thickness, smoothness) & (1.5, 2.0) & (2.0, 2.0) & (2.5, 1.7) & (2.5, 1.5) & (3.0, 1.4) \\

Spatter Effect & \makecell{(mean, standard deviation, \\ sigma, threshold,\\ scaling factor, complexity of effect)} & \makecell{(0.65, 0.3, 4, \\ 0.69, 0.6, 0)} & \makecell{(0.65, 0.3, 3, \\  0.68, 0.6, 0)} & \makecell{(0.65, 0.3, 2,\\ 0.68, 0.5, 0)} & \makecell{(0.65, 0.3, 1, \\ 0.65, 1.5, 1)} & \makecell{(0.67, 0.4, 1, \\ 0.65, 1.5, 1)} \\
\midrule
Defocus Blur & (Kernel radius, alias blur) & (3.0, 0.1) & (4.0, 0.5) & (6.0, 0.5) & (8.0, 0.5) & (10.0, 0.5) \\
Glass Blur & (Sigma, max delta, iterations) & (0.7, 1.0, 2.0) & (0.9, 2.0, 1.0) & (1.0, 2.0, 3.0) & (1.1, 3.0, 2.0) & (1.5, 4.0, 2.0) \\
Motion Blur & (Radius, sigma) & (10, 3) & (15, 5) & (15, 8) & (15, 12) & (20, 15) \\
Gaussian Blur & Sigma & 1 & 2 & 3 & 4 & 6 \\

\midrule

Gaussian Noise & Noise scale & 0.08 & 0.12 & 0.18 & 0.26 & 0.38 \\
Shot Noise & Photon number & 60 & 25 & 12 & 5 & 3 \\
Impulse Noise & Noise amount & 0.03 & 0.06 & 0.09 & 0.17 & 0.27 \\
Speckle Noise & Noise scale & 0.15 & 0.2 & 0.35 & 0.45 & 0.6 \\
\midrule

Brightness Increase & Adjustment ratio & 0.1 & 0.2 & 0.3 & 0.4 & 0.5 \\
Contrast Decrease & Adjustment of pixel mean & 0.40 & 0.30 & 0.20 & 0.10 & 0.05 \\
JPEG Compression & Compression quality & 25 & 18 & 15 & 10 & 7 \\
Pixelate & Resize factor & 0.60 & 0.50 & 0.40 & 0.30 & 0.25 \\

\bottomrule \bottomrule
\end{tabular}\label{tab:configuration_rgb_imaging}
}
\end{table}

\clearpage
\subsection{{Perturbation on Depth Sensor Imaging}}\label{subsec:perturbation-taxonomy-appendix-depth}

\begin{figure}[t]
	\centering
\includegraphics[width=0.42\textwidth]{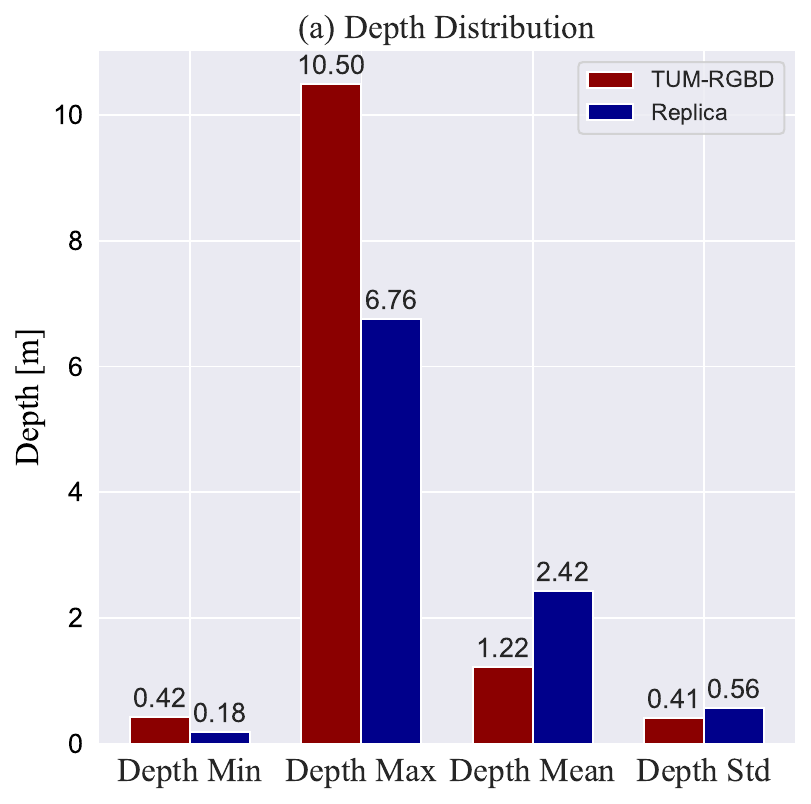}  
\includegraphics[width=0.3\textwidth]{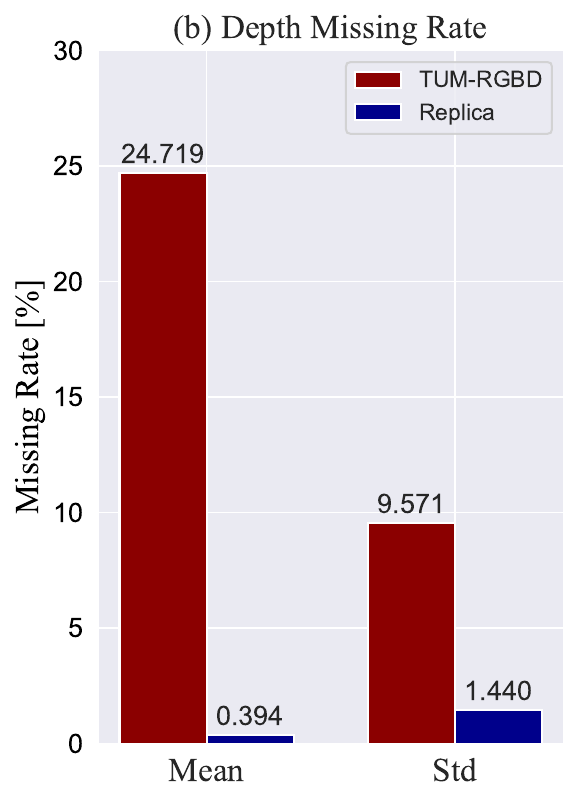} 
	\caption{\textbf{Discrepancy in depth characteristics between existing simulated and real-world SLAM datasets.} Here, we compare the depth (\textbf{a}) distribution and (\textbf{b})  missing rate between real-world collected depth from TUM-RGBD~\cite{TUM-RGBD} dataset and simulated depth from Replica~\cite{straub2019replica} dataset.}
	\label{fig:depth_distribution_comparison}
\end{figure}

As depicted in Fig.~\ref{fig:depth_distribution_comparison}, there exists a noticeable disparity between the current simulated clean depth distribution obtained from the Replica~\cite{straub2019replica} SLAM benchmark and the real noisy depth data derived from the TUM-RGBD~\cite{schubert2018tum} SLAM benchmark. In Replica, the minimum depth measures approximately 0.18 m, whereas the TUM-RGBD data exhibits a minimum depth value of 0.4 m, reflecting the limitations of real-world depth sensors. Notably, we observe a significant discrepancy in the depth missing rates, with TUM-RGBD demonstrating an approximate 25\% missing rate compared to nearly zero (0.39\%) in Replica. These observations underscore the necessity of exploring perturbation strategies for depth imaging to bridge the gap between simulated  and real-world depth.

Specifically, we consider the following four types of depth perturbations that the sensing limitations of real depth sensors:

\noindent\textbf{{{Gaussian noise}}}. \xxh{This operation simulates the random noise inherent to depth sensors, which typically follows a Gaussian distribution. Each pixel $(x,y)$ in the original depth map $D$ is perturbed by adding a Gaussian noise $\eta$, resulting in the corrupted depth map $D'$: $D'(x,y) = D(x,y) + \eta$.}

\noindent\textbf{{{Edge erosion}}}. 
    The multi-path interference effect of certain depth sensors (\textit{e.g.}, time-of-flight sensors) can lead to inaccurate depth measurements, particularly for regions with complex geometries. To simulate this perturbation, we first leverage the edge detection algorithm to obtain the edges and then remove a subset of edge pixels $\mathcal{P}$:
    \begin{equation}
    D'(x,y) = \begin{cases}
    $VOID$ & \text{if } (x,y) \in \mathcal{P} \\
    D(x,y) & \text{otherwise}
    \end{cases}
    \end{equation}

\noindent\textbf{{{Random missing depth data}}}. This perturbation introduces random masked regions to simulate occlusions or missing depth data. Specifically, a binary mask $M$ is  applied to the depth map, where the masked regions are set to a void value:
\begin{equation}
    D' = D  \odot M
    \end{equation}
Here, $\odot$ denotes element-wise multiplication. \xxh{The binary mask $M$ is generated by randomly sampling rectangular patches within the depth image.}
   
\noindent\textbf{{{Range clipping}}}. This perturbation accounts for the limited depth coverage of real-world depth sensors. Objects beyond this range will  appear as missing data in the depth image.  Specifically, any depth value $D(x,y)$ falling outside a specified range  $[D_{min}, D_{max}]$ is replaced with a predefined void value to represent depth missing.

\noindent\textbf{Implementation of depth imaging perturbations.} As shown in Table~\ref{tab:configuration_rgb_imaging}, we define five severity levels for each type of depth imaging perturbation, following the depth distribution of real depth maps~\cite{TUM-RGBD}. The specific implementation details can be found in our Depth Imaging Perturbation Synthesis Toolbox, available on \href{https://github.com/Xiaohao-Xu/SLAM-under-Perturbation/blob/main/benchmark/Co-SLAM/datasets/robustness_depth.py}{our GitHub}.

\subsection{{Perturbation on RGB-D Sensor Synchronization}}  \label{subsec:perturbation-taxonomy-appendix-rgbd}

\xxh{To emulate sensor delays in cases where multiple sensors within an RGB-D sensing system are not synchronized, we introduce temporal misalignment between sensor streams (see Fig. 2d of the main paper). Consider two initially synchronized sensor streams, denoted as $\mathbf{S}_{1}(t)$ and $\mathbf{S}_{2}(t)$. We simulate a delay in the second stream by shifting its sensor sequence by a frame interval $\Delta$. This creates perturbed streams $\mathbf{S}_{1}'(t)= \mathbf{S}_1(t)$ and $\mathbf{S}_{2}'(t)= \mathbf{S}_2(t + \Delta)$. While one sensor stream is shifted, the poses associated with each sensor reading remain unchanged. This ensures the system is operating on data grounded in the past, reflecting the real-world scenario of misaligned sensor information.}

\begin{table}[t]
\centering

\caption{Specific configurations of the depth imaging perturbations.}
\renewcommand\arraystretch{1.2}
\resizebox{\textwidth}{!} 
{ 
\begin{tabular}{l|c|ccccc}
\toprule \toprule
\textbf{Perturbation} & \textbf{Parameter} & \textbf{Level 1} & \textbf{Level 2} & \textbf{Level 3} & \textbf{Level 4} & \textbf{Level 5} \\
\midrule

Gaussian Noise & Noise scale & 0.1 & 0.2 & 0.3 & 0.4 & 0.5 \\
Edge Erosion & Erosion rate & 0.015 & 0.020 & 0.025 & 0.03 & 0.035 \\
Random missing depth data & Missing rate (\%) & 10 & 15 & 20 & 25 & 30 \\
Range clipping & (Min depth, Max depth) &(0.2, 4.4) & (0.3, 4.2)& (0.4, 4.0)& (0.5, 3.8)
& (0.6, 3.6)\\

\bottomrule \bottomrule
\end{tabular}
}\label{tab:configuration_depth_imaging}
\end{table}

\clearpage
\section{More Details about \textit{Noisy-Replica} SLAM Robustness Benchmark}\label{sec:benchmark-setup}

\subsection{Assumptions for Benchmarking Setup}\label{subsec:assumption}

We initialize the \textit{Noisy-Replica} benchmark for SLAM model robustness evaluation under the following assumptions.

\vspace{1.0mm}\noindent\textbf{Task.} We focus on the standard (passive) SLAM setting, assuming the absence of active decision-making  processes.

\vspace{1.0mm}\noindent\textbf{Model.} Our analysis is centered on vision-oriented SLAM scenarios, specifically targeting monocular and RGB-D settings. We assume the use of dense depth representation as opposed to sparse depth data obtained from a LiDAR scanner. In addition, the SLAM system is presumed to have known motion and observation models. 

\vspace{1.0mm}\noindent\textbf{Perturbation.} 
Although our noisy data synthesis pipeline is capable of generating SLAM benchmarks with multiple heterogeneous perturbations, we concentrate on investigating the performance degradation caused by individual sensor or trajectory perturbations. This focused approach is designed to dissect the system's response to isolated perturbations, allowing precise quantification of their specific impacts on SLAM performance. By analyzing the degradation induced by individual perturbations, we can effectively assess the system's robustness in a controlled manner and identify the root causes of performance degradation. This knowledge is crucial for developing targeted mitigation strategies that address the most vulnerable aspects, \textit{i.e.}, \textit{Achilles' Heel}, of the whole SLAM system.
Also, we model these perturbations using simplified linear models (\textit{e.g.}, Gaussian noise assumptions), in line with precedent set by established literature~\cite{hendrycks2019robustness,wang2020tartanair,RobustNav}. While these simplified perturbations may not fully capture the complexity of real-world scenarios, they offer interpretability and facilitate analysis across different perturbation types.

\vspace{1.0mm}\noindent\textbf{3D scene}. We assume that the environment is static, meaning there are no moving or dynamically changing objects within the scene. Also, the scene is bounded, typically referring to an indoor setting with predefined boundaries or limits.

\clearpage

\subsection{\textbf{\textit{Noisy-Replica} Benchmark Statistics}}\label{subsec:benchmark-stat}

\noindent{\textbf{Benchmark sequence number distribution.}} Using our established taxonomy of perturbations for SLAM and the noisy data synthesis pipeline, we have created a large-scale SLAM robustness benchmark called \textit{Noisy-Replica} to evaluate the robustness of monocular and multi-modal RGB-D SLAM methods by incorporating various perturbations that mimic real-world sensor and motion effects. 

Each perturbed setting of our benchmark is rendered in eight scenes from the 3D indoor scan dataset Replica~\cite{straub2019replica}. This process generates eight sequences from a single trajectory. For each perturbed setting, we calculate the average result from 24 experimental data points (eight sequences, each repeated three times). Then, we report the averaged result for each perturbed setting. Specifically, for the RGB imaging perturbation, we present the averaged result across three severity levels, under both static and dynamic perturbation modes.

\begin{figure*}[t!]
	\centering
\includegraphics[width=0.495\textwidth]{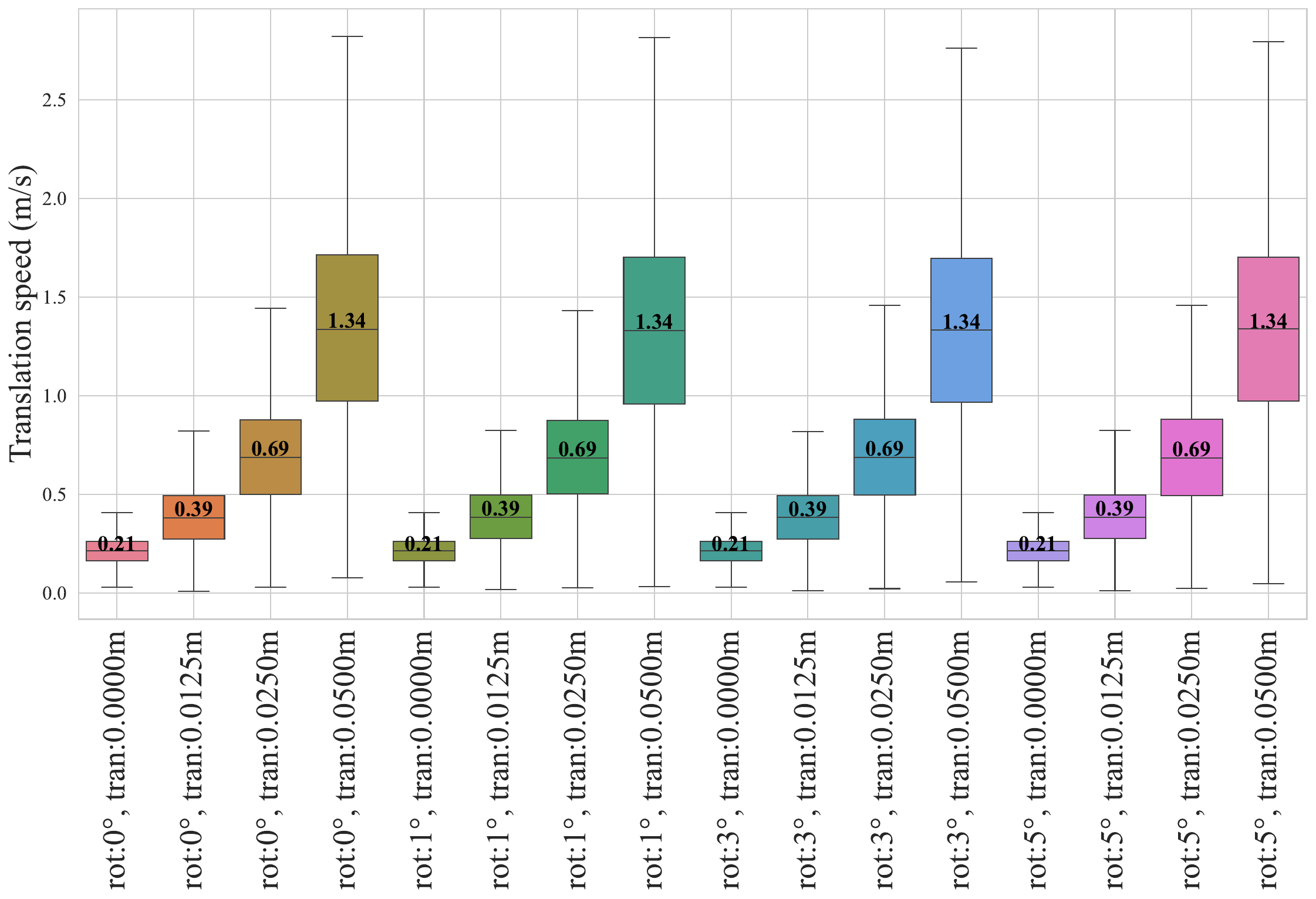}
\includegraphics[width=0.495\textwidth]{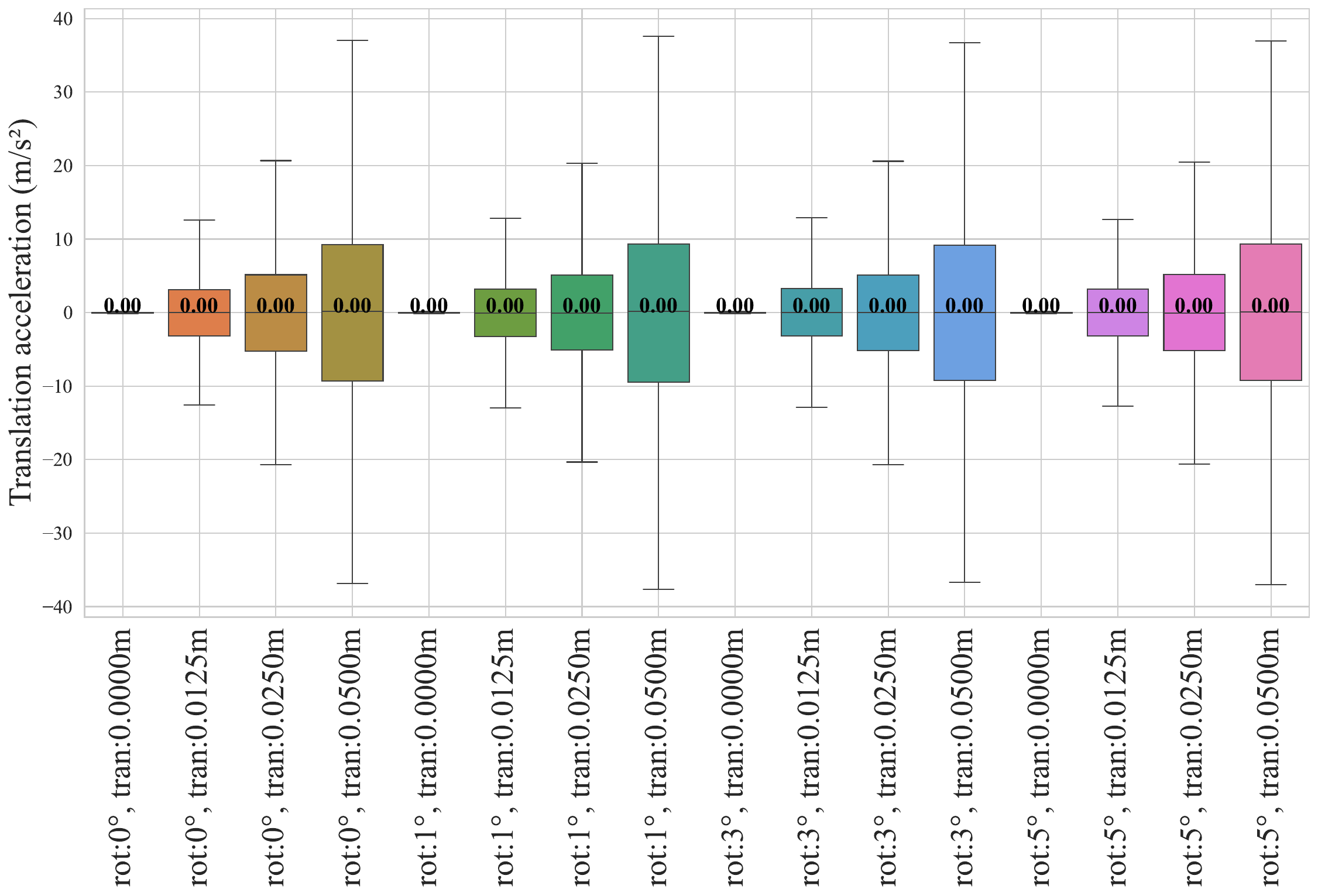}
\includegraphics[width=0.495\textwidth]{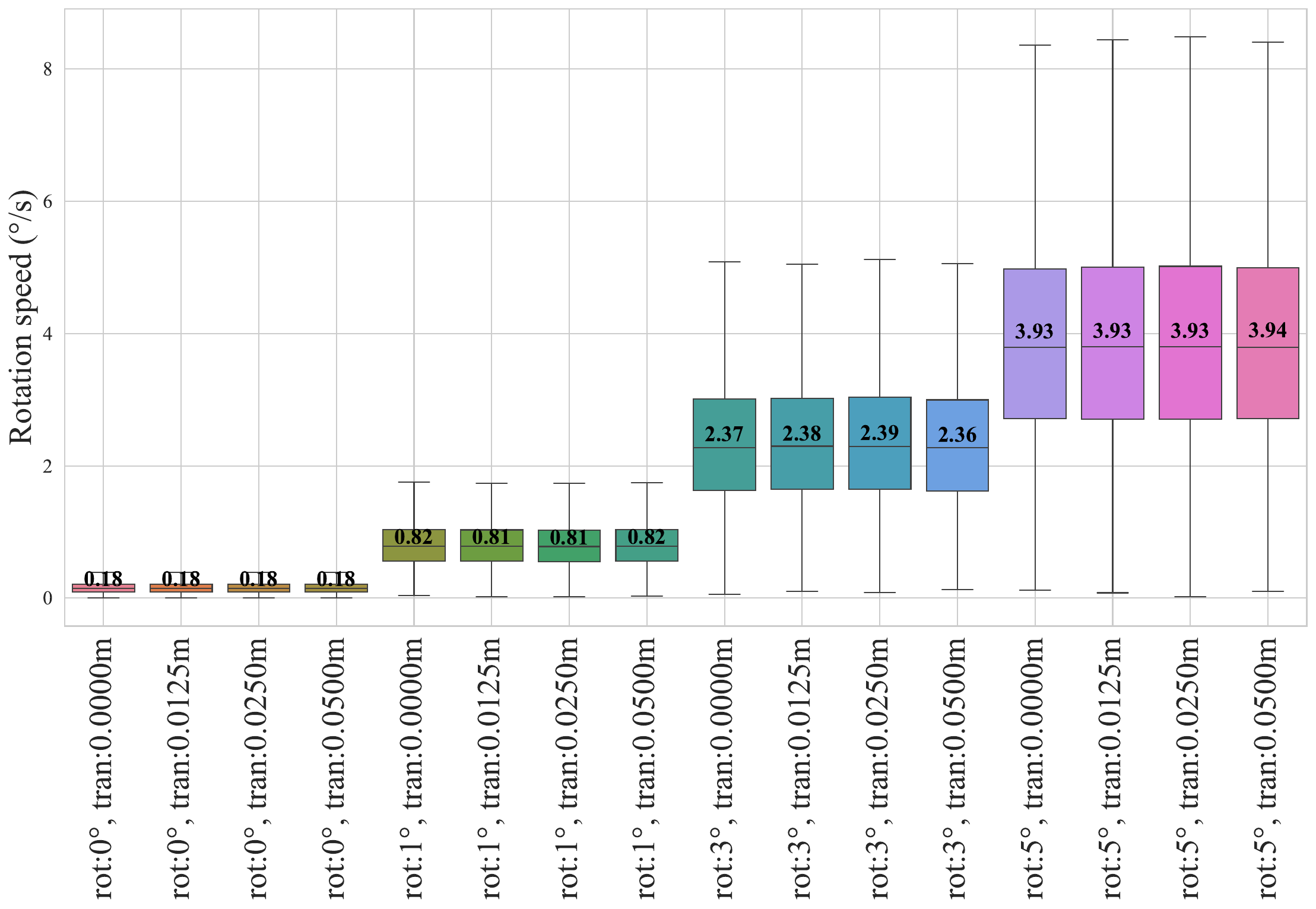}
\includegraphics[width=0.495\textwidth]{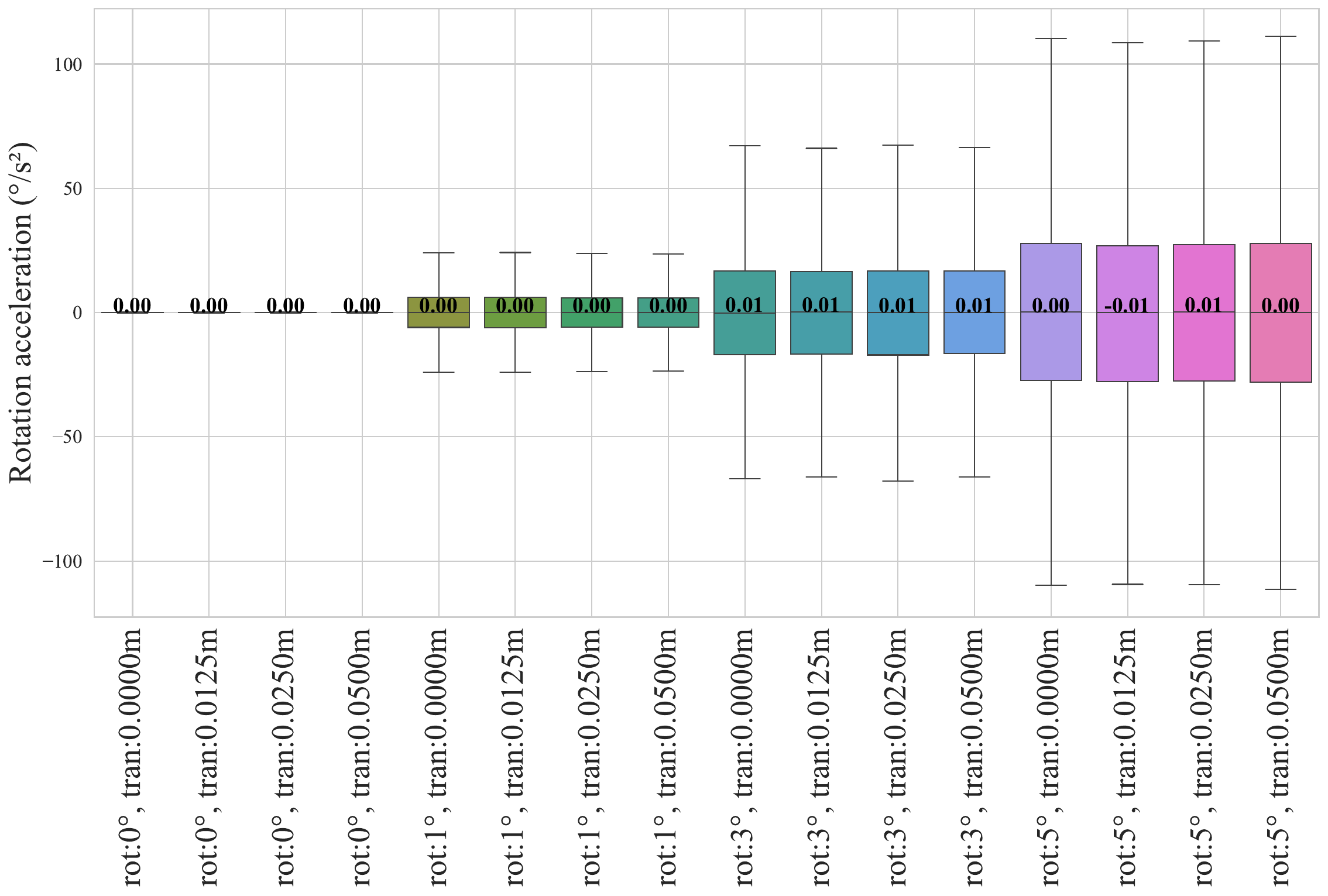}
\caption{\textbf{Motion statistics of trajectory distribution under varying combinations of translation and rotation deviations}. Assuming a frame rate of 20 frames per second for the SLAM system, \textit{i.e.}, a time interval of 0.05 seconds between neighboring pose frames, we present the motion distribution of perturbed trajectories in the proposed \textit{Noisy-Replica} benchmark. The figures show the distribution of translation speed (\textbf{Top Left}), translation acceleration (\textbf{Top Right}), rotation speed (\textbf{Bottom Left}), and rotation acceleration (\textbf{Bottom Right}). We report the mean value of each setting.}
	\label{fig:traj-motion-distribution}
\end{figure*}

We provide details about the specific distribution and setup of each perturbed setting as follows:

\begin{itemize}
  \item \textbf{8 original clean sequences:} These sequences replicate the quality and the sequence number of the original Replica SLAM dataset~\cite{imap}.
  \item \textbf{768 sequences with RGB imaging perturbations:} We apply 16 different types of image-level perturbations at 3 severity levels (Level 1, Level3, and Level 5 of Table~\ref{tab:configuration_rgb_imaging}), both under static and dynamic conditions. 
    \item \textbf{32 sequences with depth imaging perturbations:} This category consists of 4 types of perturbations. For the depth noise, we adopt the hyperparameters of the Gaussian noise distribution as specified in previous literature~\cite{hendrycks2019robustness}. Moreover, we set the depth missing rate to 20\% and establish the depth clipping range based on the real-world depth distribution of the TUM-RGBD dataset~\cite{TUM-RGBD}. The severity strength of each depth perturbation is shown in the Level 3 column of Table~\ref{tab:configuration_depth_imaging}. 
  \item \textbf{24 sequences with faster motion effects:} These sequences involve faster speed than the original sequences, with variations of two, four, and eight times the original speed.
  \item \textbf{120 sequences with motion deviations:} This category includes pure rotation deviation, pure translation deviation, and combined transformation matrix deviation. We define three severity levels for both rotation and translation deviations, and sample the deviation from a Gaussian distribution. Specifically, for rotation deviation, we introduce random deviations in rotation around the x, y, and z axes, with mean values of zero and standard deviations of 1, 3, and 5 degrees at each pose frame. For translation deviation, we introduce random deviations in the x, y, and z axes, with mean values of zero and standard deviations of 0.0125, 0.025, and 0.05 meters at each pose frame. In Fig. \ref{fig:traj-motion-distribution}, we show the motion statistics of the perturbed trajectory sequences under varying combinations of translation and rotation deviations. This category of trajectory-deviated sequences encompasses a broad spectrum of motion speeds and accelerations, enabling a progressive evaluation of the robustness of SLAM models against increasingly challenging motion types. These insights are especially valuable for evaluating the implementation of SLAM systems in high-speed scenarios or on agile robot platforms exposed to significant vibrations.
  \item \textbf{48 sequences with RGB-D sensor de-synchronization:} We consider both static and dynamic perturbation models for multi-sensor misalignment. In the static mode, a constant time delay is synthesized between the two sensor streams, while in the dynamic perturbation model, there is a varying time delay between the streams. Specifically, the multi-sensor misalignment perturbation sequences consist of 24 sequences with a fixed cross-sensor frame delay interval ($\Delta$) of 5, 10, and 20 frames, as well as 24 sequences with dynamic perturbation where $\Delta$ deviates by 1 frame from the fixed intervals of 5, 10, and 20 frames.
\end{itemize}

Overall, this benchmark dataset enables a comprehensive evaluation of existing SLAM algorithms under simulated perturbations, providing a thorough assessment of the robustness of multi-modal SLAM systems in a wide range of challenges.

\clearpage

\subsection{More Details about  Baseline Models for Benchmarking}
\label{subsec:methods}

\noindent\textbf{Additional descriptions about benchmarking models.} While previous SLAM robustness evaluations primarily focused on classical methods~\cite{wang2020tartanair,bujanca2021robust}, our benchmark encompasses both classical and learning-based SLAM systems. As shown in Table~\ref{tab:slam_methods}, in addition to ORB-SLAM3~\cite{orbslam3}, we evaluate Neural SLAM models including iMAP~\cite{imap}, Nice-SLAM~\cite{niceslam}, CO-SLAM~\cite{coslam}, GO-SLAM~\cite{zhang2023goslam}, and SplaTAM-S~\cite{keetha2023splatam}. The hyperparameters are set based on the recommendations given in the original papers or use default settings otherwise.

Below, we offer additional descriptions of SLAM models that have been benchmarked on our \textit{Noisy-Replica} benchmark.

\begin{itemize}
    \item \textbf{ORB-SLAM3}~\cite{orbslam3}: An extension of ORB-SLAM2~\cite{orbslam2} that incorporates a multi-map system and visual-inertial odometry, enhancing robustness and performance.
    \item \textbf{iMAP}~\cite{imap}: A neural RGB-D SLAM system that utilizes the MLP representation to achieve joint tracking and mapping.
    \item \textbf{Nice-SLAM}~\cite{niceslam}: A neural RGB-D SLAM model that employs a multi-level feature grid for scene representation, reducing computational overhead and improving scalability.
    \item \textbf{CO-SLAM}~\cite{coslam}: An advanced neural RGB-D SLAM system with a hybrid representation, enabling robust camera tracking and high-fidelity surface reconstruction in real time.
    \item \textbf{GO-SLAM}~\cite{zhang2023goslam}: A neural visual SLAM framework for real-time optimization of poses and 3D reconstruction. It supports both monocular and RGB-D input settings.
    \item \textbf{SplaTAM}~\cite{keetha2023splatam}: A neural RGB-D SLAM model that follows Gaussian Splatting~\cite{gaussiansplatting} to construct an adaptive map representation based on Gaussian kernels. Due to time and computational constraints, we evaluate the relatively more efficient SplaTAM-S model variant in our benchmark.
\end{itemize}

\begin{table*}[t]
\centering
\caption{\xxh{RGB-D} SLAM methods for robustness evaluation.}
\label{tab:slam_methods}
\resizebox{\textwidth}{!} 
{
\begin{tabular}{l|c|c| l |c|c|r|l|c}
\toprule\toprule
\textbf{Method } & \textbf{Type} & \begin{tabular}[c]{@{}c@{}}\textbf{Modality}\\\textbf{Mono/RGB-D}\end{tabular} & \begin{tabular}[c]{@{}c@{}}\textbf{Map} \textbf{Representation}\end{tabular} & \begin{tabular}[c]{@{}c@{}}\textbf{Loop}\\\textbf{Closure}\end{tabular} & \begin{tabular}[c]{@{}c@{}}\textbf{External}\\\textbf{Data}\end{tabular}& \multicolumn{1}{|c|}{\begin{tabular}[c]{@{}c@{}}\textbf{Speed}\end{tabular}} & \textbf{Processing} & \textbf{Year}\\ 
\midrule 
ORB-SLAM3~\cite{orbslam3}  & Classical, Sparse & $\usym{2713}$\quad/\quad$\usym{2713}$ & Keyframe+ORB, Explicit  &$\usym{2713}$ &  $\usym{2715}$& Real-time & CPU & 2020 \\
iMAP~\cite{imap}  & Neural, Dense  &  $\usym{2715}$\quad/\quad$\usym{2713}$ & NeRF-based${}^{\text{(1)}}$, Implicit &$\usym{2715}$ &  $\usym{2715}$ & Quasi Real-time & CPU+GPU & 2021 \\
Nice-SLAM~\cite{niceslam}  & Neural, Dense & $\usym{2715}$\quad/\quad$\usym{2713}$ & NeRF-based, Implicit &$\usym{2715}$ &  $\usym{2715}$& Quasi Real-time & CPU+GPU & 2022  \\
CO-SLAM~\cite{coslam}  & Neural, Dense & $\usym{2715}$\quad/\quad$\usym{2713}$ & NeRF-based, Implicit &$\usym{2715}$ &  $\usym{2715}$ & Real-time & CPU+GPU & 2023 \\
GO-SLAM~\cite{zhang2023goslam}  & Neural, Dense & $\usym{2713}$\quad/\quad$\usym{2713}$ & NeRF-based, Implicit &$\usym{2713}$ &  $\usym{2713}$${}^{\text{(2)}}$& Quasi Real-time & CPU+GPU & 2023 \\
SplaTAM-S~\cite{keetha2023splatam}  & Neural, Dense & $\usym{2715}$\quad/\quad$\usym{2713}$  & Gaussian~\cite{gaussiansplatting}, Explicit &$\usym{2715}$ &  $\usym{2715}$& Quasi Real-time & CPU+GPU & 2024  \\
\bottomrule\bottomrule
\multicolumn{9}{l}{{$(1)$ `NeRF-based' indicates methods that leverage implicit neural networks to encode the 3D scene, following the philosophy of NeRF~\cite{rosinol2022nerf}.}}\\
\multicolumn{9}{l}{{$(2)$ GO-SLAM initializes the model parameters from the DROID-SLAM model~\cite{teed2021droid} which leverages external data~\cite{wang2020tartanair} for model pre-training.}}\\
\end{tabular}}

\end{table*}

\noindent\textbf{Remark: Neural SLAM models are inherently `trained' on the testing distribution.} Neural SLAM models are optimized (\textit{i.e.}, `trained') on perturbed RGB-D observations, enabling continuous adaptation and updating of internal representations based on incoming data at each timestamp. This inherently includes `training with introduced perturbations', as the models adjust to variations during online operation. Neural SLAM methods like Nice-SLAM~\cite{niceslam}, CO-SLAM~\cite{coslam}, and GO-SLAM~\cite{zhang2023goslam} leverage Neural Radiance Field (NeRF) as the map representation, updating the parameters of NeRF network and the parameters of poses for each frame when new observations arrive during testing; SplaTAM~\cite{keetha2023splatam} leverages explicit Gaussian Splats~\cite{gaussiansplatting} as the map representation, updating the parameters of Gaussian kernels as well as the parameters of poses for each frame during testing.  We find that this test-time online learning mechanism provides better robustness compared to non-neural SLAM methods without adaptation capabilities, allowing neural SLAM models to be robust to static RGBD imaging perturbations by continuously refining their environment understanding for optimal performance. Generally speaking, our experimental setup adheres to the standard practice for evaluating neural SLAM models that have test-time online learning capabilities.

\subsection{Details about Hardware Setup for Benchmarking Experiments}\label{subsec:hardware}

Our experiments were primarily conducted on a GPU server equipped with two NVIDIA A6000 GPUs, each featuring 48 GB of memory. These resources were utilized for synthesizing perturbed noisy data and evaluating the robustness of RGB-D SLAM models. The operating system used was Ubuntu 22.04. Additionally, we tested the compatibility of our benchmarking code on a GPU server with four NVIDIA RTX6000 Ada GPUs, each with 48 GB of memory, and on a GPU server with two NVIDIA A100 GPUS, each with 40 GB of memory. 

It is important to note that the memory requirements of different SLAM methods vary based on the complexity of the perturbed RGB-D video sequences used for evaluation and the specific memory cost of each method. For instance, the CO-SLAM~\cite{coslam} model can run on a GPU with 12GB of memory. Meanwhile, only a GPU is required for all the SLAM methods evaluated in our study under each perturbed setting.

\subsection{Comparison with Existing SLAM Benchmarks}~\label{subsec:comparison_slam_benchmarks}
While acknowledging existing SLAM benchmarks~\cite{oxford-robotcar,Multi-Spectral,Burri25012016,Geiger2012CVPR,zhang2021multi,pfrommer2017penncosyvio,TUM-RGBD,zuniga2020vi,carlevaris2016university,Delmerico19icra,helmberger2022hilti,tian2023resilient,schubert2018tum,zhao2024subt}, our proposed noisy data synthesis pipeline and the instantiated benchmark \textit{Noisy-Replica} for RGB-D SLAM robustness evaluation offer several distinct advantages that can further advance the SoTA in SLAM evaluation:

\noindent\textbf{Unparalleled diversity and controllability.} With 124 perturbation settings and an extensive dataset comprising 1,000 long video sequences and nearly 2 million image-depth pairs, our tool offers unmatched diversity and controllability. Researchers can create highly customized and challenging test scenarios, exploring a wide range of real-world conditions and pushing the boundaries of SLAM algorithms. This extensive collection of perturbations allows for a comprehensive assessment of SLAM systems under diverse environmental conditions and sensor noise profiles.

\noindent\textbf{Scalability and fair comparison.} The large size of our dataset enables statistically significant evaluations and fair comparisons between different SLAM algorithms under diverse conditions. This scalability is crucial for robust benchmarking and identifying the strengths and weaknesses of various approaches. By providing a large and diverse testing ground, our tool facilitates unbiased comparisons and promotes the development of more reliable and generalizable SLAM solutions.

\noindent\textbf{Decoupled perturbation study.} Our pipeline facilitates the decoupled study of individual and mixed perturbations, providing valuable insights into the isolated and combined effects of various noise sources. This granular understanding is essential for developing targeted strategies to enhance SLAM robustness in complex environments. By disentangling the impact of individual noise sources, researchers can gain a deeper understanding of their specific effects on SLAM performance and design algorithms resilient to specific types of perturbations.

\noindent\textbf{Standardization.} Our pipeline introduces a systematic and standardized approach to generating noisy environments, ensuring consistency and reproducibility across different studies. This standardization is crucial for facilitating meaningful comparisons and advancing the field of SLAM research. By establishing a common framework for generating and evaluating SLAM datasets with perturbations, our tool promotes collaboration and accelerates the progress of the entire research community.

These unique features position our benchmark tool as a valuable resource for the SLAM community. By enabling comprehensive and standardized evaluations, our toolbox will accelerate the development of robust SLAM algorithms capable of handling the complexities of real-world environments.

\clearpage
\section{More Results and Discussions}\label{sec:additional-results}

\subsection{\textbf{More Benchmarking Analyses}}\label{subsec:additional-results-analyses}

\begin{figure*}[t!]
	\centering
\includegraphics[width=\textwidth]{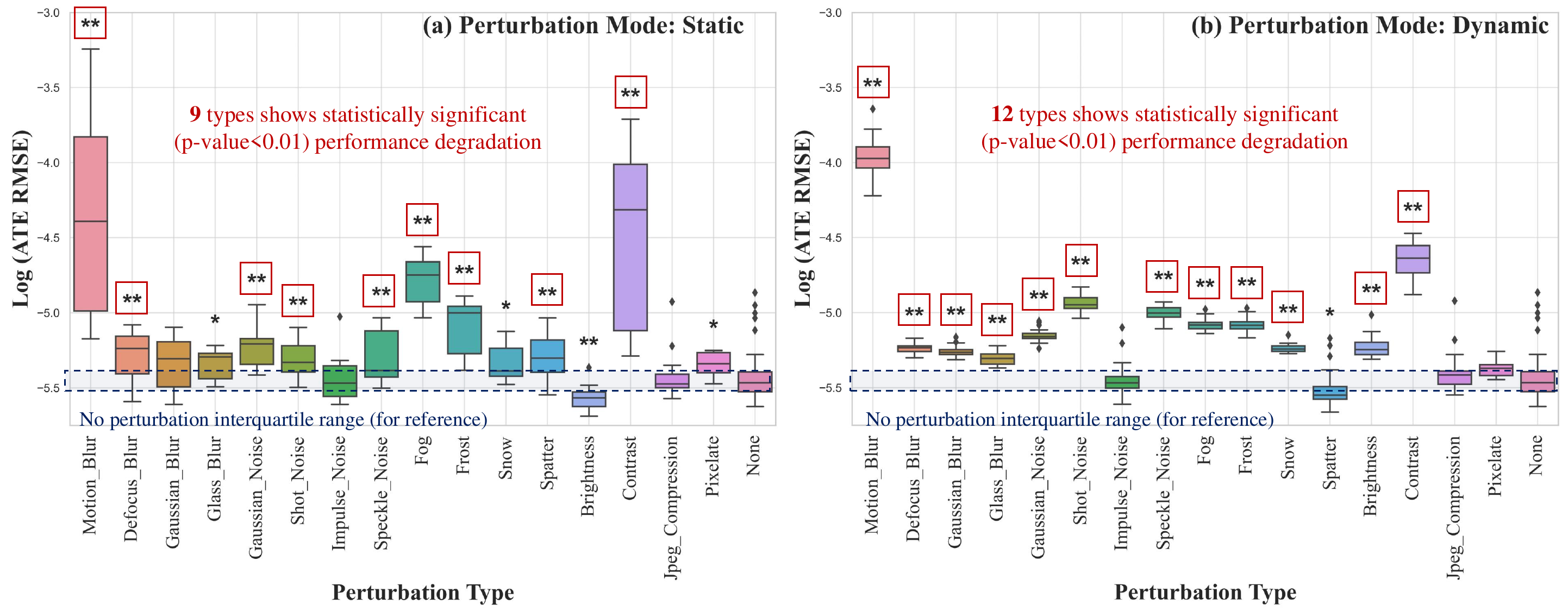}
	\caption{{Effect of each RGB imaging perturbation type
  on the trajectory estimation performance of SplaTAM-S~\cite{keetha2023splatam} model}, which shows the best overall performance under RGB imaging perturbations among all the benchmark methods. The t-test~\cite{kim2015t} is performed to compare the performance distribution between each perturbed setting and the perturbation-free setting (which is denoted as \textit{None} in the last column of each sub-figure). $**$ and $*$ indicate a significant distribution difference between the pair at the $0.01$ and $0.05$ significance level, respectively.}
	\label{fig:sensor-perturb-splatam-compare}
\end{figure*}

\noindent{\textit{\textbf{How well does the most robust SLAM model under RGB imaging perturbation, \textit{i.e.}, Splatam-S perform?}}} Even the top-performing SplaTAM-S model experiences a more substantial decrease in trajectory estimation accuracy under dynamic conditions, with statistically significant differences observed for most of the tested perturbation types (see Fig.~\ref{fig:sensor-perturb-splatam-compare}). Interestingly, increased brightness, while slightly beneficial under static conditions, leads to significant errors under dynamic conditions for SplaTAM-S.

\begin{figure*}[t]
	\centering
 \includegraphics[width=0.495\textwidth]{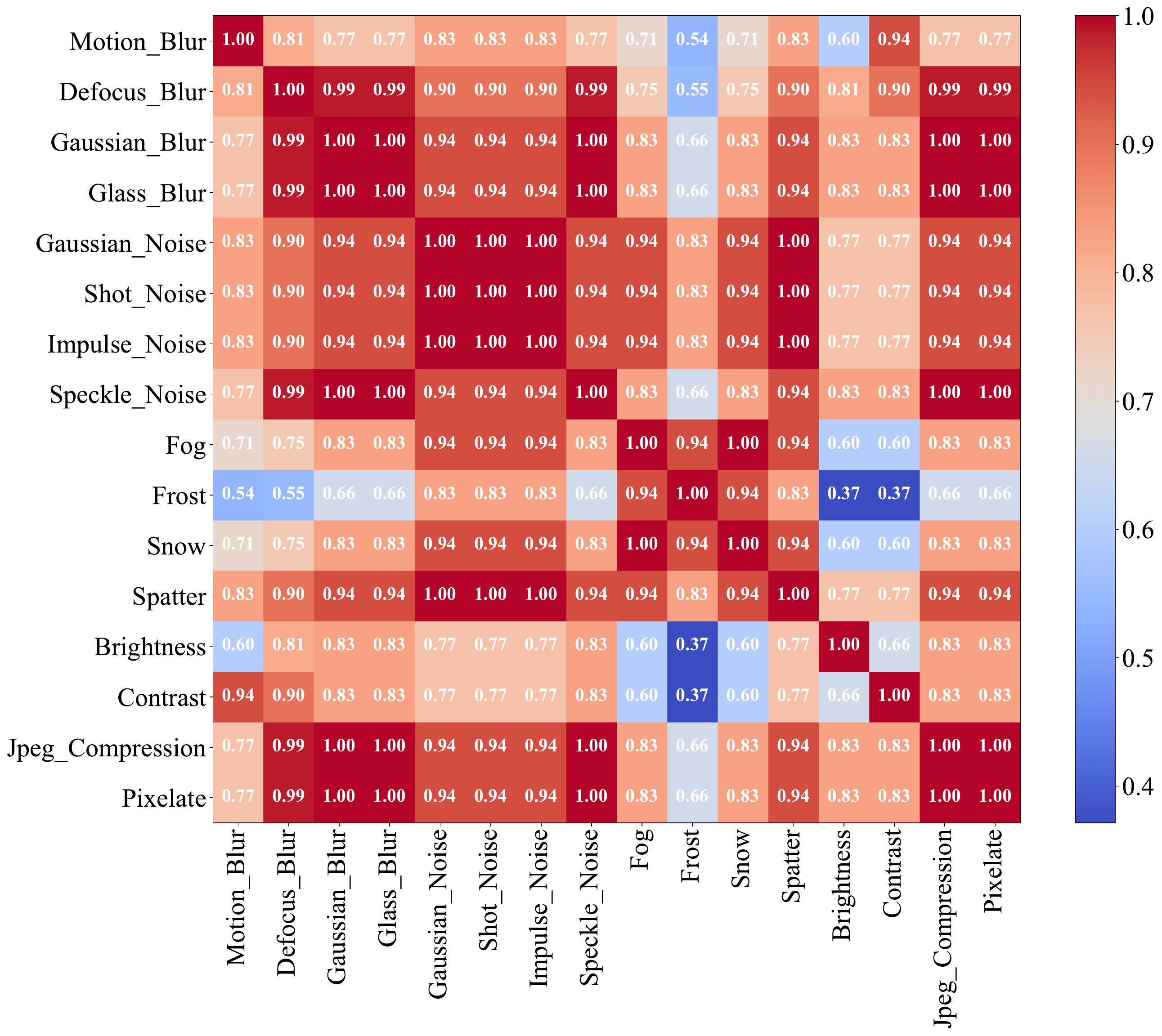} 
 \hfill 
\includegraphics[width=0.495\textwidth]{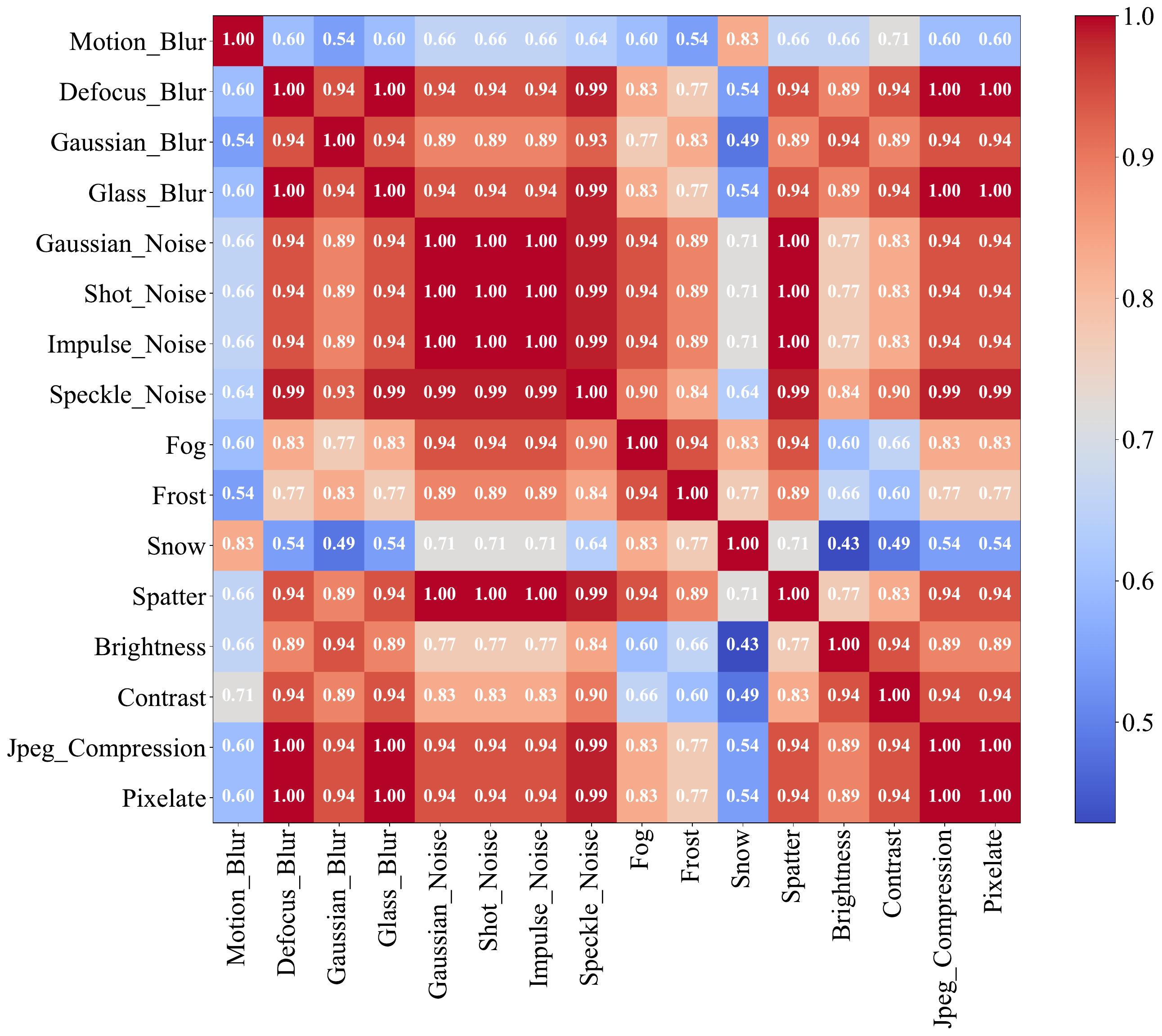} 
	\caption{\textbf{Correlation of perturbed performance (ATE) of multi-modal (RGB-D) SLAM models across different image perturbation types} under static (\textbf{Left}) and dynamic (\textbf{Right}) perturbation mode. The pair-wise correlation strength is quantified via Spearman’s rank correlation coefficient~\cite{spearman1961proof}.}
	\label{fig:correlation_perturbed_image_ate-type}
\end{figure*}

\textbf{\textit{Is there a correlation in the performance after perturbation among different image perturbation types?}} 
In Fig.~\ref{fig:correlation_perturbed_image_ate-type}, a strong correlation is observed in the combined perturbed performance vector of all evaluated RGB-D SLAM models for the majority of perturbation types. This finding suggests that the models' performance remains consistent across certain perturbation scenarios. Additionally, the correlation suggests the presence of underlying similarities in the effects of some sub-categories of image perturbation types, \textit{e.g.}, noise effects.

\begin{figure*}[t]
	\centering
\includegraphics[width=0.44\textwidth]{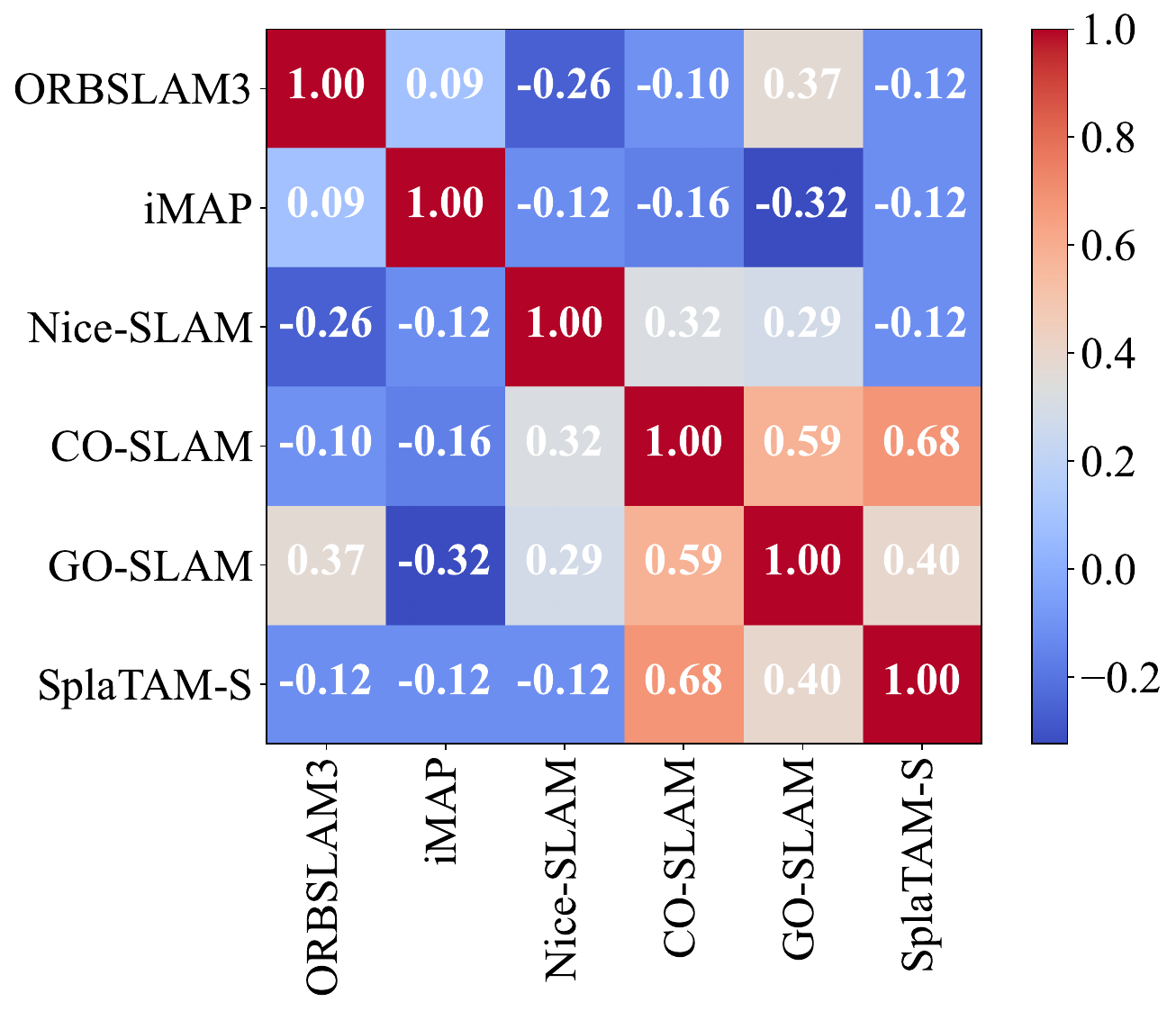} 
\includegraphics[width=0.44\textwidth]{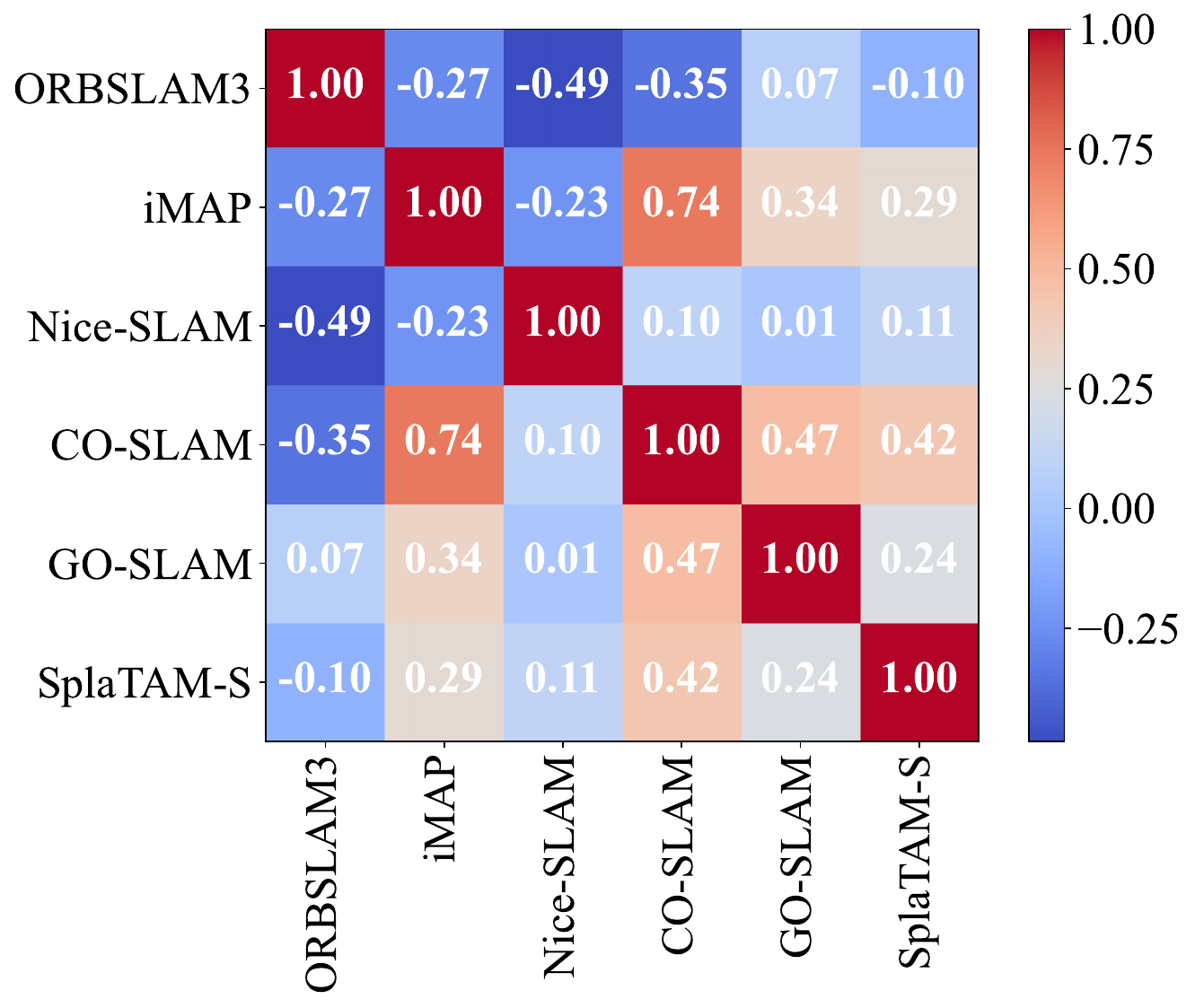} 

	\caption{\textbf{Correlation of perturbed performance (ATE) across different multi-modal (RGB-D) SLAM models} under static (\textbf{Left}) and dynamic (\textbf{Right}) image perturbation. The pair-wise correlation strength is quantified via Spearman’s rank correlation coefficient~\cite{spearman1961proof}.}
	\label{fig:correlation_perturbed_image_ate-method}
\end{figure*}

\textbf{\textit{Is there a correlation in the performance under RGB image perturbation among different methods?}} In Fig.~\ref{fig:correlation_perturbed_image_ate-method}, a weak correlation is observed in the combined perturbed performance vector, which encompasses sixteen image perturbation types across six SLAM models with the RGB-D input setting. This suggests a large divergence in the distribution of perturbed performance among the different SLAM models.

\textbf{\textit{How do image perturbations influence the mapping quality?}} We follow the mapping quality evaluation protocol in~\cite{coslam} to assess 3D reconstruction using Accuracy (ACC) [cm], Completion (Comp.) [cm], and Completion Ratio (Comp. R.) [\%] with a 5 cm threshold. Table~\ref{tab:3d-metric} details the definition for each of these metrics. Note that only certain dense SLAM models can produce 3D reconstruction results for further evaluation of mapping quality. In Fig.~\ref{tab:mapping-coslam}, we evaluate the impact of image perturbations on the mapping quality of the CO-SLAM~\cite{coslam} model, which shows a strong robustness to most of the image-level corruptions. The results reveal a direct correlation between perturbation severity and both 3D reconstruction error and completion error. Specifically, the clean setting achieves the highest accuracy (2.08 cm) and the lowest completeness score (2.17 cm), while the high perturbation severity setting exhibits the highest errors in ACC (2.39 cm) and completion (2.89 cm). Overall, our analysis shows that increasing severity levels of perturbation lead to larger errors in the reconstructed 3D map.

\begin{table}[t]
\caption{\textbf{Definitions of 3D metrics for evaluation of mesh reconstruction quality} of the reconstructed 3D mesh $P$ when given the ground-truth 3D mesh $Q$ (in the scale of meter [m]). We follows the 3D reconstruction metrics defined in the CO-SLAM~\cite{coslam} paper.}
\begin{center}
\resizebox{0.8\textwidth}{!}{
\begin{tabular}{l|c}
\toprule \toprule
\textbf{3D Reconstruction Metric} & \textbf{Definition} \\
\midrule
Accuracy (ACC) & $\frac{1}{|P|} \sum_{p \in P} \left(\min_{q \in Q} {{||p - q||{}^{2}}}\right)$ \\
Completion (Comp.) & $\frac{1}{|Q|} \sum_{q \in Q} \left(\min_{p \in P} {{||p - q||{}^{2}}}\right)$ \\
Completion Ratio (Comp. R.) & $\frac{1}{|Q|} \sum_{q \in Q} \left(\min_{p \in P} {{||p - q||{}^{2}}} \leq {0.05}\right)$ \\
\bottomrule \bottomrule
\end{tabular}\label{tab:3d-metric}}
\end{center}
\end{table}

\begin{table}[t]
\caption{\textbf{Effects of RGB imaging perturbation on 3d reconstruction (mapping) quality} for CO-SLAM~\cite{coslam} model.}
\label{tab:mapping-coslam}
\centering 
\setlength{\tabcolsep}{4.0mm}
\resizebox{0.8\textwidth}{!}{
\begin{tabular}{l|c|ccc|c}
    \toprule \toprule   
    \multirow{2}{*}{\textbf{Metrics}} &     {\textbf{Clean}}
     &  {\textbf{Low}}   & {\textbf{Middle}} & 
  {\textbf{High}}  & {\textbf{Perturb.}}     \\  
 &  \textbf{Mean}& \textbf{Severity} & \textbf{Severity} & \textbf{Severity} & \textbf{Mean}  \\ \midrule
   ACC$\downarrow$ [cm] & $\textbf{2.08}$ & $2.11$ & $2.12$ & $2.39$ & $2.21$ 
    \\
    Comp.$\downarrow$ [cm] & $\textbf{2.17}$ & $2.19$ & $2.20$ & $2.89$  & $2.43$ 
    \\
    Comp. R.$\uparrow$ [\%] & $\textbf{93.13}$ & $93.07$ & $93.04$ & $92.34$  & $92.82$ \\
    \bottomrule \bottomrule
    \multicolumn{6}{l}{{\textbf{1}) The setting with the best performance for each metric is in \textbf{bold}.}}\\
    \multicolumn{6}{l}{\textbf{2}) We compare the performance under no perturbation (clean) and  }\\
    \multicolumn{6}{l}{static image perturbations with different severity levels.}
    \end{tabular}
}
\end{table}

\clearpage
\subsection{More Discussions}\label{subsec:additional-results-discussion}

\begin{figure}[t!]
\includegraphics[width=\textwidth]{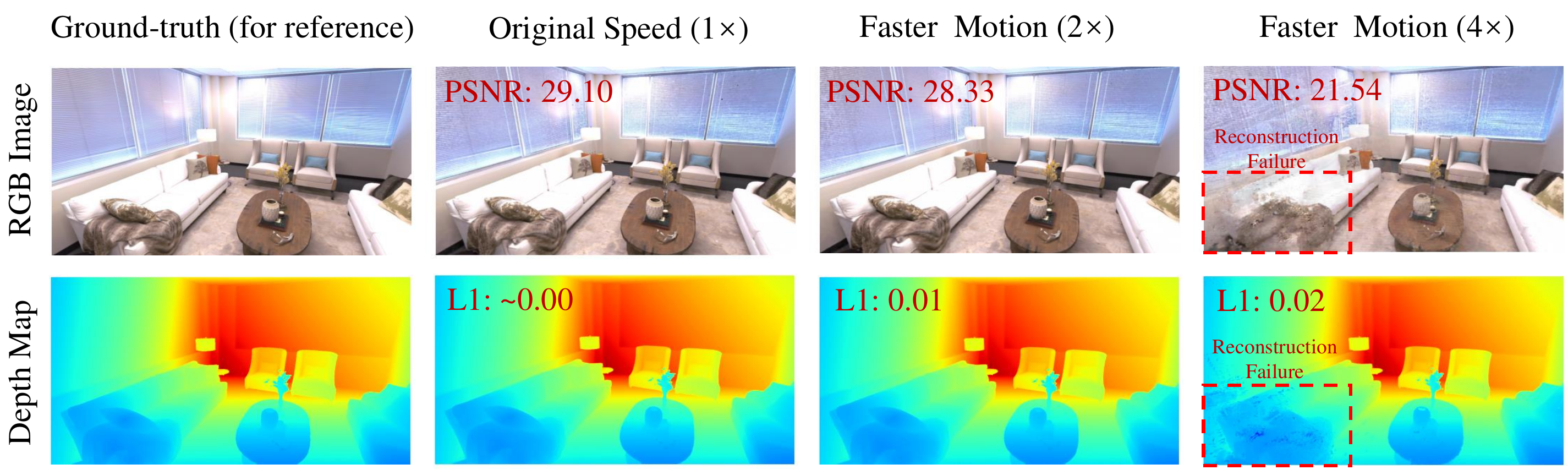} 
	\caption{{Effect of {faster motion}}  \xxh{on the 2D reconstruction losses} of RGB images (\textbf{Left}) and depth maps (\textbf{Right}), which are measured via PSNR and Depth L1 loss, for SplaTAM-S~\cite{keetha2023splatam}.}
	\label{fig:recon-fast-motion}
 \vspace{3mm}
     \centering
     \includegraphics[width=\textwidth]{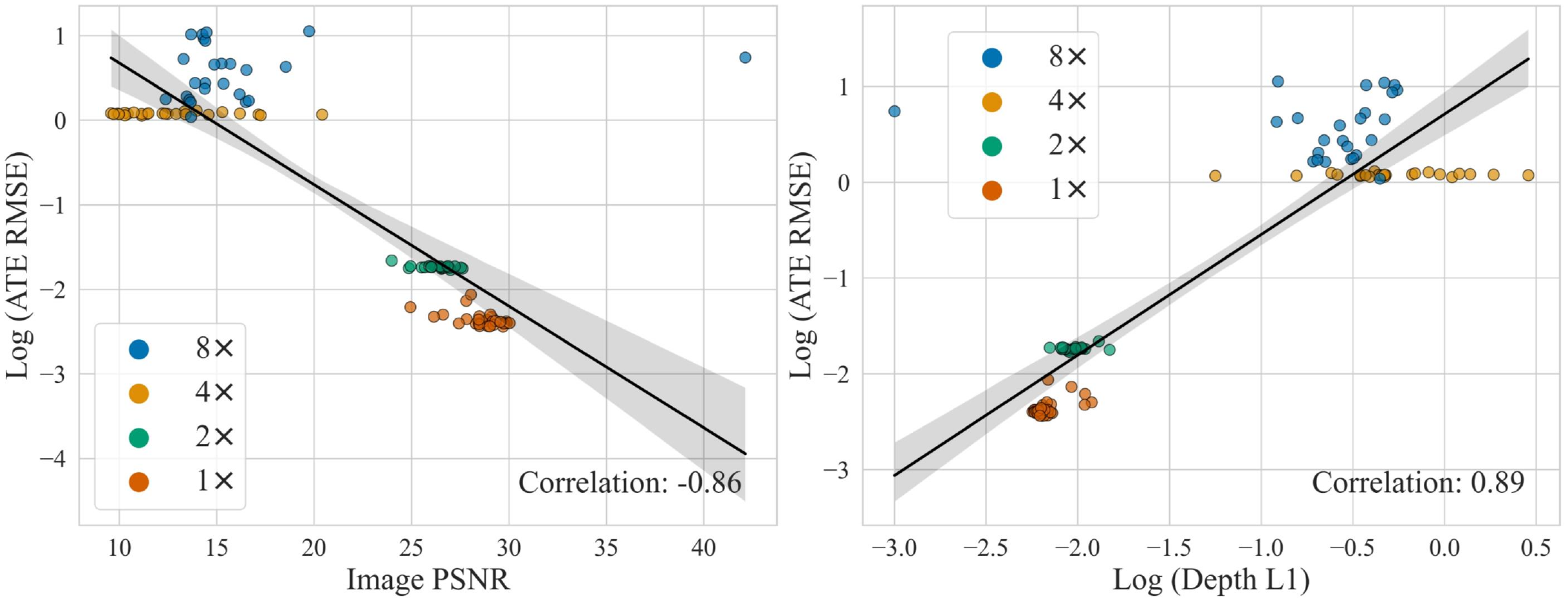} 
	\caption{Correlation between ATE (logarithm form) and \xxh{the 2D reconstruction losses of RGB images (\textbf{Left}) and depth maps (\textbf{Right}), which are used for model optimization,} under faster motion effects for SplaTAM-S~\cite{keetha2023splatam}. Pearson correlation coefficient~\cite{cohen2009pearson} is reported in the bottom-right corner. }\vspace{-4mm}
	\label{fig:correlation-fast-motion}\centering
\end{figure}

\noindent\textbf{\xxh{There exists  SLAM models  that can {perceive} perturbed observations.}}
We conduct a case study to explore the ability of SplaTAM-S model to perceive the severity of perturbations. Notice that SplaTAM-S optimizes the map and 3D pose by minimizing the 2D reconstruction loss for the RGB and depth maps during inference. In Fig.\ref{fig:recon-fast-motion}, we assess SplaTAM-S's response to different severities of faster motion perturbations. The results demonstrate that more severe perturbations lead to poorer reconstruction quality of RGB-D images. Moreover, in Fig.\ref{fig:correlation-fast-motion}, we observe a strong correlation between the accuracy of the final trajectory estimation and the RGB-D reconstruction loss. This indicates that when the model produces a larger reconstruction loss for a certain sensor stream, it is likely that the trajectory estimation is also inaccurate.  While this doesn't provide exact localization, it serves as a valuable indicator of potential observation degradation and model failure. This suggests SLAM systems could self-monitor performance using internal indicators, enabling real-time failure mitigation in safety-critical applications.

\begin{figure*}[t]
	\centering
 \vspace{1mm}
\includegraphics[width=\textwidth]{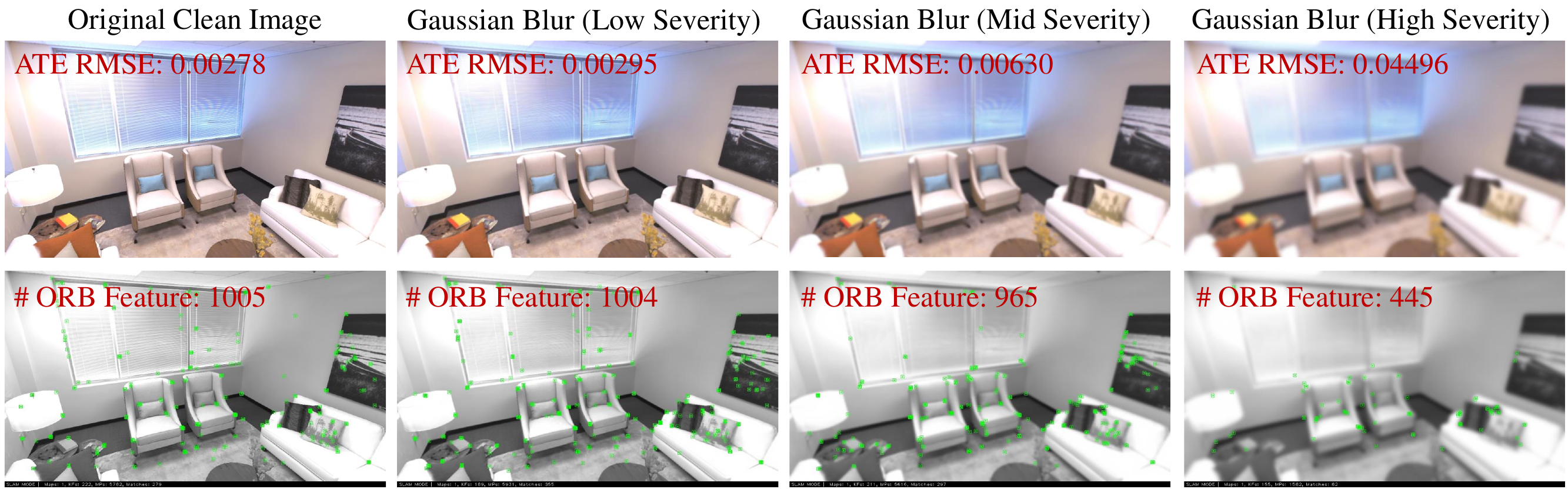} 
	\caption{\textbf{Effect of \textit{Gaussian Blur} image-level perturbation under different severity (\textbf{Top}) on the quality of detected ORB features} (\textbf{Bottom}), which are marked as green dots, for the classical SLAM model ORB-SLAM3~\cite{orbslam3}. We report the average trajectory accuracy via ATE RMSE and the average number of ORB features detected in various perturbed settings. }
	\label{fig:orbslam-gaussian-blur} 
\end{figure*}

\begin{figure*}[t]
	\centering
\includegraphics[width=0.48\textwidth]{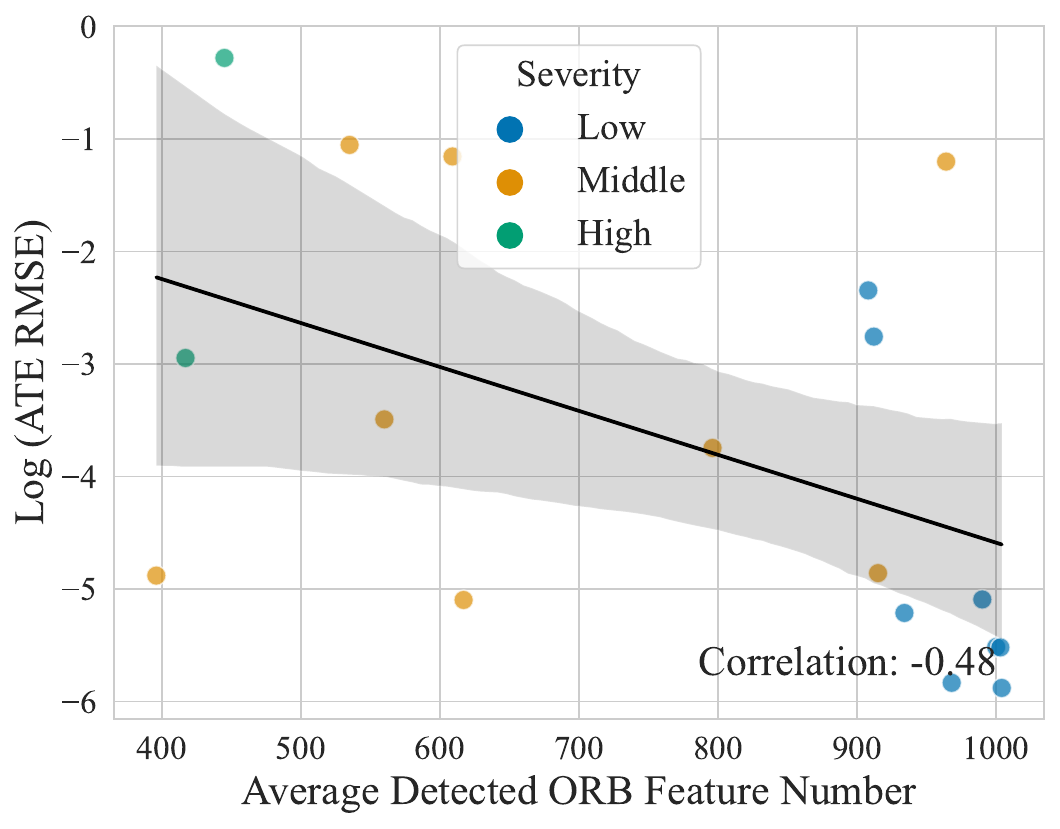} 
\includegraphics[width=0.48\textwidth]{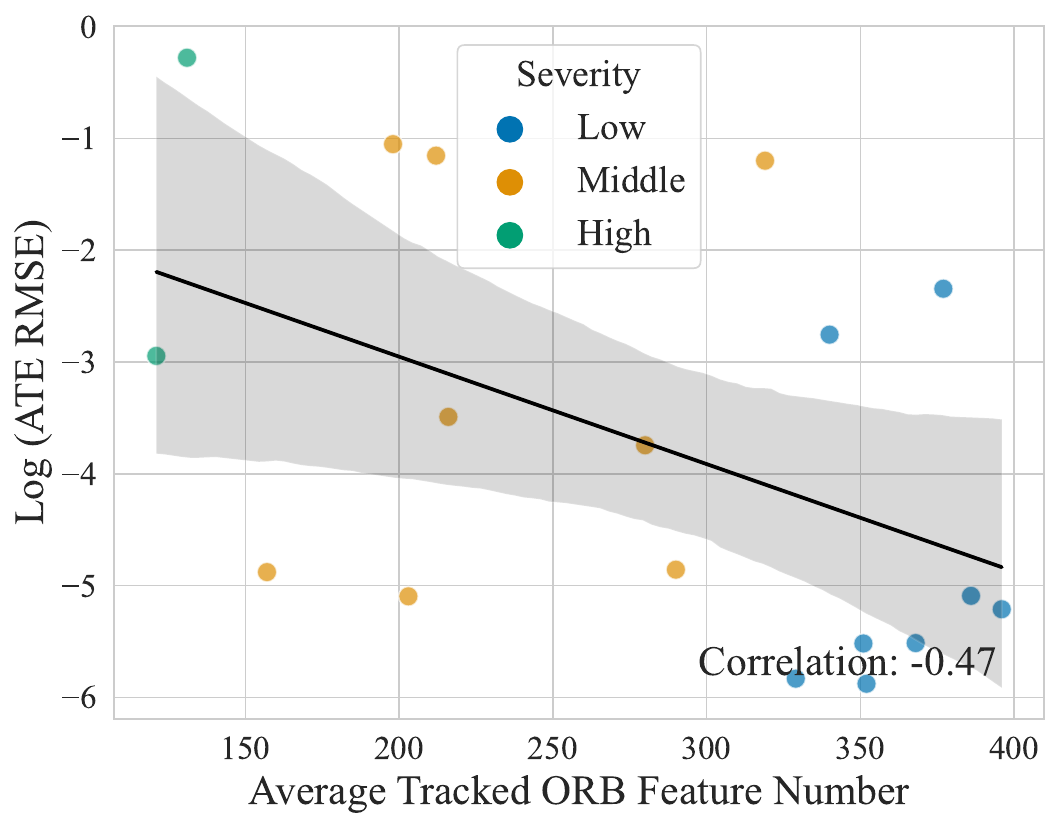} 
	\caption{\textbf{Correlation between trajectory estimation accuracy} \textbf{and the average number of detected} (\textbf{Left}) \textbf{and tracked} (\textbf{Right}) \textbf{ORB features for ORB-SLAM3}~\cite{orbslam3} model (RGB-D setting) under different severity level of Gaussian Blur image-level perturbation. We report the Pearson correlation coefficient~\cite{cohen2009pearson} at the bottom right corner. While the correlation coefficient does not indicate a significant linear correlation, there is a noticeable trend of increased trajectory estimation when fewer ORB features are detected or tracked.}
	\label{fig:correlation-orb-accuracy}
\end{figure*}

In addition to the neural SLAM model SplaTAM-S~\cite{keetha2023splatam}, we explore the ability of the classical SLAM model, \textit{i.e.}, ORB-SLAM3~\cite{orbslam3}, to `perceive' perturbation severity. Specifically, we aim to explore the correlation between the quality of ORB feature detection and the resulting overall performance.
In Fig.~\ref{fig:orbslam-gaussian-blur}, we present qualitative comparisons of the ORB feature detection results of ORB-SLAM3 under the influence of varying severity levels of Gaussian blur image-level perturbations. It is evident that more severe perturbations result in a lower number of detected ORB features. In addition, Fig.\ref{fig:correlation-orb-accuracy} depicts the correlation between the number of (detected or matched) ORB feature descriptors and the accuracy of trajectory estimation. With increasing severity levels, a noticeable reduction in the number of ORB features is observed, accompanied by a subsequent rise in trajectory estimation error, \textit{i.e.}, ATE. This trend indicates that the deterioration of detected feature descriptors could serve as an informative indicator for identifying degraded and anomalous observations. Furthermore, it has the potential to provide a rough estimation of the overall performance of ORB-SLAM3 in situations where ground-truth annotation is unavailable.


\clearpage
\section{Qualitative Results}\label{sec:qualitative_results}

\subsection{Qualitative Results of SLAM Model Performance under Perturbations}\label{subsec:qualitative_results-SLAM}

This section presents qualitative results of trajectory estimation and 3D reconstruction in SLAM models, highlighting both successful and failure cases under specific perturbations.

\begin{figure*}[ht!]
	\centering
\includegraphics[width=\textwidth]{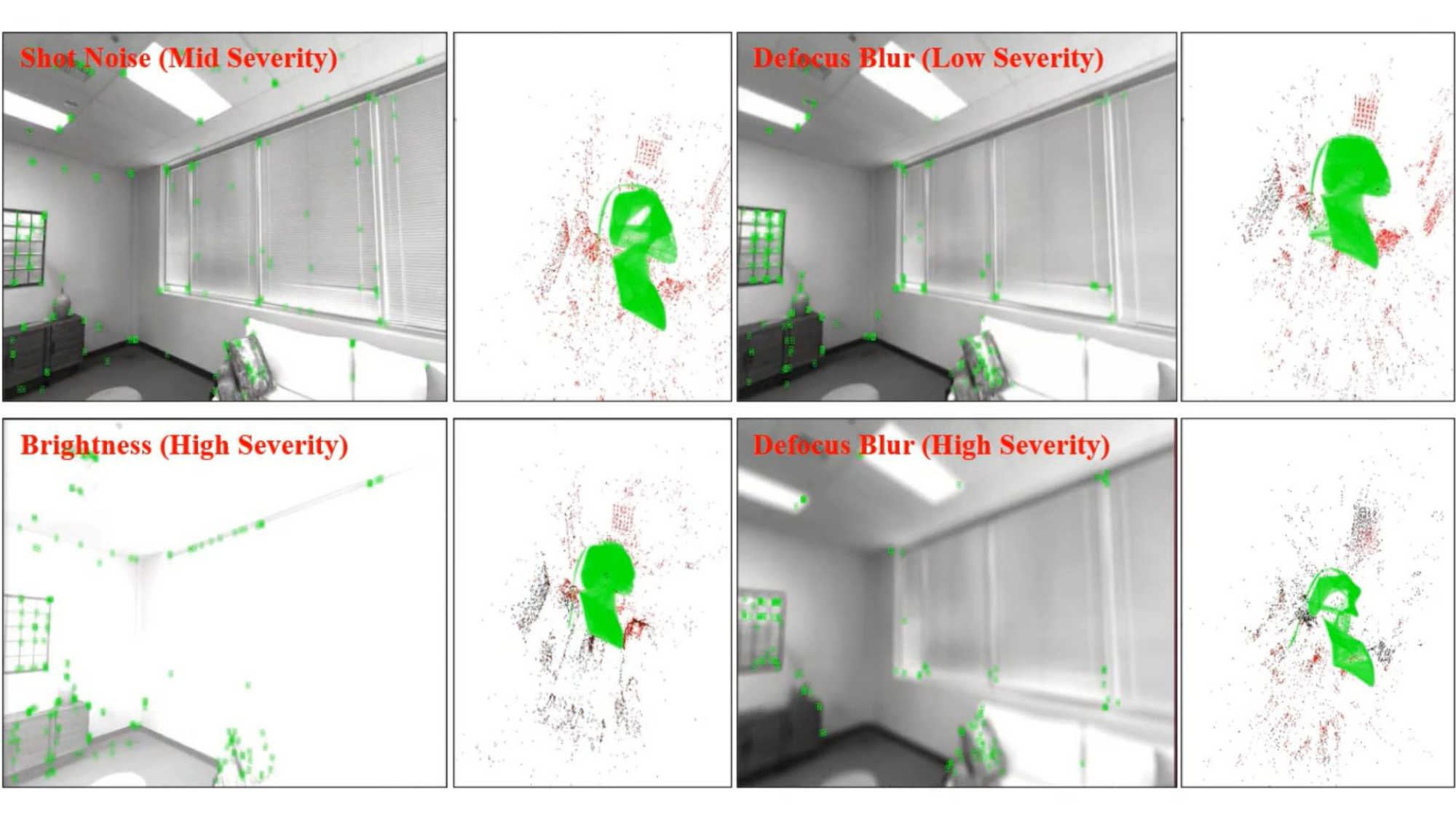} 
\caption{\textbf{Qualitative results of the successful cases of ORB-SLAM3 model~\cite{orbslam3} with  RGB-D input.}}
\label{fig:orbslam3-succeed} 
\end{figure*}

\begin{figure*}[ht!]
	\centering
\includegraphics[width=\textwidth]{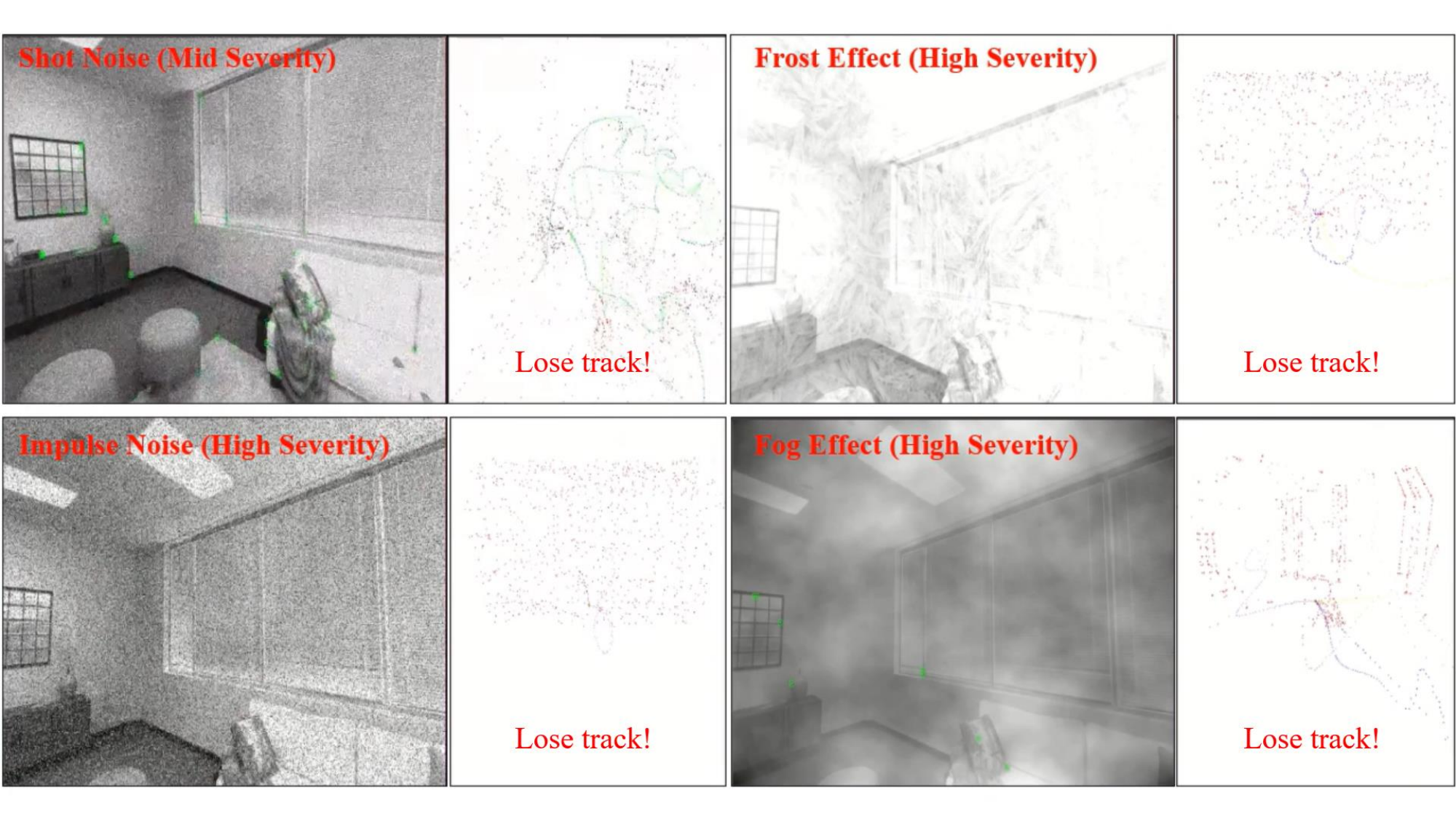} 
\caption{\textbf{Qualitative results of the failure cases of ORB-SLAM3 model~\cite{orbslam3} with  RGB-D input.}}
\label{fig:orbslam3-fail} 
\end{figure*}

\noindent\textbf{ORB-SLAM3.} Fig.~\ref{fig:orbslam3-succeed} demonstrates the resilience of the ORB-SLAM3~\cite{orbslam3} model to certain image corruptions, \textit{e.g.}, brightness changing and defocus blur. However, we observed that noise-related perturbations can cause the failure of ORB feature detection of the ORB-SLAM3 model, resulting in complete loss of tracking, as depicted in Fig.~\ref{fig:orbslam3-fail}.

\begin{figure*}[ht!]
	\centering
\includegraphics[width=\textwidth]{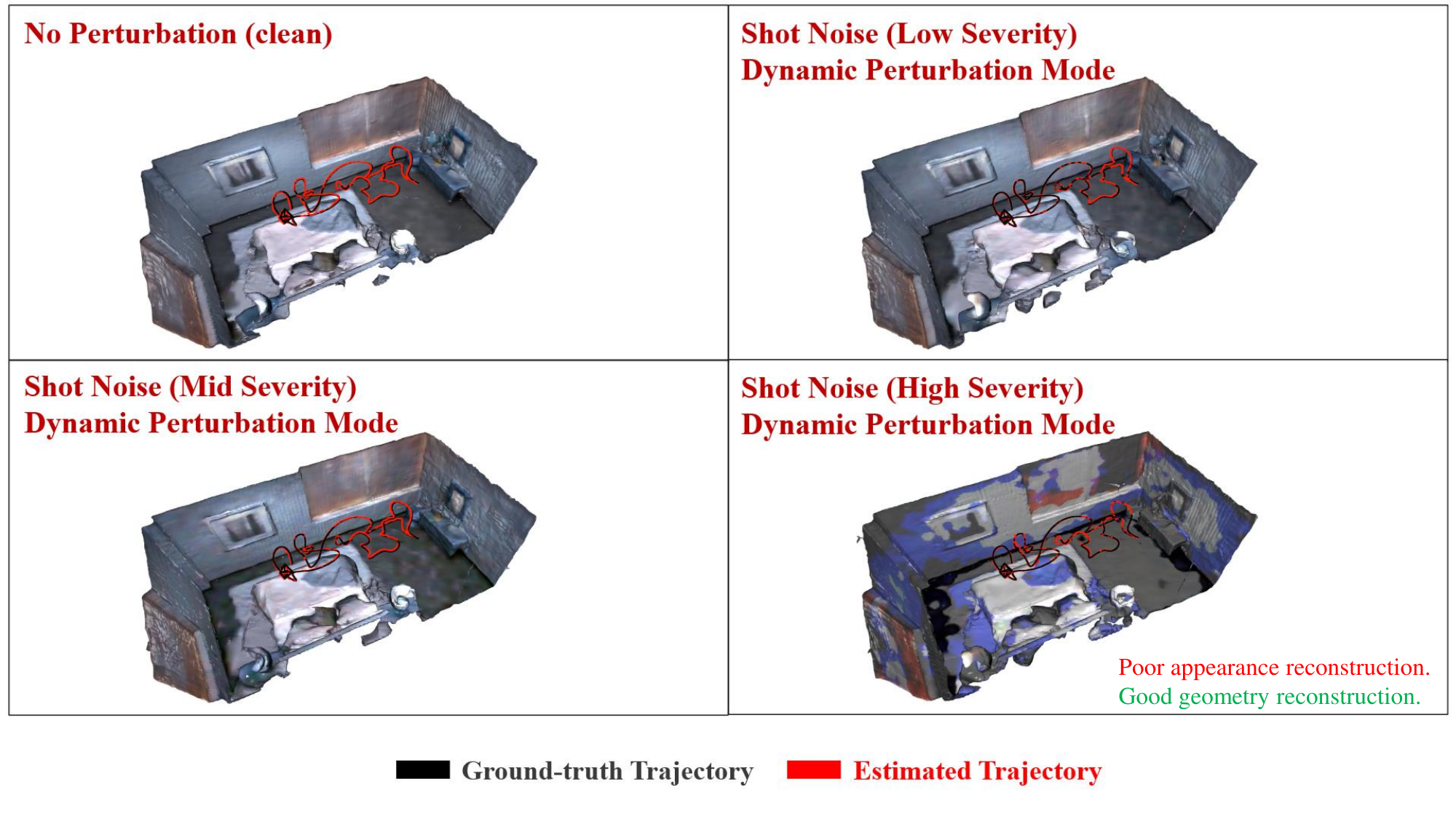} 
\caption{\textbf{Qualitative results of successful cases of Nice-SLAM model~\cite{niceslam} with RGB-D input.}}
\label{fig:niceslam-succeed} 
\end{figure*}

\begin{figure*}[ht!]
	\centering
\includegraphics[width=\textwidth]{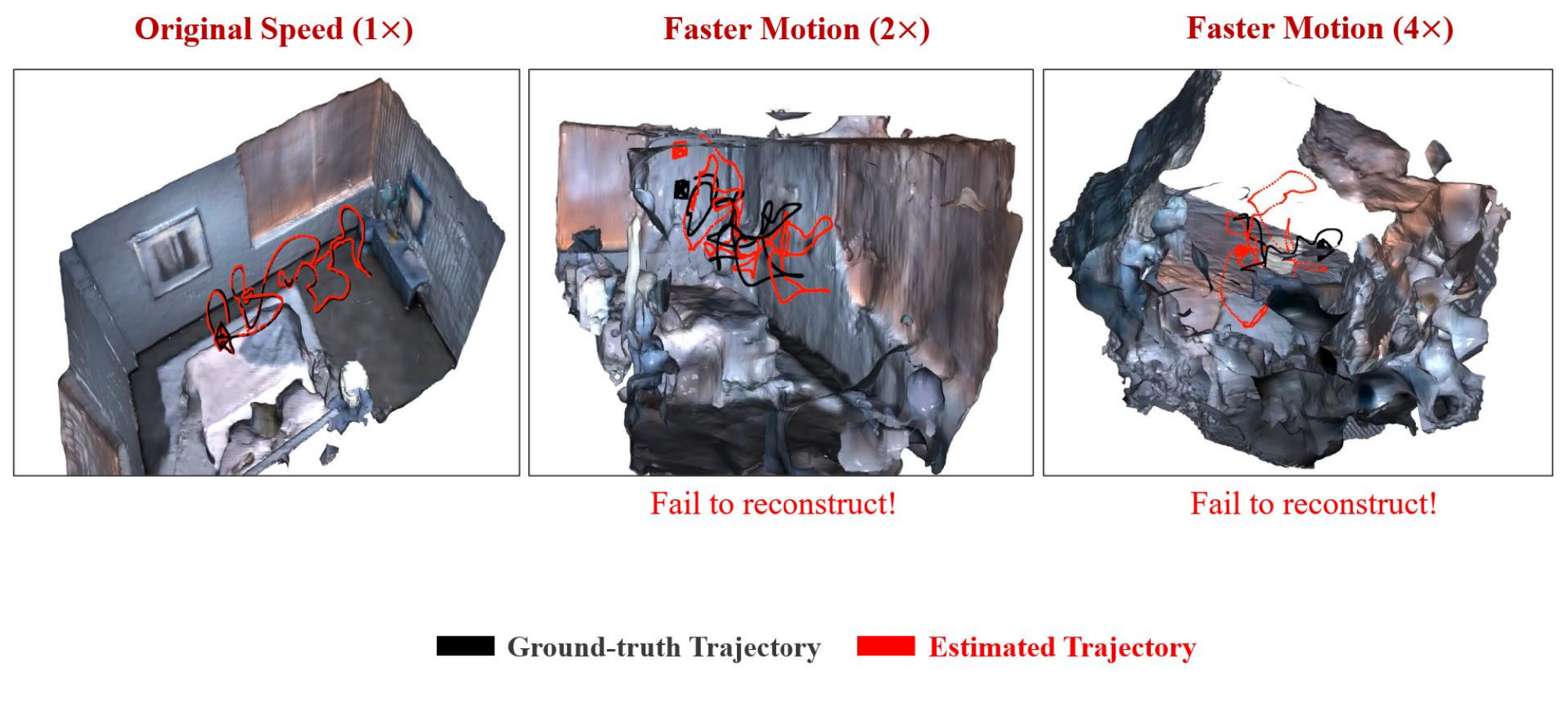} 
\caption{\textbf{Qualitative results of the failure cases of Nice-SLAM model~\cite{niceslam} with  RGB-D input.}}
\label{fig:niceslam-fail} 
\end{figure*}

\noindent\textbf{Nice-SLAM.} Fig.~\ref{fig:niceslam-succeed} showcases the 3D reconstruction and trajectory estimation results of the Nice-SLAM~\cite{niceslam} model under varying levels of shot noise perturbation on the RGB image. Nice-SLAM consistently produces high-quality geometry reconstructions even when subjected to high severity levels of shot noise, which we attribute to the nearly error-free, unperturbed depth map aiding geometry reconstruction. However, we observe that the model struggles to accurately predict and reconstruct appearance details. Consequently, as the noise in the RGB images intensifies, color reconstruction quality diminishes. Furthermore, Fig.~\ref{fig:niceslam-fail} highlights the complete failure of the Nice-SLAM model in reconstructing 3D geometry and maintaining tracking under rapid motion.

\begin{figure*}[ht!]
	\centering
\includegraphics[width=\textwidth]{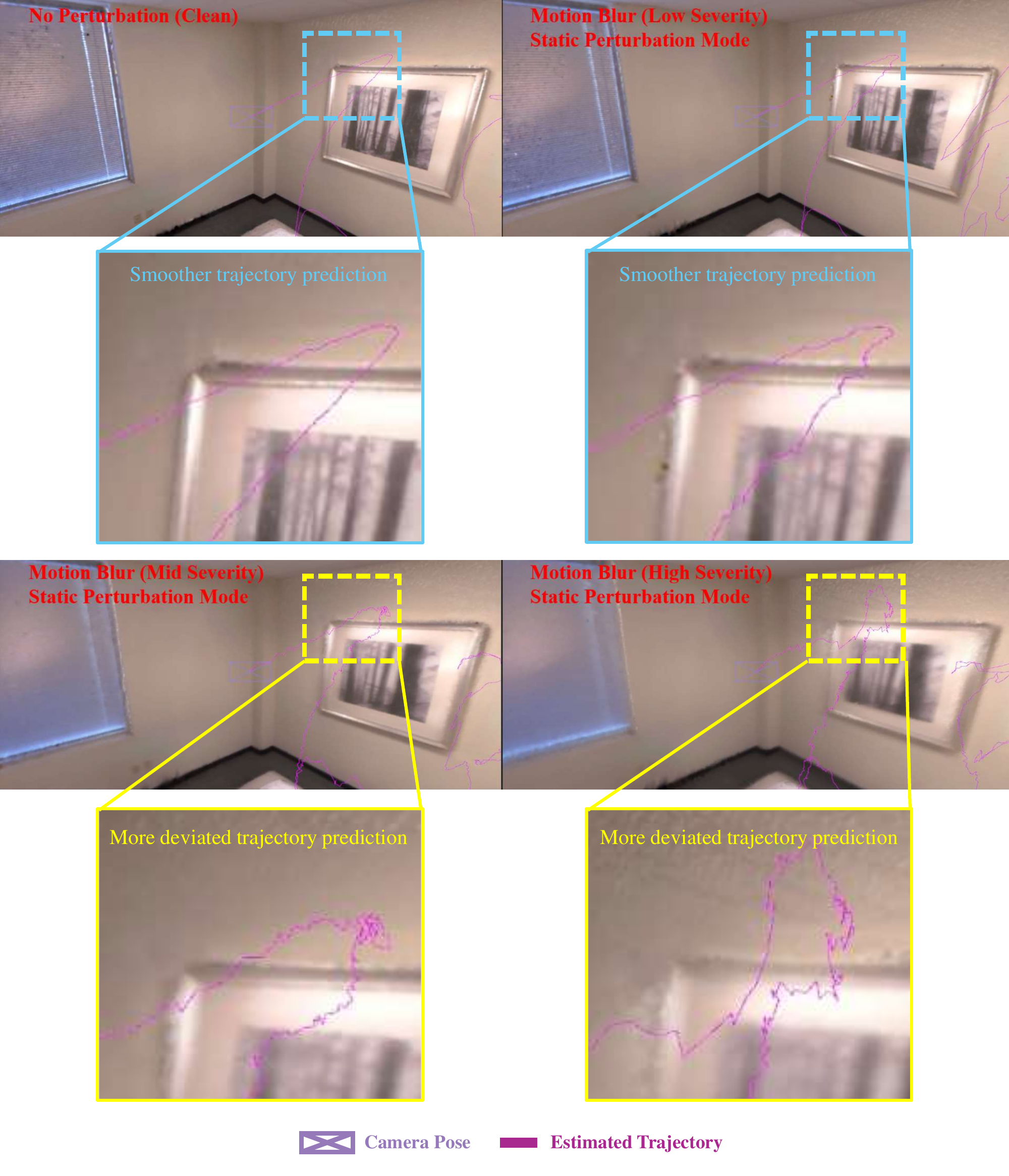} 
\caption{\textbf{Qualitative results of successful cases of SplaTAM-S model~\cite{keetha2023splatam} with  RGB-D input.}}
\label{fig:splatam-succeed} 
\end{figure*}

\begin{figure*}[ht!]
	\centering
\includegraphics[width=\textwidth]{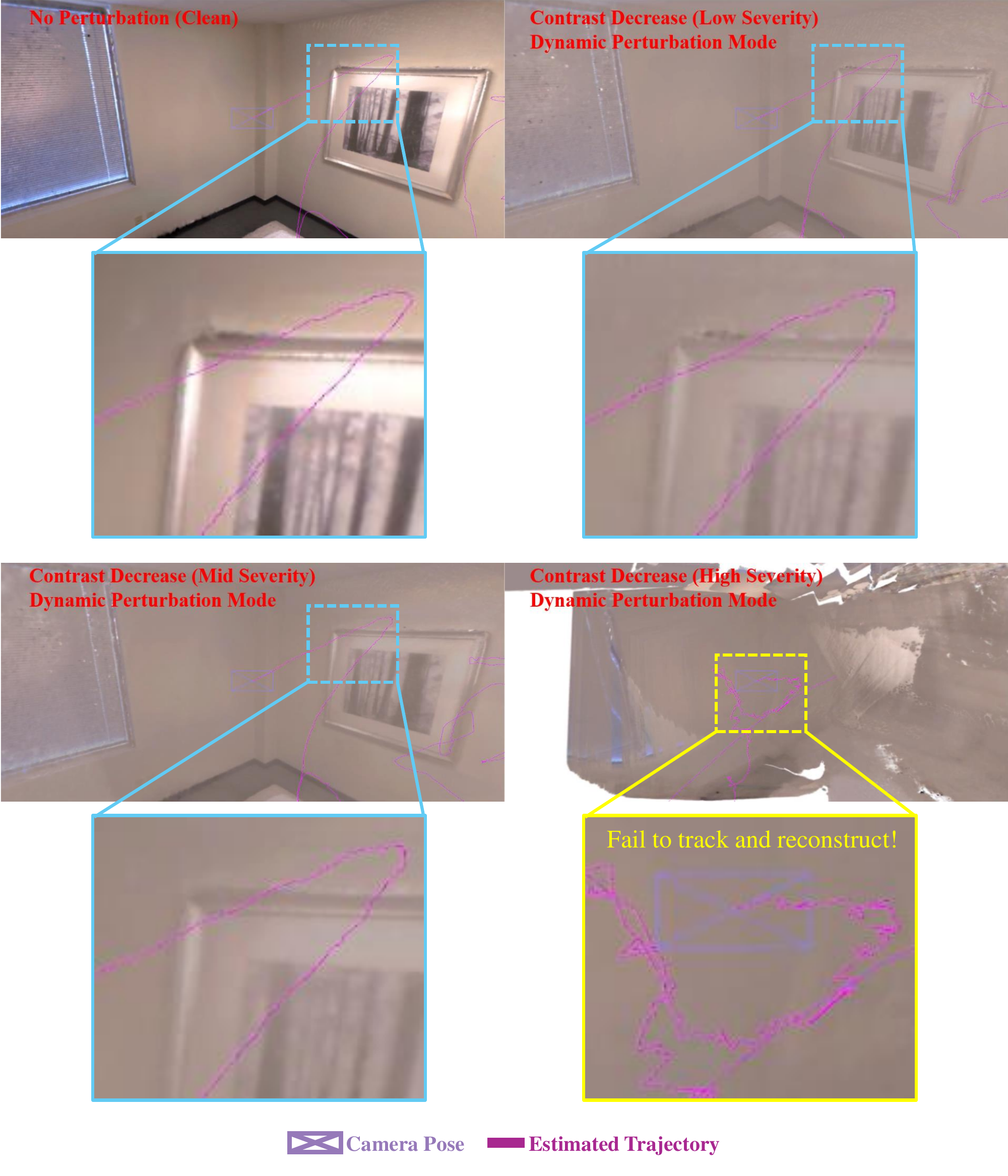} 
\caption{\textbf{Qualitative results of the failure cases of SplaTAM-S model~\cite{niceslam} with RGB-D input.}}
\label{fig:splatam-fail} 
\end{figure*}

\begin{figure*}[ht!]
	\centering
\includegraphics[width=\textwidth]{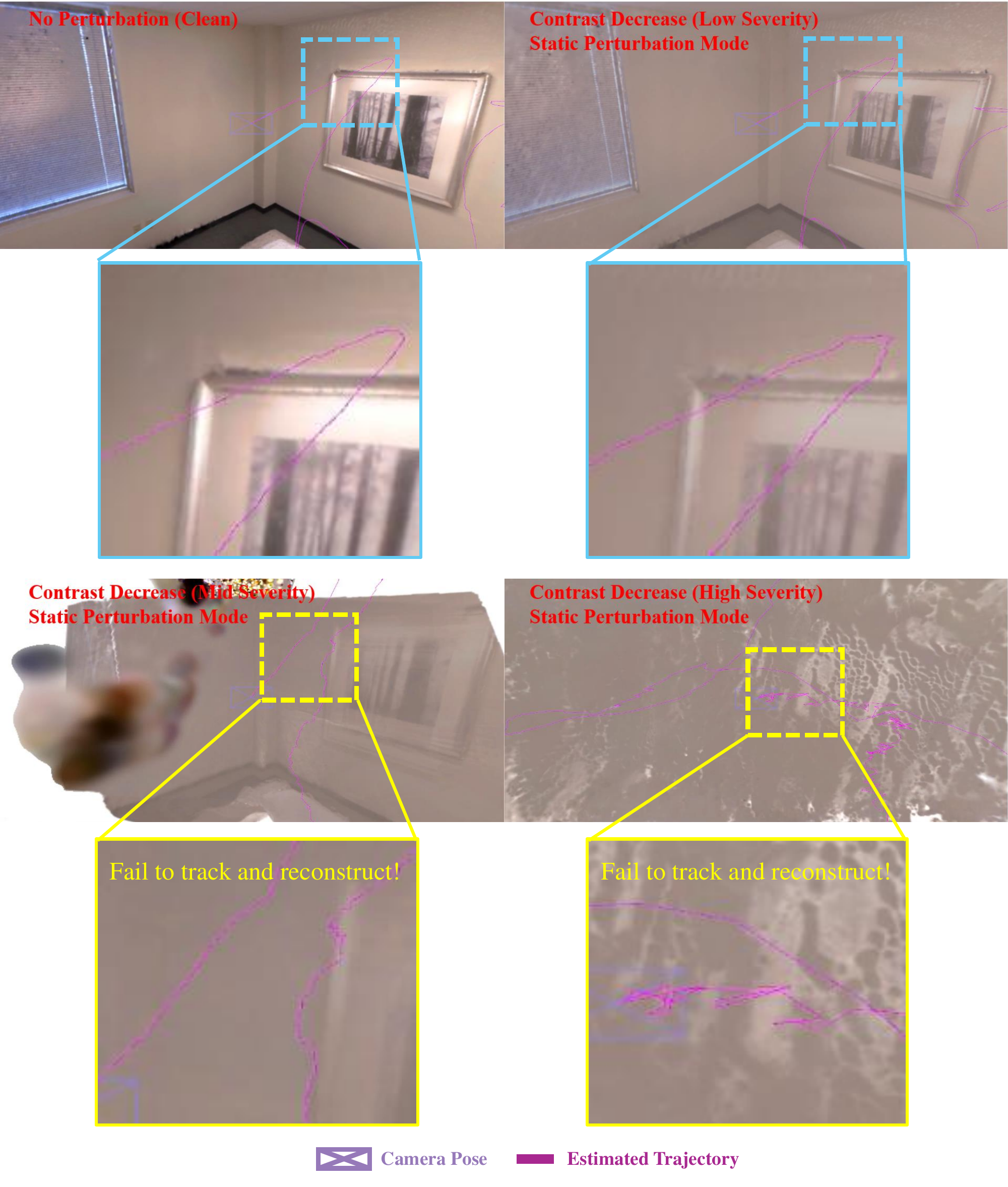} 
\caption{\textbf{Qualitative results of the failure cases of SplaTAM-S model~\cite{niceslam} with  RGB-D input.}}
\label{fig:splatam-fail-2} 
\end{figure*}

\noindent\textbf{SplaTAM-S.} Fig.~\ref{fig:splatam-succeed} presents the qualitative results of the SplaTAM-S~\cite{keetha2023splatam} model under different severity levels of motion blur image-level perturbations. The trajectory estimation reveals that, in the absence of perturbation or with low levels of motion blur, the model produces smooth trajectories. However, as perturbation severity increases to a moderate or high level, the predicted trajectory exhibits more deviations. Notably, the 3D reconstruction consistently maintains high quality despite the increasing blurring caused by observation degradation. In addition, Fig.\ref{fig:splatam-fail} and Fig.\ref{fig:splatam-fail-2} depict failure instances of the SplaTAM-S~\cite{keetha2023splatam} model. These failures occur when subjected to varying levels of contrast decrease image-level perturbation under both static and dynamic perturbation modes. Higher severity levels of perturbation result in complete tracking loss and reconstruction failure.

To provide a better viewing experience, we kindly refer you to the \href{https://youtu.be/jNM94naSPXA}{video demo}.

\subsection{Video Demo}\label{subsec:video-demo}

We provide a video demo on YouTube (\url{https://youtu.be/jNM94naSPXA}) showcasing the visualization of synthesized noisy data for evaluating SLAM robustness, including both sensor and trajectory perturbations. The video also displays qualitative per-frame prediction results, such as trajectory estimation and 3D reconstruction, of advanced SLAM models under these perturbations, highlighting both successful and failed cases.

\clearpage

\clearpage
\section{Potential Other Directions to Explore}\label{sec:more future work}
Towards more robust and deployable SLAM, we present additional potential research avenues:

\noindent\textbf{Robustness evaluation in unbounded 3D scene.} Our work does not encompass the robustness of SLAM systems in unbounded scenes, such as outdoor environments~\cite{dosovitskiy2017carla,zhao2023subt}. Investigating the robustness of SLAM systems in such scenarios holds significant potential for future research. It can contribute to a better understanding of the practical applicability and generalizability of SLAM systems in complex scenes.

\noindent\textbf{Computationally-efficient robustness evaluation.} Our findings have identified discernible indicators within certain SLAM models that can reflect degraded observations, \textit{i.e.}, the reconstruction quality of RGB images and depth maps for the SplaTAM-S~\cite{keetha2023splatam} model. Future research could explore leveraging and designing robustness indicators to evaluate the robustness of SLAM systems more efficiently. By incorporating such indicators, we have the potential to enable unsupervised performance evaluation of SLAM, especially in scenarios where obtaining ground-truth annotations is challenging or costly.

\noindent\textbf{Real-world robustness evaluation.} While our work primarily focuses on synthesis-based robustness analysis, we recognize the value of real-world verification and validation for SLAM systems. Conducting extensive field tests in more challenging environments~\cite{SubT}, where SLAM systems are subjected to agile locomotion types~\cite{kaufmann2023champion}, would provide empirical validation of simulation results and uncover additional challenges. This real-world evaluation would bridge the gap between simulated environments and actual deployment conditions, ensuring the practical reliability and robustness of SLAM systems.

\section{Social Impact}\label{sec:social-impact}

The proposed pipeline for noisy data synthesis and the Noisy-Replica benchmark have the potential to advance the development of robust Simultaneous Localization and Mapping (SLAM) systems. By enabling the evaluation of SLAM models under diverse perturbations, this work can facilitate the creation of more resilient robotic systems capable of operating reliably in unstructured and challenging environments.
On the positive side, robust SLAM systems could enhance safety, efficiency, and effectiveness in various domains, such as autonomous navigation, exploration, and mapping in hazardous or inaccessible areas. These applications span industries like disaster response, construction, mining, and search and rescue operations, where reliable robotic operation is crucial.

However, it is essential to acknowledge potential negative implications. The synthesized perturbations and noisy environments, while designed to evaluate robustness, might not fully capture the complexities of real-world scenarios. Overreliance on these simulated environments could lead to overlooking unforeseen challenges, highlighting the importance of complementing simulations with real-world testing and validation.
Furthermore, while we have curated a diverse set of perturbations, inherent biases or blindspots in the perturbation taxonomy or synthesis process may exist. We emphasize the need for continuous refinement and expansion of the perturbation taxonomy to capture a broader range of real-world disturbances and mitigate potential biases.
Additionally, the potential for unethical misuse of robust SLAM systems should be considered. Manipulated or biased data could be used to construct environments that present a distorted view of reality, leading to erroneous decision-making or harmful consequences. To mitigate this risk, robust validation, fact-checking mechanisms, and adherence to ethical guidelines must be integrated into the development and deployment processes.

\clearpage

\section{Availability and Maintenance}
\label{sec:availablility-maintenance}

The code and datasets utilized in this study are publicly accessible. The project repository, \href{https://github.com/Xiaohao-Xu/SLAM-under-Perturbation}{SLAM-under-Perturbation}, contains the following resources:

\begin{itemize}
    \item \textbf{Robustness Benchmark}: The repository includes code for SLAM robustness evaluation under customized perturbations.
    \item \textbf{Baseline Models}: Detailed instructions are provided for running all baseline models, facilitating the reproduction of all results presented in this paper.
    \item \textbf{Experiment Reproduction}: Comprehensive guidelines for reproducing all experiments can be found in the \href{https://github.com/Xiaohao-Xu/SLAM-under-Perturbation/blob/main/benchmark/Instructions.md}{Instructions.md} file within the repository.
\end{itemize}

These resources ensure that researchers and practitioners can effectively utilize and extend the work presented in this study.

We encourage the community to propose more robust SLAM models to further advance the frontier of robust embodied agents, ensuring their safe and reliable deployment in real-world environments.

\section{License}\label{sec:license}
The benchmark and code are under the Apache License 2.0. Please refer to
\url{https://github.com/Xiaohao-Xu/SLAM-under-Perturbation/blob/main/LICENSE} for details.

\section{Public Resources Used}
\label{sec:public-assets}

We gratefully acknowledge the use of the following public resources in this work:

\begin{itemize}

\item Classification-Robustness\footnote{\url{https://github.com/hendrycks/robustness}} \dotfill Apache License 2.0
\item Replica\footnote{\url{https://github.com/facebookresearch/Replica-Dataset}.} \dotfill \href{https://github.com/facebookresearch/Replica-Dataset?tab=License-1-ov-file#readme}{Research-only License}

\item Nice-SLAM\footnote{\url{https://github.com/cvg/nice-slam}.} \dotfill Apache License 2.0

\item Co-SLAM\footnote{\url{https://github.com/HengyiWang/Co-SLAM}.} \dotfill Apache License 2.0

\item SplaTAM\footnote{\url{https://github.com/spla-tam/SplaTAM}.} \dotfill BSD 3-Clause "New" or "Revised" License

\item GO-SLAM\footnote{\url{https://github.com/youmi-zym/GO-SLAM}.} \dotfill Apache License 2.0

\item ORB-SLAM3\footnote{\url{https://github.com/UZ-SLAMLab/ORB_SLAM3}.} \dotfill GNU General Public License v3.0

\end{itemize}

\clearpage

\section{\textbf{Detailed Benchmarking Tables}}\label{sec:detailed-result-with-error-bar}
To mitigate potential randomness, for each perturbed setting, we conduct each experiment three times on eight 3D scenes (totaling 24 experiments per perturbed result) and report the averaged results. Specifically, for the RGB imaging perturbation, we present the averaged result across three severity levels, under both static and dynamic perturbation modes.
This approach reduces the impact of randomness while maintaining computational efficiency, striking a balance between mitigating randomness and ensuring feasibility.

To facilitate future quantitative comparison on our benchmark, we have provided additional detailed benchmarking tables of RGB-D SLAM methods in this section, which includes:
\begin{itemize}
    \item Neural SLAM methods under depth imaging perturbation (Table~\ref{tab:depth-perturb}).
        \item  Neural SLAM methods under the faster motion effect (Table~\ref{tab:faster-motion-perturbation}).
        \item Neural SLAM methods under motion deviations (Table~\ref{tab:trajectory-perturbation_sr_metric}).
        \item Neural SLAM methods under RGB-D de-synchronization (Table~\ref{tab:sensor-misalign-perturb}).
    \item Classical SLAM method ORB-SLAM3 under RGB imaging perturbation (Table~\ref{tab:image_perturb_ate_metric},Table~\ref{tab:image_perturb_rpe_metric}, Table~\ref{tab:image_perturb_sr_metric}).
    \item Classical SLAM method ORB-SLAM3 under depth imaging perturbation (Table~\ref{tab:depth-perturb_ate_metric}, Table~\ref{tab:depth-perturb_rpe_metric}, Table~\ref{tab:depth-perturb_sr_metric}).
        \item  Classical SLAM method ORB-SLAM3 under the faster motion effect (Table~\ref{tab:faster-motion-perturbation_ate_metric}, Table~\ref{tab:faster-motion-perturbation_rpe_metric}, Table~\ref{tab:faster-motion-perturbation_sr_metric}).
        \item Classical SLAM method ORB-SLAM3 under the motion deviations (Table~\ref{tab:motion-deviation-perturb_ate_metric}, Table~\ref{tab:motion-deviation-perturb_rpe_metric},
        Table~\ref{tab:motion-deviation-perturb_sr_metric} ).                
        \item Classical SLAM method ORB-SLAM3 under RGB-D de-synchronization (Table~\ref{tab:sensor-misalign-perturb_ate_metric}, Table~\ref{tab:sensor-misalign-perturb_rpe_metric}, Table~\ref{tab:sensor-misalign-perturb_sr_metric}).
\end{itemize}

Furthermore, we have noticed large performance deviations with each perturbed setting for the ORB-SLAM3 model. To reflect potential randomness, we have included the performance standard deviation of this model in our experiment results. We acknowledge these potential performance deviations, which indicate low model robustness – a phenomenon our work aims to highlight to encourage the SLAM community to develop more robust and stable SLAM systems.


\begin{table}[ht!]
\caption{Trajectory estimation error (ATE [m]) under {depth imaging perturbation} for RGB-D SLAM methods.}
\label{tab:depth-perturb}
\centering 
\resizebox{\textwidth}{!}{
\begin{tabular}{p{3.955cm}|p{1.795cm}<{\centering}|p{1.795cm}<{\centering}p{1.795cm}<{\centering}p{1.795cm}<{\centering}p{1.795cm}<{\centering}}
    \toprule \toprule   
    \multirow{2}{*}{\textbf{Method}} &    \multirow{2}{*}{\textbf{Clean}}
     &  {\textbf{Gaussian}}   & {\textbf{Edge}} & 
  {\textbf{Random}}  & {\textbf{Range}}      \\  
 & & \textbf{Noise} & \textbf{Erosion} & \textbf{Missing} & \textbf{Clipping}\\ \midrule
    iMAP (RGB-D)~\cite{imap} & $0.1209$ & $\usym{2715}$ & $0.0307$ & $0.1083$  & $0.2438$ 
    \\ 
    Nice-SLAM (RGB-D)~\cite{niceslam}& $0.0147$ & $\usym{2715}$ & $0.0149$ & $0.0154$  & $0.1183$ 
    \\
    CO-SLAM (RGB-D)~\cite{coslam}& $0.0090$ & $0.5794$ & $0.0096$ & $0.0094$  & $0.0122$ 
    \\
    GO-SLAM (RGB-D)~\cite{zhang2023goslam} & ${0.0046}$ & ${0.0378}$ & $0.0046$ & ${0.0046}$ & ${{0.0045}}$   
    \\
    SplaTAM-S (RGB-D)~\cite{keetha2023splatam}& ${0.0045}$ & $\usym{2715}$ & ${{0.0042}}$ & ${0.0042}$  & ${0.0048}$ 
    \\
    \bottomrule \bottomrule
    \multicolumn{6}{l}{{$\usym{2715}$ indicates  completely unacceptable, \textit{i.e.}, performance (ATE $\geq$ 1.0 [m])}} \\
    \end{tabular}
}
 
\end{table}

\begin{table}[t!]
\centering
\caption{Trajectory estimation error (ATE [m]) of monocular (\textbf{Top}) and RGB-D (\textbf{Bottom}) SLAM under faster motion.}
\label{tab:faster-motion-perturbation}
\resizebox{\textwidth}{!}
{
\begin{tabular}{p{3.955cm}|p{2.240cm}<{\centering}|p{2.240cm}<{\centering}p{2.240cm}<{\centering}p{2.240cm}<{\centering}}
\toprule \toprule
 {\textbf{Speed-up Ratio}}
 & {$\textbf{1}\times$} & \textbf{$\textbf{2}\times$} & \textbf{$\textbf{4}\times$} & \textbf{$\textbf{8}\times$}    \\ \midrule 
GO-SLAM (Mono)~\cite{zhang2023goslam} & ${0.0039}$ & ${0.0042}$ & ${0.0046}$ & ${0.0048}$ 
\\
\midrule 
iMAP (RGB-D)~\cite{imap}& $0.1209$ & $0.4675$ & $0.9445$ & ${1.0000}$ 
\\
Nice-SLAM (RGB-D)~\cite{niceslam} & $0.0147$ & $0.1702$ & ${1.0000}$ & ${1.0000}$  
\\
CO-SLAM (RGB-D)~\cite{coslam}& $0.0090$ & $0.1062$ & $0.9510$ & ${1.0000}$ 
\\
GO-SLAM (RGB-D)~\cite{zhang2023goslam} & ${0.0046}$ & ${{0.0046}}$ & ${{0.0046}}$ & ${0.0050}$  
\\
SplaTAM-S (RGB-D)~\cite{keetha2023splatam}& ${0.0045}$ & $0.0184$ & ${1.0000}$ & ${1.0000}$ 
\\
\bottomrule \bottomrule

\end{tabular}

}\end{table}

\begin{table}[ht]
\centering
\caption{Trajectory estimation error (ATE [m]) of Neural SLAM methods under motion deviations.}
\centering\setlength{\tabcolsep}{1.0mm}
\resizebox{\textwidth}{!}{
\begin{tabular}{l|c|ccc|cccc|cccc|cccc}
\toprule \toprule
\textbf{Rotate [$\deg$]} & \multirow{2}{*}{\textbf{Clean}}  & \multicolumn{3}{c|}{\textbf{\textit{0}}} & \multicolumn{4}{c|}{\textbf{\textit{1}}} & \multicolumn{4}{c|}{\textbf{\textit{3}}} & \multicolumn{4}{c}{\textbf{\textit{5}}}
\\ \cmidrule{1-1} \cmidrule{3-5} \cmidrule{6-9} \cmidrule{10-13} \cmidrule{14-17}
\textbf{Translate  [$m$]} &  & \textit{\textbf{0.0125}} &\textit{\textbf{0.025}} & \textbf{\textit{0.05}} & \textit{\textbf{0}} & \textit{\textbf{0.0125}} &\textit{\textbf{0.025}} & \textbf{\textit{0.05}} & \textit{\textbf{0}}  & \textit{\textbf{0.0125}} &\textit{\textbf{0.025}} & \textbf{\textit{0.05}} & \textit{\textbf{0}} & \textit{\textbf{0.0125}} &\textit{\textbf{0.025}} & \textbf{\textit{0.05}} 
\\ \midrule\midrule
GO-SLAM (Mono)~\cite{zhang2023goslam} & ${0.0039}$ & ${0.0084}$ & ${0.0077}$ & ${0.0091}$ & ${0.0083}$ & ${0.0079}$ & ${0.0082}$ & ${0.0094}$ & $\usym{2715}$ & $\usym{2715}$ & $\usym{2715}$ & $\usym{2715}$ & $\usym{2715}$ & $\usym{2715}$ & $\usym{2715}$ & $\usym{2715}$  
\\ \midrule
iMAP (RGB-D)~\cite{imap} & $0.1209$ & $0.0334$ & $0.1386$ & $0.0442$ & $0.2438$ & $0.2135$ & $0.3754$ & $0.2801$ & $\usym{2715}$ & $\usym{2715}$ & $\usym{2715}$ & $\usym{2715}$ & $\usym{2715}$ & $\usym{2715}$ & $\usym{2715}$ & $\usym{2715}$   \\
Nice-SLAM (RGB-D)~\cite{niceslam}& $0.0147$ & $0.5812$ & $\usym{2715}$ & $\usym{2715}$ & $\usym{2715}$ & $\usym{2715}$ & $\usym{2715}$ & $\usym{2715}$ & $\usym{2715}$ & $\usym{2715}$ & $\usym{2715}$ & $\usym{2715}$ & $\usym{2715}$ & $\usym{2715}$ & $\usym{2715}$ & $\usym{2715}$ 
\\
CO-SLAM (RGB-D)~\cite{coslam}& $0.0090$ & $0.0420$ & $0.0848$ & $0.3087$ & $0.4579$ & $0.5069$ & $0.2998$ & $0.5040$ & $0.6443$ & $0.6630$ & $0.7532$ & $0.5772$ & $0.8457$ & $0.7966$ & $0.8277$ & $\usym{2715}$ 
\\
GO-SLAM (RGB-D)~\cite{zhang2023goslam} & ${0.0046}$ & ${0.0082}$ & ${0.0082}$ & ${0.0081}$ & ${0.0080}$ & ${0.0080}$ & ${0.0078}$ & ${0.0077}$ & $\usym{2715}$ & $\usym{2715}$ & $\usym{2715}$ & $\usym{2715}$ & $\usym{2715}$ & $\usym{2715}$ & $\usym{2715}$ & $\usym{2715}$  
\\
SplaTAM-S (RGB-D)~\cite{keetha2023splatam}& ${0.0045}$ & $0.0545$ & $0.0980$ & $0.2964$ & $0.297F$ & $0.2272$ & $0.2313$ & $\usym{2715}$ & $\usym{2715}$ & $\usym{2715}$ & $\usym{2715}$ & $\usym{2715}$ & $\usym{2715}$ & $\usym{2715}$ & $\usym{2715}$ & $\usym{2715}$ 
\\ 
\bottomrule \bottomrule
\multicolumn{17}{l}{{Notation \textit{\textbf{F}} represents settings that include failure sequences where no final trajectory is generated due to tracking loss. The number in front of \textit{\textbf{F}} represents the}}\\
\multicolumn{17}{l}{{average ATE as failure sequences are set as a value of 1.0. Notation $\usym{2715}$ indicates completely unacceptable trajectory estimation performance, \textit{i.e.}, ATE $\geq$ 1.0 [m].}}\\ 
\end{tabular}
}
\label{tab:trajectory-perturbation_sr_metric}
\end{table}

\begin{table}[ht]
\caption{Trajectory estimation error (ATE [m]) under sensor de-synchronization  for RGB-D SLAM methods.}
\label{tab:sensor-misalign-perturb}
\centering 
\resizebox{\textwidth}{!}{
\begin{tabular}{l|c|ccc|ccc}
    \toprule \toprule
     \multirow{2}{*}{\textbf{Method}} & \textbf{Clean} &  \multicolumn{3}{c|}{\textbf{Static Mode}}& \multicolumn{3}{c}{\textbf{Dynamic Mode}} \\ \cmidrule{2-5} \cmidrule{6-8}
     & \textbf{$\Delta=0$} & \textbf{$\Delta=5$} & \textbf{$\Delta=10$} & \textbf{$\Delta=20$} & \textbf{$\Delta=5$} & \textbf{$\Delta=10$} & \textbf{$\Delta=20$}      \\ \midrule
    iMAP (RGB-D)~\cite{imap} & $0.1209$ & $0.4672$ & $0.5344$ & $0.6345$  & $0.5104$ & $0.6803$ & $0.6745$ 
    \\
    Nice-SLAM (RGB-D)~\cite{niceslam}& $0.0147$ & $0.3820$ & $0.4062$ & $0.5216$  & $0.5433$ & $0.5548$ & $0.7020$
    \\
     CO-SLAM (RGB-D)~\cite{coslam}& $0.0090$ & $0.0520$ & $0.1005$ & $0.1939$ & $0.0740$ & $0.1164$ & $0.2108$ \\

    GO-SLAM (RGB-D)~\cite{zhang2023goslam} & $0.0046$ & ${0.0148}$ & ${0.0292}$ & ${0.0646}$  & {${0.0151}$} & {${0.0297}$} & {${0.0650}$} 
    \\
    SplaTAM-S (RGB-D)~\cite{keetha2023splatam}& ${0.0045}$ & $0.0554$ & $0.0629$ & $0.0880$  & $0.0402$ & $0.0645$ & $0.0850$ 
    \\
    \bottomrule  \bottomrule
      \multicolumn{8}{l}{{  $\Delta$ denotes the misaligned frame interval between RGB and depth streams. }}\\
    \end{tabular}
}
\end{table}

\clearpage

\begin{table*}[t]
\caption{ATE metric of trajectory estimation under RGB imaging perturbations for ORB-SLAM3~\cite{orbslam3}. }
\label{tab:image_perturb_ate_metric}
\centering\setlength{\tabcolsep}{0.6mm}
\resizebox{\textwidth}{!}{
\begin{tabular}{cc|c|cccc|cccc|cccc|cccc}
    \toprule \toprule
  \textbf{Perturb.} &   \textbf{Input} &  \multirow{2}{*}{\textbf{Clean}}  & \multicolumn{4}{|c|}{\textbf{Blur Effect}} & \multicolumn{4}{c|}{\textbf{Noise Effect}} & \multicolumn{4}{c|}{\textbf{Environmental Interference}}  & \multicolumn{4}{c}{\textbf{Post-processing Effect}} 
    \\ \cmidrule{4-7} \cmidrule{8-11} \cmidrule{12-15} \cmidrule{16-19}
 \textbf{Mode}  & \textbf{Modality}   &   & \textbf{Motion} & \textbf{Defocus} & \textbf{Gaussian} & \textbf{Glass} & \textbf{Gaussian} & \textbf{Shot} & \textbf{Impulse} & \textbf{Speckle} & \textbf{Fog} & \textbf{Frost} & \textbf{Snow} & \textbf{Spatter} & \textbf{Bright} & \textbf{Contra.} & \textbf{JPEG}  & \textbf{Pixelate}
    \\\midrule\midrule
    \multirow{4}{*}{\textbf{Static}} &\multirow{2}{*}{Mono} & ${0.014}$  & ${0.891}$ & ${0.535}$ & ${0.505}$ & ${0.797}$ & ${0.917}$ & ${0.969}$ & ${1.000}$ & ${0.923}$ & $0.629$ & ${0.917}$ & ${1.000}$  & ${0.719}$ & ${0.027}$ & ${0.612}$ & $0.173$ & ${0.768}$ 
\\  
   &   & $\pm{0.028}$& $\pm0.285$ & $\pm0.487$ & $\pm0.506$ & $\pm0.404$ & $\pm0.281$ & $\pm0.152$ & $\pm0.000$ & $\pm0.261$ & $\pm0.490$ & $\pm0.280$ & $\pm0.000$  & $\pm0.449$ & $\pm0.059$ & $\pm0.478$ & $\pm0.325$ & $\pm0.410$ 
    \\ \cmidrule{2-19}
    &\multirow{2}{*}{RGB-D} & ${0.082}$  & ${0.300}$ & ${0.340}$ & ${0.292}$ & ${0.211}$ & ${0.470}$ & ${0.433}$ & ${0.591}$ & ${0.351}$ & ${0.397}$ & ${0.640}$ & ${0.795}$ & ${0.508}$ & ${0.065}$ & ${0.527}$ & ${0.132}$ & ${0.703}$
    \\
        &  & ${\pm0.179}$ & $\pm0.365$ & $\pm0.408$ & $\pm0.425$ & $\pm0.335$ & $\pm0.498$ & $\pm0.490$ & $\pm0.494$ & $\pm0.469$ & $\pm0.480$ & $\pm0.474$ & $\pm0.409$ & $\pm0.503$ & $\pm0.142$ & $\pm0.492$ & $\pm0.191$ & $\pm0.354$
    \\
  \midrule\midrule
   \multirow{4}{*}{\textbf{Dynamic}} &\multirow{2}{*}{Mono}  & ${0.014}$  & $0.876$ & $0.506$ & $0.760$ & $1.000$ & $1.000$ & $1.000$ & $1.000$ & $0.885$ & ${0.751}$ & $1.000$ & $1.000$ & $1.000$ & $0.052$ & $1.000$ & ${0.166}$ & $0.718$
    \\
     &   & ${\pm0.028}$ & $\pm0.351$ & $\pm0.528$ & $\pm0.445$ & $\pm0.000$ & $\pm0.000$ & $\pm0.000$ & $\pm0.000$ & $\pm0.325$ & $\pm0.461$ & $\pm0.000$ & $\pm0.000$ & $\pm0.000$ & $\pm0.092$ & $\pm0.000$ & $\pm0.341$ & $\pm0.412$
    \\ \cmidrule{2-19}
   &\multirow{2}{*}{RGB-D} & ${0.082}$  & $0.279$ & $0.303$ & $0.052$ & $0.168$ & $0.508$ & $0.513$ & $0.755$ & $0.262$ & $0.659$ & $0.256$ & $0.876$ & $0.753$ & $0.066$ & $0.751$ & $0.645$ & $0.756$
    \\ 
     &   & ${\pm0.179}$   & $\pm0.330$ & $\pm0.435$ & $\pm0.080$ & $\pm0.245$ & $\pm0.526$ & $\pm0.520$ & $\pm0.453$ & $\pm0.455$ & $\pm0.476$ & $\pm0.459$ & $\pm0.352$ & $\pm0.458$ & $\pm0.145$ & $\pm0.462$ & $\pm0.491$ & $\pm0.294$ 
    \\
    \bottomrule   \bottomrule  
    \end{tabular}
}
\end{table*}
\begin{table*}[t]
\caption{RPE metric of trajectory estimation under RGB imaging perturbations for ORB-SLAM3~\cite{orbslam3}. }
\label{tab:image_perturb_rpe_metric}
\centering\setlength{\tabcolsep}{0.6mm}
\resizebox{\textwidth}{!}{
\begin{tabular}{cc|c|cccc|cccc|cccc|cccc}
    \toprule \toprule
  \textbf{Perturb.} &   \textbf{Input} &  \multirow{2}{*}{\textbf{Clean}} & \multicolumn{4}{|c|}{\textbf{Blur Effect}} & \multicolumn{4}{c|}{\textbf{Noise Effect}} & \multicolumn{4}{c|}{\textbf{Environmental Interference}}  & \multicolumn{4}{c}{\textbf{Post-processing Effect}} 
    \\ \cmidrule{4-7} \cmidrule{8-11} \cmidrule{12-15} \cmidrule{16-19}
 \textbf{Mode}  & \textbf{Modality}   &    & \textbf{Motion} & \textbf{Defocus} & \textbf{Gaussian} & \textbf{Glass} & \textbf{Gaussian} & \textbf{Shot} & \textbf{Impulse} & \textbf{Speckle} & \textbf{Fog} & \textbf{Frost} & \textbf{Snow} & \textbf{Spatter} & \textbf{Bright} & \textbf{Contra.} & \textbf{JPEG}  & \textbf{Pixelate}
    \\\midrule\midrule
    \multirow{4}{*}{\textbf{Static}} &\multirow{2}{*}{Mono} & ${0.197}$& $0.853$ & $0.605$ & $0.601$ & $0.811$ & $0.927$ & $0.960$ & $0.100$ & $0.932$ & $0.680$ & $0.927$ & $1.000$ & $0.760$ & $0.162$ & $0.699$ & $0.247$ & $0.525$
    \\ 
     &   & ${\pm0.030}$   & $\pm0.335$ & $\pm0.408$ & $\pm0.410$ & $\pm0.376$ & $\pm0.247$ & $\pm0.194$ & $\pm0.000$ & $\pm0.230$ & $\pm0.423$ & $\pm0.246$ & $\pm0.000$ & $\pm0.383$ & $\pm0.028$ & $\pm0.373$ & $\pm0.294$ & $\pm0.485$
    \\ \cmidrule{2-19}
    &\multirow{2}{*}{RGB-D} & ${0.114}$ & ${0.238}$ & ${0.323}$ & ${0.329}$ & ${0.084}$ & ${0.493}$ & ${0.445}$ & ${0.601}$ & ${0.377}$ & ${0.421}$ & ${0.617}$ & ${0.798}$ & ${0.529}$ & ${0.078}$ & ${0.570}$ & ${0.080}$ & ${0.032}$
    \\
        &  & ${\pm0.023}$ & $\pm0.350$ & $\pm0.400$ & $\pm0.397$ & $\pm0.036$ & $\pm0.477$ & $\pm0.479$ & $\pm0.482$ & $\pm0.450$ & $\pm0.459$ & $\pm0.470$ & $\pm0.403$ & $\pm0.481$ & $\pm0.022$ & $\pm0.442$ & $\pm0.028$ & $\pm0.026$
    \\
  \midrule\midrule
   \multirow{4}{*}{\textbf{Dynamic}} &\multirow{2}{*}{Mono}  & ${0.197}$  & $0.879$ & $0.572$ & $0.782$ & $1.000$ & $1.000$ & $1.000$ & $1.000$ & $0.899$ & ${0.785}$ & $1.000$ & $1.000$ & $1.000$ & $0.144$ & $1.000$ & ${0.242}$ & $0.287$
    \\
     &   & ${\pm0.030}$   & $\pm0.342$ & $\pm0.459$ & $\pm0.403$ & $\pm0.000$ & $\pm0.000$ & $\pm0.000$ & $\pm0.000$ & $\pm0.284$ & $\pm0.398$ & $\pm0.000$ & $\pm0.000$ & $\pm0.000$ & $\pm0.034$ & $\pm0.000$ & $\pm0.307$ & $\pm0.440$
    \\ \cmidrule{2-19}
   &\multirow{2}{*}{RGB-D}  & ${0.114}$   & ${0.043}$ & ${0.293}$ & ${0.033}$ & ${0.046}$ & ${0.517}$ & ${0.523}$ & ${0.755}$ & ${0.284}$ & $0.642$ & ${0.272}$ & ${0.876}$  & ${0.758}$ & ${0.074}$ & ${0.792}$ & $0.656$ & ${0.031}$ 
\\  
   &   & $\pm{0.023}$  & $\pm0.024$ & $\pm0.437$ & $\pm0.027$ & $\pm0.021$ & $\pm0.517$ & $\pm0.511$ & $\pm0.453$ & $\pm0.443$ & $\pm0.494$ & $\pm0.450$ & $\pm0.350$  & $\pm0.448$ & $\pm0.016$ & $\pm0.393$ & $\pm0.474$ & $\pm0.011$ 
\\  
    \bottomrule   \bottomrule  

    \end{tabular}
}
\end{table*}

\begin{table*}[t]
\caption{Success rate (SR) of pose tracking under RGB imaging perturbations for ORB-SLAM3~\cite{orbslam3}. }
\label{tab:image_perturb_sr_metric}
\centering\setlength{\tabcolsep}{0.6mm}
\resizebox{\textwidth}{!}{
\begin{tabular}{cc|c|cccc|cccc|cccc|cccc}
    \toprule \toprule
  \textbf{Perturb.} &   \textbf{Input} &  \multirow{2}{*}{\textbf{Clean}} & \multicolumn{4}{|c|}{\textbf{Blur Effect}} & \multicolumn{4}{c|}{\textbf{Noise Effect}} & \multicolumn{4}{c|}{\textbf{Environmental Interference}}  & \multicolumn{4}{c}{\textbf{Post-processing Effect}} 
    \\ \cmidrule{4-7} \cmidrule{8-11} \cmidrule{12-15} \cmidrule{16-19}
 \textbf{Mode}  & \textbf{Modality}   &   & \textbf{Motion} & \textbf{Defocus} & \textbf{Gaussian} & \textbf{Glass} & \textbf{Gaussian} & \textbf{Shot} & \textbf{Impulse} & \textbf{Speckle} & \textbf{Fog} & \textbf{Frost} & \textbf{Snow} & \textbf{Spatter} & \textbf{Bright} & \textbf{Contra.} & \textbf{JPEG}  & \textbf{Pixelate}
    \\\midrule\midrule
    \multirow{4}{*}{\textbf{Static}} &\multirow{2}{*}{Mono} & ${0.854}$   & $0.059$ & $0.331$ & $0.382$ & $0.122$ & $0.055$ & $0.016$ & $0.000$ & $0.052$ & $0.229$ & $0.057$ & $0.000$ & $0.211$ & $0.915$ & $0.320$ & $0.712$ & $0.064$
    \\ 
     &   & ${\pm0.149}$   & $\pm0.152$ & $\pm0.385$ & $\pm0.415$ & $\pm0.307$ & $\pm0.186$ & $\pm0.076$ & $\pm0.000$ & $\pm0.183$ & $\pm0.362$ & $\pm0.197$ & $\pm0.000$ & $\pm0.352$ & $\pm0.142$ & $\pm0.405$ & $\pm0.349$ & $\pm0.089$
    \\ \cmidrule{2-19}
    &\multirow{2}{*}{RGB-D} & ${0.960}$ & ${0.621}$ & ${0.606}$ & ${0.617}$ & ${0.778}$ & ${0.388}$ & ${0.391}$ & ${0.219}$ & ${0.443}$ & ${0.276}$ & ${0.111}$ & ${0.081}$ & ${0.259}$ & ${0.971}$ & ${0.412}$ & ${0.818}$ & ${0.361}$
    \\
        &  & ${\pm0.046}$  & $\pm0.423$ & $\pm0.443$ & $\pm0.430$ & $\pm0.319$ & $\pm0.468$ & $\pm0.475$ & $\pm0.409$ & $\pm0.457$ & $\pm0.421$ & $\pm0.282$ & $\pm0.236$ & $\pm0.405$ & $\pm0.030$ & $\pm0.461$ & $\pm0.284$ & $\pm0.232$
    \\
  \midrule\midrule
   \multirow{4}{*}{\textbf{Dynamic}} &\multirow{2}{*}{Mono}  & ${0.854}$  & $0.008$ & $0.267$ & $0.158$ & $0.000$ & $0.000$ & $0.000$ & $0.000$ & $0.096$ & ${0.169}$ & $0.000$ & $0.000$ & $0.000$ & $0.904$ & $0.000$ & ${0.703}$ & $0.142$
    \\
     &   & ${\pm0.149}$ &${\pm0.270}$ & $\pm0.000$  & $\pm0.022$ & $\pm0.000$ & $\pm0.000$ & $\pm0.000$ & $\pm0.000$ & $\pm0.273$ & $\pm0.315$ & $\pm0.000$ & $\pm0.000$ & $\pm0.000$ & $\pm0.046$ & $\pm0.000$ & $\pm0.328$ & $\pm0.168$
    \\ \cmidrule{2-19}
   &\multirow{2}{*}{RGB-D}  & ${0.960}$  & ${0.871}$ & ${0.354}$ & ${0.301}$ & ${0.991}$ & ${0.207}$ & ${0.277}$ & ${0.040}$ & ${0.303}$ & $0.027$ & ${0.005}$ & ${0.000}$  & ${0.009}$ & ${0.975}$ & ${0.003}$ & $0.376$ & ${0.432}$ 
\\  
   &   & $\pm{0.046}$   & $\pm0.364$ & $\pm0.429$ & $\pm0.437$ & $\pm0.081$ & $\pm0.378$ & $\pm0.473$ & $\pm0.084$ & $\pm0.441$ & $\pm0.045$ & $\pm0.009$ & $\pm0.000$  & $\pm0.026$ & $\pm0.027$ & $\pm0.007$ & $\pm0.519$ & $\pm0.218$ 
\\  
    \bottomrule   \bottomrule  
    \end{tabular}
}
\end{table*}

\begin{table}[t]
\caption{ATE metric of trajectory estimation under depth perturbations for ORB-SLAM3~\cite{orbslam3}  with RGB-D input. }
\label{tab:depth-perturb_ate_metric}
\centering 
\resizebox{\textwidth}{!}{
\begin{tabular}{p{2.495cm}<{\centering}|p{2.495cm}<{\centering}p{2.495cm}<{\centering}p{2.495cm}<{\centering}p{2.495cm}<{\centering}}
    \toprule \toprule   
       \multirow{2}{*}{\textbf{Clean}}
     &  {\textbf{Gaussian}}   & {\textbf{Edge}} & 
  {\textbf{Random}}  & {\textbf{Range}}      \\  
  & \textbf{Noise} & \textbf{Erosion} & \textbf{Missing} & \textbf{Clipping}\\ \midrule
    ${0.082}\pm 0.179$ & $0.803\pm 0.301$ & $0.807\pm 0.365$ & $0.756\pm 0.397$ & $0.994\pm 0.283$
    \\
    \bottomrule  \bottomrule 
    \end{tabular}
}
\end{table}
\begin{table}[t]
\caption{RPE metric of trajectory estimation under depth perturbations for ORB-SLAM3~\cite{orbslam3}  with RGB-D input. }
\label{tab:depth-perturb_rpe_metric}
\centering 
\resizebox{\textwidth}{!}{
\begin{tabular}{p{2.495cm}<{\centering}|p{2.495cm}<{\centering}p{2.495cm}<{\centering}p{2.495cm}<{\centering}p{2.495cm}<{\centering}}
    \toprule \toprule   
       \multirow{2}{*}{\textbf{Clean}}
     &  {\textbf{Gaussian}}   & {\textbf{Edge}} & 
  {\textbf{Random}}  & {\textbf{Range}}      \\  
  & \textbf{Noise} & \textbf{Erosion} & \textbf{Missing} & \textbf{Clipping}\\ \midrule
    ${0.114}\pm 0.023$ & $0.064\pm 0.018$ & $0.154\pm 0.342$ & $0.054\pm 0.022$ & $0.047\pm 0.017$
    \\
    \bottomrule  \bottomrule 
    \end{tabular}
}
\end{table}

\begin{table}
\vspace{2mm}
\caption{Success rate (SR) of pose tracking under depth perturbations for ORB-SLAM3~\cite{orbslam3}  with RGB-D input. }
\label{tab:depth-perturb_sr_metric}
\centering 
\resizebox{\textwidth}{!}{
\begin{tabular}{p{2.495cm}<{\centering}|p{2.495cm}<{\centering}p{2.495cm}<{\centering}p{2.495cm}<{\centering}p{2.495cm}<{\centering}}
    \toprule \toprule   
       \multirow{2}{*}{\textbf{Clean}}
     &  {\textbf{Gaussian}}   & {\textbf{Edge}} & 
  {\textbf{Random}}  & {\textbf{Range}}      \\  
  & \textbf{Noise} & \textbf{Erosion} & \textbf{Missing} & \textbf{Clipping}\\ \midrule
    ${0.960}\pm 0.046$ & $0.421\pm 0.331$ & $0.379\pm 0.281$ & $0.322\pm 0.354$ & $0.286\pm 0.181$
    \\
    \bottomrule  \bottomrule 
    \end{tabular}
}
\end{table}

\begin{table}[t]
 \caption{ATE metric of trajectory estimation under the faster motion perturbation for ORB-SLAM3~\cite{orbslam3}.}
\centering 
\label{tab:faster-motion-perturbation_ate_metric}
\resizebox{\textwidth}{!}{
\begin{tabular}{p{2.495cm}|p{2.495cm}<{\centering}|p{2.495cm}<{\centering}p{2.495cm}<{\centering}p{2.495cm}<{\centering}}
    \toprule \toprule
     {\textbf{Speed-up Ratio}}
     & {$\textbf{1}\times$} & \textbf{$\textbf{2}\times$} & \textbf{$\textbf{4}\times$} & \textbf{$\textbf{8}\times$}    \\ \midrule
Monocular & ${0.014}\pm 0.028$ & ${{0.009}}\pm 0.009$ & ${0.023}\pm 0.051$ & ${{0.023}}\pm 0.053$  \\ 
    RGB-D &  ${0.082}\pm 0.179$  & ${0.019}\pm 0.023$ & ${{0.064}}\pm 0.156$ & ${{0.077}}\pm 0.125$ \\
    \bottomrule \bottomrule
    \end{tabular}
}
\end{table}

\begin{table}[t]
 \caption{RPE metric of trajectory estimation under the faster motion perturbation for ORB-SLAM3~\cite{orbslam3}.}
\centering 
\label{tab:faster-motion-perturbation_rpe_metric}
\resizebox{\textwidth}{!}{
\begin{tabular}{p{2.495cm}|p{2.495cm}<{\centering}|p{2.495cm}<{\centering}p{2.495cm}<{\centering}p{2.495cm}<{\centering}}
    \toprule \toprule
     {\textbf{Speed-up Ratio}}
     & {$\textbf{1}\times$} & \textbf{$\textbf{2}\times$} & \textbf{$\textbf{4}\times$} & \textbf{$\textbf{8}\times$}    \\ \midrule
Monocular & ${0.197}\pm 0.030$ & ${{0.194}}\pm 0.035$ & ${0.227}\pm 0.035$ & ${{0.289}}\pm 0.064$  \\ 

    RGB-D &  ${0.114}\pm 0.023$  & ${0.152}\pm 0.032$ & ${{0.190}}\pm 0.030$ & ${{0.268}}\pm 0.059$ \\
    \bottomrule \bottomrule
    \end{tabular}
}
\end{table}

\begin{table}
 
 \caption{Success rate (SR) of pose tracking under the faster motion perturbation for ORB-SLAM3~\cite{orbslam3}.}
\centering 
\label{tab:faster-motion-perturbation_sr_metric}
\resizebox{\textwidth}{!}{
\begin{tabular}{p{2.495cm}|p{2.495cm}<{\centering}|p{2.495cm}<{\centering}p{2.495cm}<{\centering}p{2.495cm}<{\centering}}
    \toprule \toprule
     {\textbf{Speed-up Ratio}}
     & {$\textbf{1}\times$} & \textbf{$\textbf{2}\times$} & \textbf{$\textbf{4}\times$} & \textbf{$\textbf{8}\times$}    \\ \midrule
Monocular & ${0.854}\pm 0.149$ & ${{0.893}}\pm 0.081$ & ${0.909}\pm 0.049$ & ${{0.837}}\pm 0.115$  \\ 
    RGB-D &  ${0.960}\pm 0.046$  & ${0.964}\pm 0.012$ & ${{0.938}}\pm 0.029$ & ${{0.866}}\pm 0.129$ \\
    \bottomrule \bottomrule
    \end{tabular}
}
\end{table}

\begin{table*}[th]
    \centering
    \vspace{-2mm}
       \caption{ATE metric of trajectory estimation under motion deviations for {ORB-SLAM3}~\cite{orbslam3}.}
             \label{tab:motion-deviation-perturb_ate_metric}
\centering\setlength{\tabcolsep}{1.0mm}
\resizebox{\textwidth}{!}{
\begin{tabular}{l|c|ccc|cccc|cccc|cccc}
    \toprule \toprule
     \textbf{Rotate [$\deg$]} & \multirow{2}{*}{\textbf{Clean}}  & \multicolumn{3}{c|}{\textbf{\textit{0}}} & \multicolumn{4}{c|}{\textbf{\textit{1}}} & \multicolumn{4}{c|}{\textbf{\textit{3}}} & \multicolumn{4}{c}{\textbf{\textit{5}}}
    \\ \cmidrule{1-1} \cmidrule{3-5} \cmidrule{6-9} \cmidrule{10-13} \cmidrule{14-17}
     \textbf{Translate  [$m$]} &  & \textit{\textbf{0.0125}} &\textit{\textbf{0.025}} & \textbf{\textit{0.05}} & \textit{\textbf{0}} & \textit{\textbf{0.0125}} &\textit{\textbf{0.025}} & \textbf{\textit{0.05}} & \textit{\textbf{0}}  & \textit{\textbf{0.0125}} &\textit{\textbf{0.025}} & \textbf{\textit{0.05}} & \textit{\textbf{0}} & \textit{\textbf{0.0125}} &\textit{\textbf{0.025}} & \textbf{\textit{0.05}} 
    \\\midrule\midrule
      \multirow{2}{*}{Monocular} & ${0.014}$ & ${0.017}$ & $0.136$ & $0.190$ & ${0.063}$ & $0.146$ & $0.348$ & ${0.200}$ & ${0.170}$ & $0.055$ & $0.053$ & $0.061$ & $0.073$ & $0.023$ & ${0.041}$ & ${0.080}$ 
     \\ 
        & $\pm0.028$ & $\pm0.027$ & $\pm0.349$ & $\pm0.339$ & $\pm0.047$ & $\pm0.196$ & $\pm0.378$ & $\pm0.355$ & $\pm0.338$ & $\pm0.057$ & $\pm0.061$ & $\pm0.061$ & $\pm0.135$ & $\pm0.039$ & $\pm0.042$ & $\pm0.108$ 
     \\ \midrule
    \multirow{2}{*}{RGB-D} & ${0.082}$ & $0.191$ & ${0.085}$ & ${0.171}$ & $0.059$ & ${0.163}$ & ${0.084}$ & $0.228$ & $0.148$ & ${0.090}$ & ${0.158}$ & ${0.085}$ & ${0.164}$ & ${0.072}$ & $0.050$ & $0.062$ 
    \\
        & $\pm0.179$ & $\pm0.369$ & $\pm0.174$ & $\pm0.337$ & $\pm0.067$ & $\pm0.280$ & $\pm0.052$ & $\pm0.361$ & $\pm0.210$ & $\pm0.072$ & $\pm0.106$ & $\pm0.037$ & $\pm0.340$ & $\pm0.100$ & $\pm0.054$ & $\pm0.076$ 
    \\
    \bottomrule \bottomrule
    \end{tabular}
}

\end{table*}%
\begin{table*}[th]
    \centering
    \vspace{-2mm}
       \caption{RPE metric of trajectory estimation under motion deviations for {ORB-SLAM3}~\cite{orbslam3}.}
             \label{tab:motion-deviation-perturb_rpe_metric}
\centering\setlength{\tabcolsep}{1.0mm}
\resizebox{\textwidth}{!}{
\begin{tabular}{l|c|ccc|cccc|cccc|cccc}
    \toprule \toprule
     \textbf{Rotate [$\deg$]} & \multirow{2}{*}{\textbf{Clean}}  & \multicolumn{3}{c|}{\textbf{\textit{0}}} & \multicolumn{4}{c|}{\textbf{\textit{1}}} & \multicolumn{4}{c|}{\textbf{\textit{3}}} & \multicolumn{4}{c}{\textbf{\textit{5}}}
    \\ \cmidrule{1-1} \cmidrule{3-5} \cmidrule{6-9} \cmidrule{10-13} \cmidrule{14-17}
     \textbf{Translate  [$m$]} &  & \textit{\textbf{0.0125}} &\textit{\textbf{0.025}} & \textbf{\textit{0.05}} & \textit{\textbf{0}} & \textit{\textbf{0.0125}} &\textit{\textbf{0.025}} & \textbf{\textit{0.05}} & \textit{\textbf{0}}  & \textit{\textbf{0.0125}} &\textit{\textbf{0.025}} & \textbf{\textit{0.05}} & \textit{\textbf{0}} & \textit{\textbf{0.0125}} &\textit{\textbf{0.025}} & \textbf{\textit{0.05}} 
    \\\midrule\midrule
      \multirow{2}{*}{Monocular} & ${0.197}$ & ${0.191}$ & $0.291$ & $0.297$ & ${0.182}$ & $0.194$ & $0.291$ & ${0.197}$ & ${0.349}$ & $0.231$ & $0.265$ & $0.235$ & $0.355$ & $0.324$ & ${0.368}$ & ${0.371}$ 
     \\ 
        & $\pm0.030$ & $\pm0.037$ & $\pm0.289$ & $\pm0.297$ & $\pm0.071$ & $\pm0.036$ & $\pm0.294$ & $\pm0.075$ & $\pm0.270$ & $\pm0.086$ & $\pm0.105$ & $\pm0.087$ & $\pm0.052$ & $\pm0.083$ & $\pm0.067$ & $\pm0.040$ 
     \\ \midrule
    \multirow{2}{*}{RGB-D} & ${0.114}$ & $0.229$ & ${0.119}$ & ${0.208}$ & $0.114$ & ${0.116}$ & ${0.153}$ & $0.231$ & $0.197$ & ${0.220}$ & ${0.203}$ & ${0.220}$ & ${0.417}$ & ${0.313}$ & $0.298$ & $0.334$ 
    \\
        & $\pm0.023$ & $\pm0.313$ & $\pm0.028$ & $\pm0.322$ & $\pm0.039$ & $\pm0.038$ & $\pm0.046$ & $\pm0.313$ & $\pm0.046$ & $\pm0.027$ & $\pm0.035$ & $\pm0.026$ & $\pm0.239$ & $\pm0.047$ & $\pm0.052$ & $\pm0.042$ 
    \\
    \bottomrule \bottomrule
    \end{tabular}
}

  \vspace{-2mm}
\end{table*}%

\begin{table*}[th]
    \centering
       \caption{Success rate (SR) of pose tracking under motion deviations for {ORB-SLAM3}~\cite{orbslam3}.}
                  \label{tab:motion-deviation-perturb_sr_metric}
\centering\setlength{\tabcolsep}{1.0mm}
\resizebox{\textwidth}{!}{
\begin{tabular}{l|c|ccc|cccc|cccc|cccc}
    \toprule \toprule
     \textbf{Rotate [$\deg$]} & \multirow{2}{*}{\textbf{Clean}}  & \multicolumn{3}{c|}{\textbf{\textit{0}}} & \multicolumn{4}{c|}{\textbf{\textit{1}}} & \multicolumn{4}{c|}{\textbf{\textit{3}}} & \multicolumn{4}{c}{\textbf{\textit{5}}}
    \\ \cmidrule{1-1} \cmidrule{3-5} \cmidrule{6-9} \cmidrule{10-13} \cmidrule{14-17}
     \textbf{Translate  [$m$]} &  & \textit{\textbf{0.0125}} &\textit{\textbf{0.025}} & \textbf{\textit{0.05}} & \textit{\textbf{0}} & \textit{\textbf{0.0125}} &\textit{\textbf{0.025}} & \textbf{\textit{0.05}} & \textit{\textbf{0}}  & \textit{\textbf{0.0125}} &\textit{\textbf{0.025}} & \textbf{\textit{0.05}} & \textit{\textbf{0}} & \textit{\textbf{0.0125}} &\textit{\textbf{0.025}} & \textbf{\textit{0.05}} 
    \\\midrule\midrule
      \multirow{2}{*}{Monocular} & ${0.854}$ & ${0.489}$ & $0.247$ & $0.128$ & ${0.662}$ & $0.340$ & $0.221$ & ${0.120}$ & ${0.373}$ & $0.096$ & $0.087$ & $0.059$ & $0.413$ & $0.155$ & ${0.214}$ & ${0.144}$ 
     \\ 
        & $\pm0.149$ & $\pm0.107$ & $\pm0.129$ & $\pm0.091$ & $\pm0.329$ & $\pm0.189$ & $\pm0.156$ & $\pm0.069$ & $\pm0.329$ & $\pm0.154$ & $\pm0.081$ & $\pm0.056$ & $\pm0.418$ & $\pm0.160$ & $\pm0.112$ & $\pm0.039$ 
     \\ \midrule
    \multirow{2}{*}{RGB-D} & ${0.960}$ & $0.462$ & ${0.321}$ & ${0.152}$ & $0.596$ & ${0.345}$ & ${0.228}$ & $0.114$ & $0.325$ & ${0.259}$ & ${0.146}$ & ${0.118}$ & ${0.422}$ & ${0.158}$ & $0.180$ & $0.129$ 
    \\
        & $\pm0.046$ & $\pm0.213$ & $\pm0.085$ & $\pm0.100$ & $\pm0.491$ & $\pm0.218$ & $\pm0.151$ & $\pm0.115$ & $\pm0.270$ & $\pm0.226$ & $\pm0.121$ & $\pm0.087$ & $\pm0.428$ & $\pm0.156$ & $\pm0.148$ & $\pm0.092$ 
    \\
    \bottomrule \bottomrule
    \end{tabular}
}
         \label{tab:trajectory-perturbation}
  \vspace{-2mm}
\end{table*}%

\clearpage

\begin{table}[t]
\vspace{3mm}
\caption{ATE Metric under sensor de-synchronization for  RGBD-based {ORB-SLAM3}~\cite{orbslam3}. }
\centering 
\label{tab:sensor-misalign-perturb_ate_metric}
\resizebox{\textwidth}{!}{
\begin{tabular}{p{1.495cm}|p{2.495cm}<{\centering}|p{2.495cm}<{\centering}p{2.495cm}<{\centering}p{2.495cm}<{\centering}}
    \toprule \toprule
     \multirow{2}{*}{\textbf{Perturb}} & \textbf{Clean} &  \multicolumn{3}{c}{\textbf{Misaligned Frame Interval ($\Delta$)}} \\ \cmidrule{2-5} 
   \textbf{Mode}  & \textbf{$\Delta=0$} & \textbf{$\Delta=5$} & \textbf{$\Delta=10$} & \textbf{$\Delta=20$}       \\ \midrule
    Static &   \multirow{2}{*}{${0.082}\pm 0.179$} & ${0.069}\pm 0.168$ & ${0.066}\pm 0.154$  & ${0.065}\pm 0.163$ 
    \\
    Dynamic &  & $0.070\pm 0.157$ & $0.077\pm 0.161$ & ${0.083}\pm 0.178$ 
    \\
    \bottomrule  \bottomrule
    \end{tabular}
}\vspace{-2mm}
\end{table}

\begin{table}[ht]
\caption{RPE metric under sensor de-synchronization for  RGBD-based {ORB-SLAM3}~\cite{orbslam3}. }
\centering 
\label{tab:sensor-misalign-perturb_rpe_metric}
\resizebox{\textwidth}{!}
{
\begin{tabular}{p{1.495cm}|p{2.495cm}<{\centering}|p{2.495cm}<{\centering}p{2.495cm}<{\centering}p{2.495cm}<{\centering}}
    \toprule \toprule
     \multirow{2}{*}{\textbf{Perturb}} & \textbf{Clean} &  \multicolumn{3}{c}{\textbf{Misaligned Frame Interval ($\Delta$)}} \\ \cmidrule{2-5} 
   \textbf{Mode}  & \textbf{$\Delta=0$} & \textbf{$\Delta=5$} & \textbf{$\Delta=10$} & \textbf{$\Delta=20$}       \\ \midrule
    Static &   \multirow{2}{*}{${0.114}\pm 0.023$} & ${0.123}\pm 0.024$ & ${0.114}\pm 0.024$  & ${0.115}\pm 0.023$ 
    \\
    Dynamic &  & $0.116\pm 0.025$ & ${0.112}\pm 0.019$ & ${0.117}\pm 0.027$ 
    \\
    \bottomrule  \bottomrule
    \end{tabular}
}\vspace{-2mm}
\end{table}

\begin{table}[ht]
\caption{Success rate (SR) of pose tracking under sensor de-synchronization for  RGBD-based {ORB-SLAM3}~\cite{orbslam3}. }
\centering 
\label{tab:sensor-misalign-perturb_sr_metric}
\resizebox{\textwidth}{!}{
\begin{tabular}{p{1.495cm}|p{2.495cm}<{\centering}|p{2.495cm}<{\centering}p{2.495cm}<{\centering}p{2.495cm}<{\centering}}
    \toprule \toprule
     \multirow{2}{*}{\textbf{Perturb}} & \textbf{Clean} &  \multicolumn{3}{c}{\textbf{Misaligned Frame Interval ($\Delta$)}} \\ \cmidrule{2-5} 
   \textbf{Mode}  & \textbf{$\Delta=0$} & \textbf{$\Delta=5$} & \textbf{$\Delta=10$} & \textbf{$\Delta=20$}       \\ \midrule
    Static &   \multirow{2}{*}{${0.960}\pm 0.046$} & ${0.960}\pm 0.036$ & ${0.958}\pm 0.029$  & ${0.954}\pm 0.030$ 
    \\
    Dynamic &  & $0.955\pm 0.039$ & $0.942\pm 0.050$ & ${0.948}\pm 0.041$ 
    \\
    \bottomrule  \bottomrule
    \end{tabular}
}
\end{table}

\end{document}